\documentclass[11pt]{article}

\usepackage{graphicx}
\usepackage[position=top]{subfig}
\usepackage{calc}
\usepackage[outdir=./Graphs]{epstopdf}
\usepackage{amsmath}
\usepackage{amssymb}
\usepackage{booktabs}
\usepackage[all]{xy}
\usepackage{tcolorbox}
\usepackage{authblk}
\usepackage[ruled,vlined,commentsnumbered]{algorithm2e}
\usepackage{tikz}
\usepackage{threeparttable}

\usepackage{xr}
\usepackage[round]{natbib}
\usepackage{cleveref}

\def\argmin{\mathop{\rm argmin}}

\newcommand{\real}{\ensuremath{\mathbb{R}}}
\newcommand{\ltwo}{\ensuremath{\mathbb{L}^2}}

\newcommand{\inner}[2]{\left\langle #1,#2 \right\rangle}

\newcommand{\s}{\ensuremath{\mathbb{S}}}

\def\argmin{\mathop{\rm argmin}}

\title{\huge Elastic Shape Analysis of Brain Structures for Predictive Modeling of PTSD}
\author[a,*]{Yuexuan Wu}
\author[b]{Suprateek Kundu}
\author[c]{Jennifer S. Stevens}
\author[c]{Negar Fani}
\author[a]{Anuj Srivastava}
\affil[*]{Corresponding author: Yuexuan Wu (yw17g@my.fsu.edu)}
\affil[a]{Department of Statistics, Florida State University, Tallahassee, FL, USA}
\affil[b]{Department of Biostatistics and Bioinformatics, Emory University, Atlanta, GA, USA}
\affil[c]{Department of Psychiatry and Behavioral Sciences, Emory University, Atlanta, GA, USA}
\date{}

\begin{document}
\maketitle{}

\begin{abstract}
There is increasing evidence on the importance of brain morphology in predicting and classifying mental disorders and behavior. However, the vast majority of current shape approaches rely heavily on vertex-wise analysis that may not successfully capture complexities of subcortical structures. Additionally, the past works do not include interactions between these structures and exposure factors when modeling clinical or behavioral outcomes. Predictive modeling with such interactions is of paramount interest in heterogeneous mental disorders such as PTSD, where trauma exposure and other factors interact with brain shape changes to influence behavior. We propose a comprehensive framework that overcomes these limitations by representing brain substructures as continuous parameterized surfaces and quantifying their shape differences using elastic shape metrics. 
Using the elastic shape metric, we compute shape summaries (mean, covariance, PCA) of subcortical data and represent individual shapes by their principal scores under a PCA basis. These representations allow visualization tools that help understand localized changes when these PCs are varied. Subsequently, these PCs, the auxiliary exposure variables, and their interactions are used for regression modeling and prediction. We apply our method to data from the Grady Trauma Project (GTP), where the goal is to predict clinical measures of PTSD using shapes of brain substructures such as the hippocampus, amygdala, and putamen. Our analysis revealed considerably greater predictive power under the elastic shape analysis than widely used approaches such as the standard vertex-wise shape analysis and even a volumetric analysis. It helped identify local deformations in brain shapes related to change in PTSD severity. To our knowledge, this is one of the first brain shape analysis approaches that can seamlessly integrate the pre-processing steps under one umbrella for improved accuracy and are naturally able to account for interactions between brain shape and additional covariates to yield superior predictive performance when modeling clinical outcomes.
\end{abstract}


\newpage

\section{Introduction}\label{introduction}

Scientists have increasingly become aware of the extent and commonality of traumatic
experiences in our society. It has been reported that 
60.7\% of men and 51.2\% of women experience at least one potentially traumatic event \citep{kessler1995posttraumatic}, and a significant proportion of these events occurs during young age; 
for example, 26\% of males and 18\% of females reported having experienced at least 
one traumatic event at a young age in a representative community sample \citep{perkonigg2000traumatic}.
Of those experiencing potentially traumatic events, 10-40\% develop psychiatric symptoms of clinical relevance \citep{breslau1999previous, hoge2004combat, stein2007increased, o2008mental} such as post-traumatic stress disorder (PTSD). PTSD is the fourth most common mental disorder in USA and 
results in significant impairments of psychological and physical health \citep{kessler1995posttraumatic, sareen2014posttraumatic}, and they are most notable in adults who undergo trauma during childhood. However, one major (but often overlooked) aspect in PTSD is the presence of heterogeneity, since not everyone who experiences trauma suffers from PTSD. In order to address such heterogeneity and develop personalized risk assessment models, it is critical to incorporate information rich biomarkers such as brain imaging data in PTSD analysis.

The problem of diagnosing psychiatric disorders using brain imaging data is a widely studied topic in the literature. An important family of physical biomarkers for evaluating and predicting psychopathology comes from volumes and shapes of anatomical structures in the human brain. Subcortical structures, in particular, have been implicated in preclinical neuroscience research in behaviors and phenotypes with key relevance to mental health. Voxel based morphometry (VBM)  approaches have identified associations between PTSD and the volumes of a variety of brain regions including the amygdala, prefrontal cortex, temporal cortex, insula, thalamus, anterior cingulate cortex (ACC) and hippocampus \citep{francati2007functional, mahan2012fear, nemeroff2006posttraumatic}. All of these regions contain sub-regions with specialization in function, and likely different links to psychopathology. Sub-regions have been investigated in PTSD using automated segmentation methods \citep{van2009automated, saygin2017high}. For example, PTSD-related alterations in the hippocampus have been isolated to lower volume of the CA1 sub-field \citep{chen2018smaller}. Findings are more mixed for amygdalar subnuclei. In adult military veterans, PTSD is linked with smaller paralaminar and lateral subnuclei, but larger central, medial, and cortical nuclei which are critical to the behavioral and physiological outputs of fear \citep{morey2020amygdala}. In contrast, in youth exposed to a terror attack, PTSD symptoms were associated with smaller volumes across all major subnuclei \citep{ousdal2020association}. These findings suggest that the developmental timing and type of trauma exposure may have important effects on neural phenotypes in PTSD.  However, the analyses of subnuclei are limited by the limits in tissue contrast and spatial resolution available in typical 3T MRI research scans.  Alternative methods for understanding the morphology of subcortical regions are likely to provide important neural biomarkers of various psychiatric conditions.

There is a limited literature on brain shape changes in PTSD, although the relationship between morphologies of structures such as hippocampus and putamen shapes with disorders such ADHD, Alzheimers and Schizophrenia are well established \citep{joshi2016surface, kurtek2011classification}. The value of brain shape analysis lies in their ability to reveal local regions of variation within a structure’s surface. This is valuable complementary information to the volumetric descriptions of a structure which depict gross variation in a single direction (i.e., increased or decreased volume). The addition of localized descriptions of topology allows the detection of subtler changes in the morphometry of a surface whose signal may be lost when averaged across the whole ROI. For example although hippocampal changes are expected between PTSD and control groups, volumetric analyses in \citep{veer2015evidence, bae2019volume} found that the differences in bilateral hippocampus were not significant. In contrast, shape analysis allows investigators to report regions of equal but opposite variation within a single surface which would have otherwise been canceled out had they been reduced to a single, scalar value. However, in spite of the development of limited literature concerning the impact of brain structural and shape changes in assessing PTSD severity, there are important unanswered questions regarding how alterations in the brain structure and shape after trauma exposure results in PTSD onset and progression.

The most commonly used tools for shape analysis in PTSD involve an FSL toolbox pipeline that is utilized to identify the correlation between PTSD and subcortical volumes and shapes, most focused on amygdala and hippocampus \citep{veer2015evidence, knight2017lifetime, akiki2017association, bae2019volume, klaming2019expansion}. This pipeline, known as FIRST, is a surface-based vertex-wise shape analysis \citep{patenaude2007bayesian} method that compares the brain surface distances between the PTSD and control populations via  multivariate statistics such as radial distance and Jacobian determinant. Other methods \citep{tate2016volumetric} have used spherical shape registration tools proposed in \citep{gutman2015medial}. The latter uses a combination of spherical and medial axis representations to achieve a final surface registration. Although such analyses are useful, they essentially rely on a vertex-wise analysis that visualizes the brain surface as a collection of discrete vertices represented by voxels, which overlooks the interpretation of the brain shape as a continuously varying object in three dimensions. Moreover, due to a large number of voxels included in the multivariable analysis, it is challenging to accommodate interactions between the brain shape and confounding variables such as trauma exposure without giving rise to an inflated number of parameters in the model. Existing approaches in \citep{knight2017lifetime, klaming2019expansion} overcome this difficulty by including interactions via a voxel-wise analysis, which ignores the spatial nature of the brain shape, and requires stringent multiplicity adjustments for testing significant effects. These limitations are likely to result in biological findings that may not be reproducible across studies, especially in studies with moderate sample sizes. For example, the vertex-wise analysis in \citep{knight2017lifetime} did not find significant subcortical volume or shape differences between PTSD and control groups, although there were weakly significant interaction effects between depression and PTSD severity in the left amygdala. In contrast, \citep{veer2015evidence} found smaller right amygdala volume for PTSD versus control, while \citep{bae2019volume} discovered larger left amygdala volume for PTSD versus control group. Both studies suggested that the difference in bilateral hippocampus was not significant.

In this article, we develop a fundamentally novel approach for brain shape analysis for continuous clinical phenotypes using an {\it elastic shape analysis} method~\citep{book_AC3} that characterizes the brain shape as a continuously varying parametrized surface in 3-dimensions, instead of visualizing it as a collection of vertex-wise points. A key advantage of the elastic shape analysis framework is that it incorporates dense registration of points across objects as an in-built step. This key idea results in an integrated framework that bypasses pre-processing registration steps in shape analysis and hence prevents error propagation resulting from additional registration steps. Such an integrated registration step in the elastic shape analysis enables us to quantify pure shape variability, modulo shape-preserving transformations, and to discern subtle variations across populations by minimizing mis-registration variability. Another distinguishing feature of the proposed shape analysis is that is invariant to rotations, translations, and scaling via the use of invariant Riemannian metrics, and provides an in-built square-root normal field (SRNF) representation to align, register, and compare shapes of surfaces. Finally, the elastic shape analysis provides optimal deformations or geodesics between anatomical structures, providing important tools for visualizing and investigating shape differences.

From the perspective of statistical analysis, elastic shape analysis can be used to compute statistical means and covariances of shapes sampled from a population. The characterization of covariance automatically paves the way for a classical principal component analysis (PCA), which provides an efficient representation of shapes by their principal coefficients. The set of principal coefficients (PC) provides parsimonious lower dimensional representation of the brain shape, and a subset of PCs that adequately represents the brain shape can then be used in any classification or prediction approach, along with confounding variables such as age and trauma exposure. By including interactions between a subset of PCs and confounding variables, the proposed approach provides a parsimonious and flexible classification or prediction approach that is not restricted to vertex-wise analysis and enables non-linear associations between the brain shape and potential confounders and their interactions with the clinical phenotype of interest. This results in superior prediction performance compared to state of the art shape analysis methods in PTSD literature.

An important strength of these shape features is that one can reconstruct full shapes from these feature vectors (principal components) via mathematical invertibility inherent under the elastic shape analysis method. Consequently, one can visualize changes in shapes of a structure by varying only one or multiple features at a time. This provides an important tool for physicians and clinicians to visualize localized changes or deformations in the brain anatomy for statistically significant shape features or principal components. Furthermore, one can use this tool to visualize and analyze salient deformations in brain sub-structures such as the hippocampus, putamen, amygdala in relation to changes in PTSD clinical measurements. Starting from T1-weighted MRI data, we extract, analyze and represent objects of interest from images and perform a comprehensive statistical analysis of these objects. This process involves representing objects as parameterized surfaces, and solving for optimal registrations of points across surfaces. Multiple surfaces are registered by computing a mean surface, and registering individual surfaces to this mean, in a recursive fashion. In our analysis, we focus on a pre-specified subset of subcortical structures and their interactions with demographic and exposure confounding variables for classification and prediction of PTSD severity. In contrast to most existing studies that focus on brain shape changes in military veterans, our study is one of the first to investigate brain shape changes in PTSD in conjunction with co-morbidities such as trauma in a civilian minority population of AA females.


\section{Materials and Methods}
\begin{figure}[!ht]
		\centering
		\fbox{
 		\includegraphics[width = \linewidth]{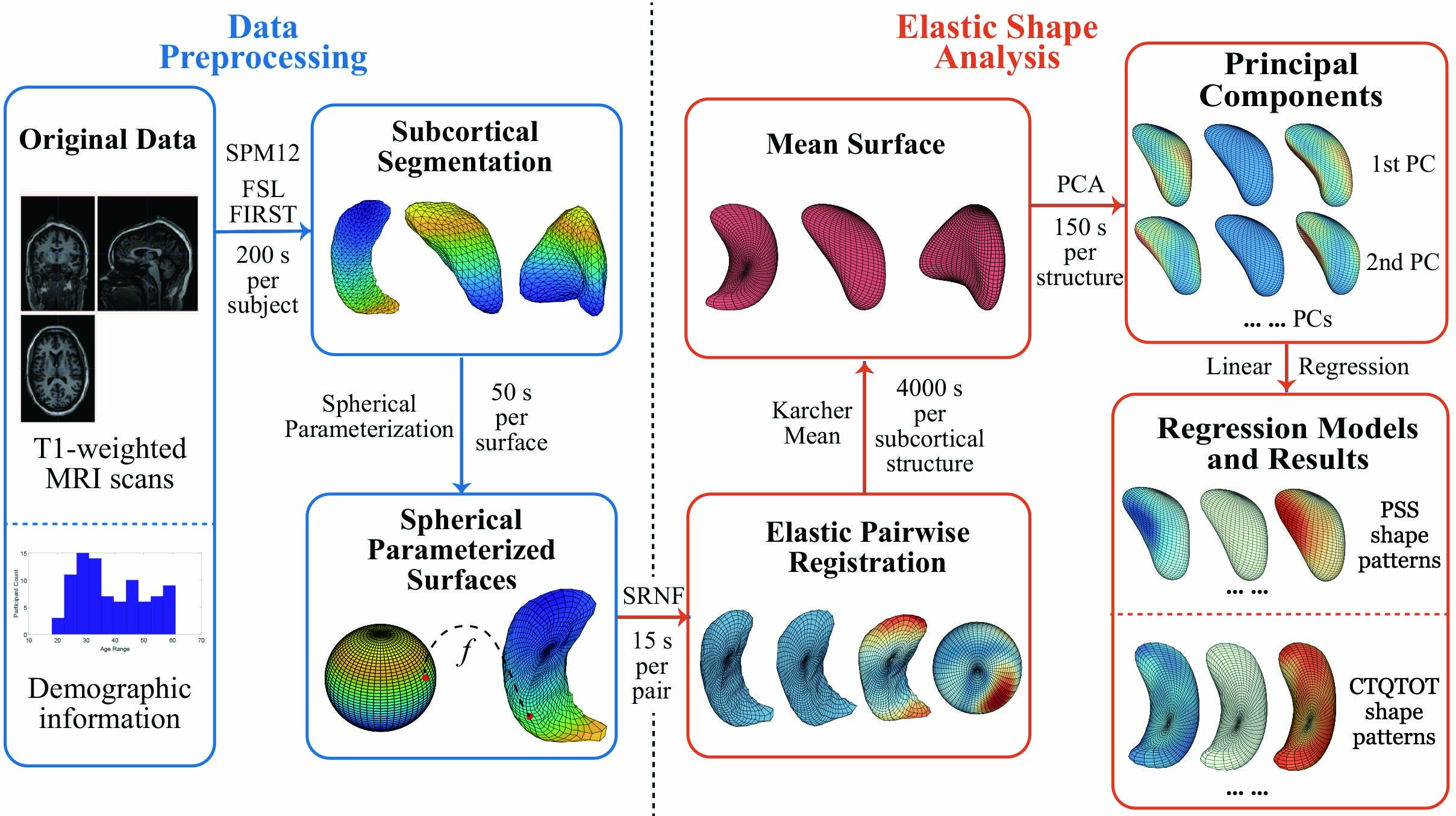}
		}
			\caption{Pipeline steps and time cost. The time cost is computed with Intel i7-8705G. SRNF: square root normal field, PCA: principal component analysis, PC: principal component, PSS: PTSD symptom scale, and CTQTOT: childhood trauma questionnaire total score.} 
		\label{figure:pipeline}
\end{figure}

In this section, we will describe the full pipeline for extracting and analyzing shapes of 
subcortical brain surfaces. 
This pipeline is illustrated in Fig.~\ref{figure:pipeline} with the time costs 
computed on a laptop with Intel i7-8705G processor. As the figure shows, 
the proposed pipeline has three main steps: (i) pre-processing of the original data; 
(ii) registration and shape analysis of 3D surfaces; and (iii) regression models for analysis 
of post-traumatic stress disorder. We describe these steps next, starting with an introduction
of the data used
in the experiments presented later. 

\subsection{Data Description}
In this study, we include 90 experimental subjects' T1-weighted MRI scans of brain. The dataset contains demographic information with the questionnaire results on PTSD symptoms and traumatic experiences for each experimental subject.\\

\noindent {\bf T1-weighted MRI scans}:
The original data is T1-weighted MRI scans acquired by Emory University Grady Trauma Project using Siemens Tim Trio \citep{enigma}. Field of view is $224mm \times 256mm$, while repetition time and echo time are 2600 ms and 3.02 ms separately.
\\

\begin{figure}[!ht]
\centering
\includegraphics[height=5cm]{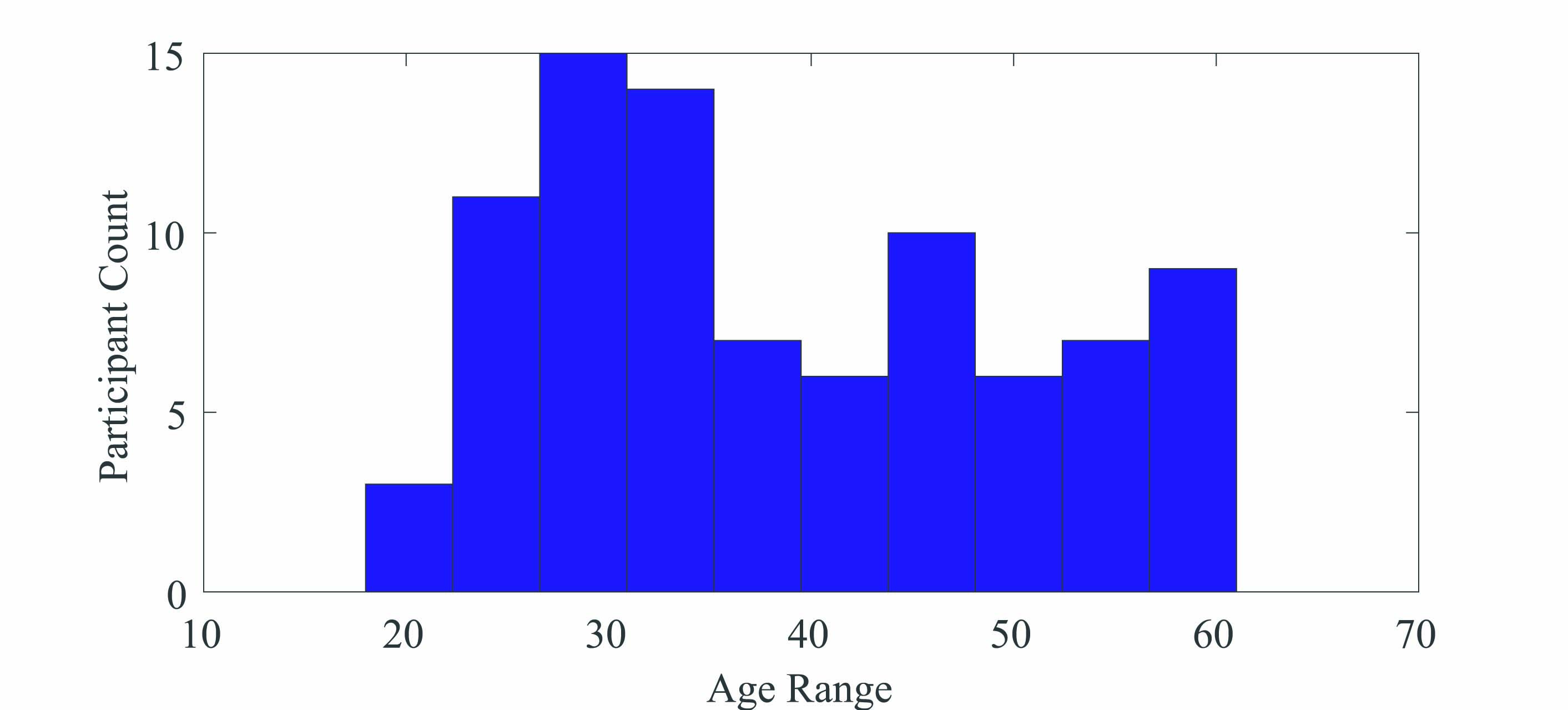}
\caption{Histogram of participants' age. Age ranges from 18-61 and the average age is 38.6 years old.}
\label{figure:age_hist}
\end{figure}

\noindent {\bf Demographic Information}:
Participants of the experimental data collection are all African American women, so sex and race are all set to 1. 
The other demographic information included in the data are: 
age (18-61), education (0-5), employment (0,1), disability (0,1) and income (0-4). 
The average age of participants is 38.6 years old and Fig.~\ref{figure:age_hist} displays a histogram of participants' age. 
\\

\noindent  {\bf Questionnaire Results}
In the study, we are most interested in participants' answers to questionnaires that are related to PTSD symptoms and traumatic experiences. Specifically, we focus on three questionnaire results: (i) PTSD Symptom Scale (PSS) \citep{fani}, which measures the presence and frequency of current PTSD symptoms and has a range of 0-42; (ii) Childhood Trauma Questionnaire \citep{CTQ} Total Score (CTQTOT), which is a 25-item inventory of different types of childhood maltreatment including abuse and neglect, and has a range of 25-125; and (iii) Beck Depression Inventory (BDI), which is a 21-question multiple-choice self-report inventory and has a range of 0-63.

\subsection{Data Pre-processing}
Here we describe the pre-processing steps for extracting subcortical structures from brain imaging data. 
We use state-of-art packages and software to pre-process the MRI scans obtained as the original data. 
We first convert 176 DICOM scan files for each subject into a single NIfTI file using 
SPM12 \citep{SPM12} . The NIfTI images each have a resolution of 240 $\times$ 256 $\times$ 176.

Next, we utilize the FMRIB Software Library (FSL), which is a software library containing image analysis and statistical tools for functional, 
structural and diffusion MRI brain imaging data. 
Among the tools in FSL, FSL FIRST \citep{FSL} is a model-based segmentation/registration tool. FSL FIRST can segment a T1-weighted MRI image into $15$ subcortical structures' surfaces. Using some manually segmented images, in which the subcortical labels are parameterized as surface meshes and modeled as a point distribution model, FSL FIRST trains an automatic segmentation model using a
Bayesian approach. The inputs of FSL FIRST are T1-weighted MRI images in NIfTI file format, and the outputs are triangular meshed surfaces of 15 subcortical brain structures. The labeled subcortical structures are: thalamus proper, caudate, putamen, pallidum, hippocampus, amygdala and accumbens area for both left and right cerebral hemispheres and brain stem. 
Although there are several structures available for study, this paper mainly focuses on three 
structures: left hippocampus, left amygdala and left putamen.
The outputs of FSL FIRST are $m \times 3$ vertex coordinates and $(2m-4) \times 3$ edges. In this data, $m$ is 732 for hippocampus surfaces and 642 for all other subcortical structures.

\begin{figure}[!ht]
		\centering
		\resizebox{.9\columnwidth}{!}{
		\begin{tikzpicture}
				\node at (0,0) {\includegraphics[width=4cm]{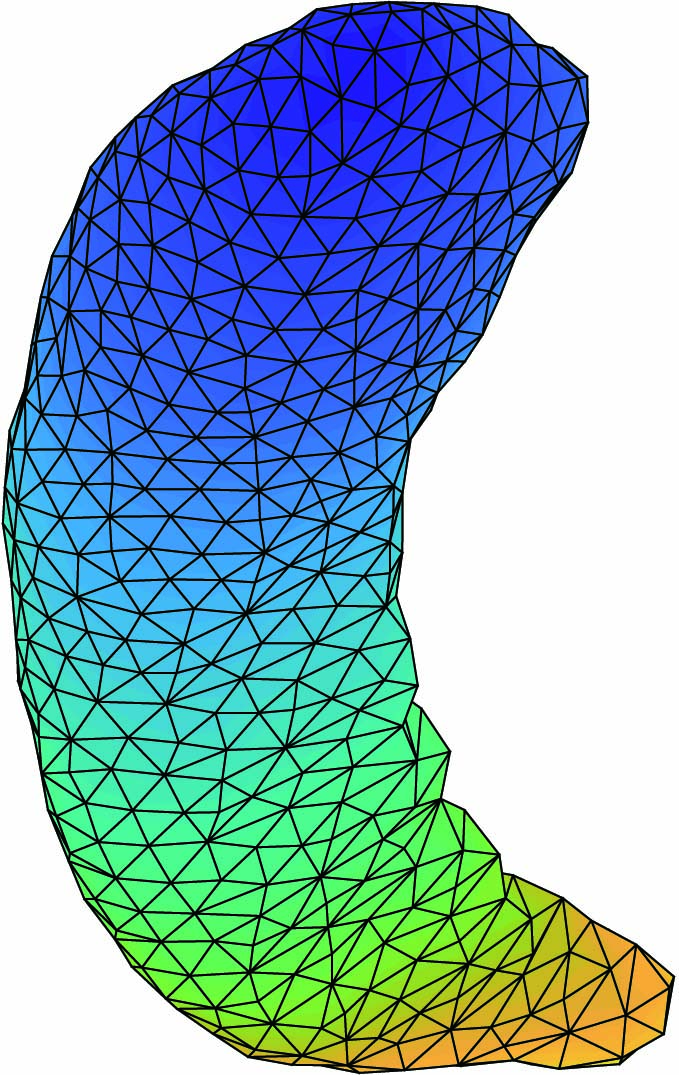}};
				\node at (5,0) {\includegraphics[width=4cm]{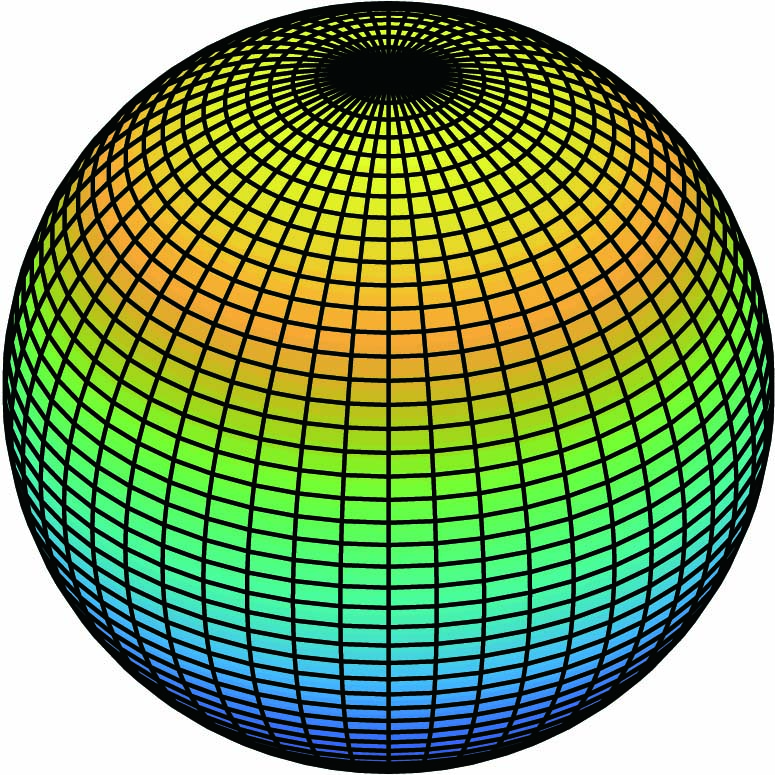}};
				\node at (11,0) {\includegraphics[width=4cm]{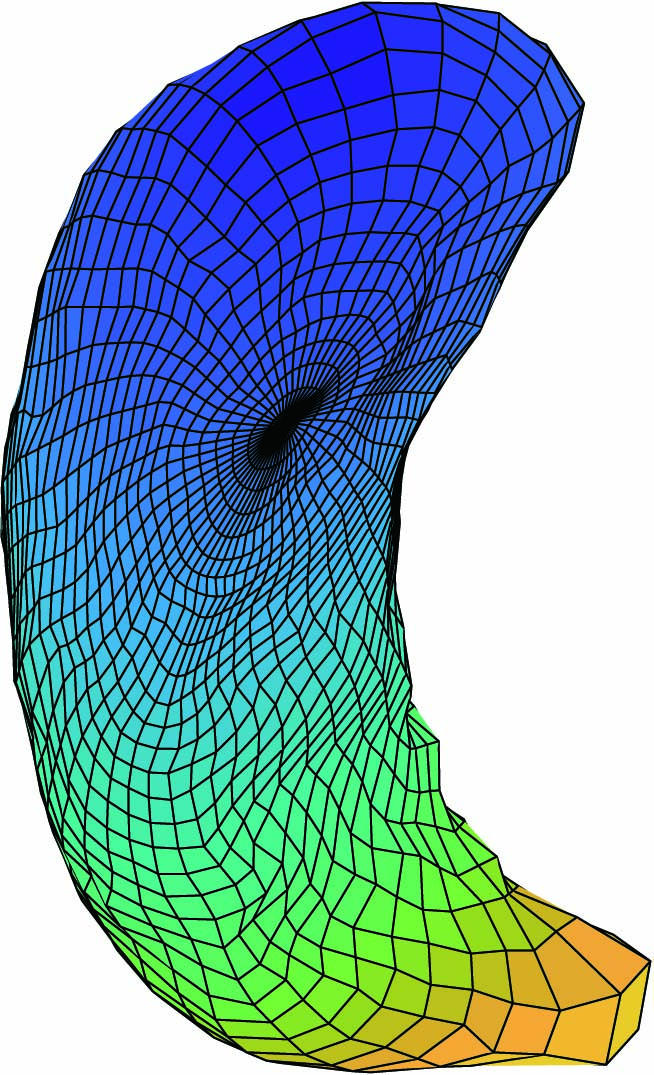}};
				\filldraw [red] (6.5,1.2) circle (3.5pt);
				\filldraw [red] (12.4,-2.35) circle (3.5pt);
				\filldraw [green] (4.3,-1.5) circle (3.5pt);
				\filldraw [green] (10,1.2) circle (3.5pt);
				\filldraw [yellow] (5.5,-1.5) circle (3.5pt);
				\filldraw [yellow] (11.5,1.9) circle (3.5pt);
				\draw (0,4) node {\large (a) Triangular Mesh Surface}
						(5,4)	node {\large (b) Unit Sphere} 
						(11,4)	node {\large (c) Spherical Parameterized Surface};
		\end{tikzpicture} }
		\caption{Triangular mesh surface and spherical parameterized surface. Triangular mesh surface has $732 \times 3$ vertices and spherical parameterized surface has a $51 \times 51$ mesh. Each vertex on the spherical parameterized surface corresponds with a vertex on the $51 \times 51$ mesh unit sphere $\s^2$.}
		\label{Triangular_2_Sphere}
\end{figure}
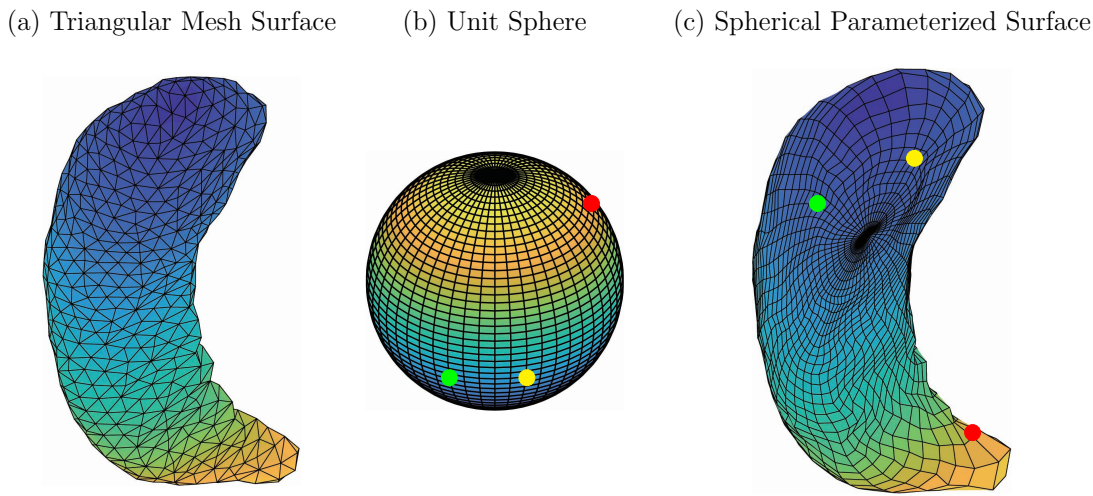

We then apply the spherical conformal mapping and tuette mapping algorithms in \citep{book_AC3} to transform the triangulated mesh into {\it spherically-parameterized} surfaces. The method first creates progressive mesh structure with the triangular mesh, and then embeds the mesh vertices into a sphere. The surfaces are called spherical parameterized since each point on the surface corresponds with a point on the unit sphere $\s^2$. This provides a representation of the surface as an 
embedding: $f : \s^2 \rightarrow \real^3$. Figure \ref{Triangular_2_Sphere} illustrates the spherical parameterization of triangulated meshes. Column (a) shows an example of triangular mesh surface, column(b) and (c) are sphere $\s^2$ and the corresponded spherical parameterized surface with $51 \times 51$ mesh. Points in the same color indicate the corresponding relationship.

\subsection{Framework: Elastic Shape Analysis}
To analyze shapes of subcortical surfaces and to discern shape changes with traumatic experiences, 
we utilize an {\it elastic shape analysis} approach, developed in the book~\cite{book_AC3}. This comprehensive theory provides several 
tools for analyzing shapes of 3D objects, including (1) metric for quantifying differences in their shapes, (2) deforming objects into each other using optimal paths,  
(3) optimally registrating points across surfaces being compared, and (4) computing
mean, covariance, and PCA of shapes. An important aspect is that these 
results are invariant to the chosen shape-preserving transformations (rigid motions, 
global scaling, and parameterizations of surfaces). In the past, this framework has been applied to shape 
analysis of human bodies \citep{laga-TPAMI-2017} and brain morphology associated with Alzheimers \citep{joshi2016surface}
and ADHD \citep{Sebastian-TMI-2012}. 

\subsubsection{Elastic Metric and SRNF}
\begin{figure}[!ht]
		\centering
		\resizebox{.8\columnwidth}{!}{
		\begin{tikzpicture}
			\node at (0,0) {\includegraphics[height=5cm]{Unit_sphere.jpg}};
			\node at (9,0) {\includegraphics[height=5cm]{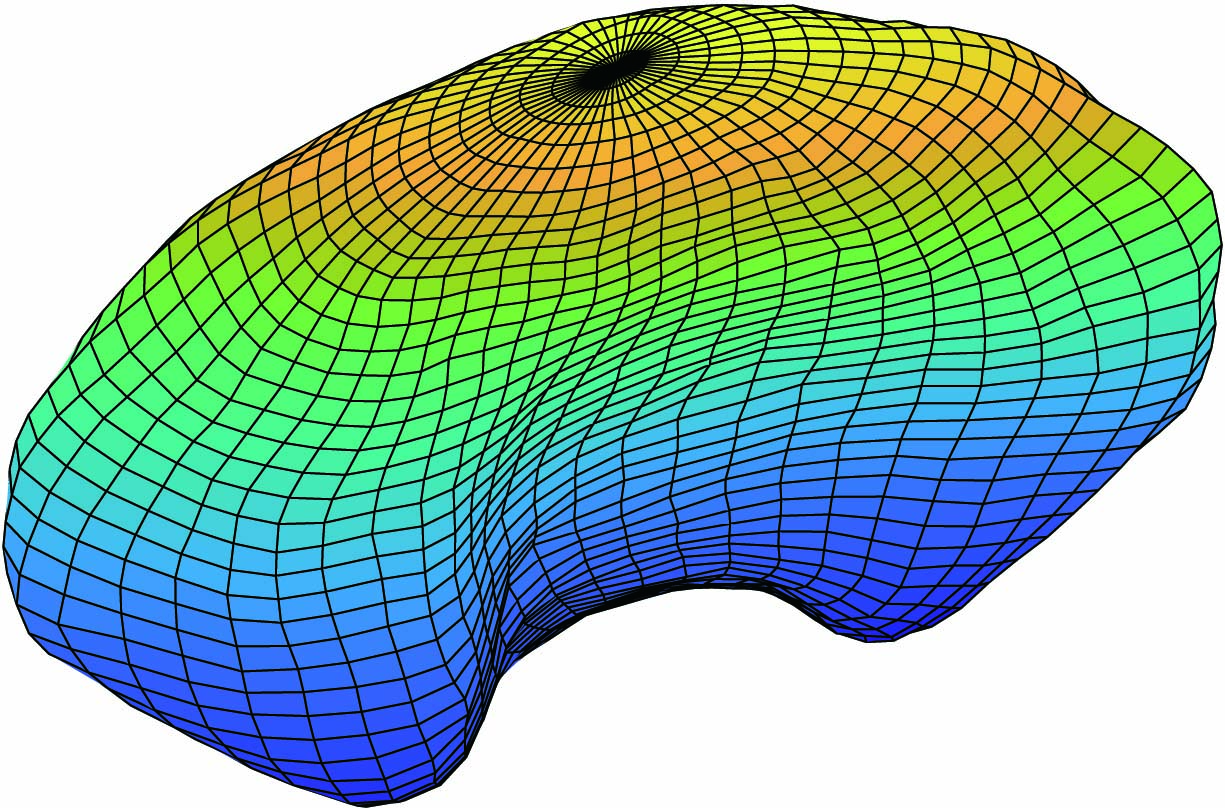}};
			\filldraw [black] (0.51,-0.5) circle (3.5pt) node[{below left}] {\huge $s$};;
			\draw[->,ultra thick] (0.51,-0.5) -- (3.1,-0.5) node (xaxis) [right] {\LARGE $\frac{\partial}{\partial v}$};
			\draw[->,ultra thick] (0.51,-0.5) -- (0.51,2.8) node (yaxis) [right] {\LARGE $\frac{\partial}{\partial u}$};
			\filldraw [black] (10.5,1.2) circle (3.5pt) node[{below right}] {\huge $s$};
			\draw[->,ultra thick] (10.5,1.2) -- (12.3,2.5) node (xaxis) [right] {\LARGE $f_v$};
			\draw[->,ultra thick] (10.5,1.2) -- (9.3,3) node (yaxis) [right] {\LARGE $f_u$};
			\draw[->,ultra thick] (10.5,1.2) -- (10.2,0) node (norm) [right] {\LARGE $\bold{n}$};
			\draw [dotted,->,ultra thick,out=70,in=-220] (0.51,-0.5) to node[below]{\LARGE $f$} (10.5,1.2);
			\draw (0,-3) node {\Large $\s^2$};
			\draw (9,-3) node {\Large $\mathbb{R}^3$};
		\end{tikzpicture} }
		\caption{Local geometry of a parameterized surface viewed as a mapping $f: \s^2 \to \real^3$. 
		The figure shows the tangent vectors $f_v$ and $f_u$ and the normal vector $\bold{n}$ at point $f(s)$ on $f$.}
		\label{Embedding and normal}
\end{figure}

Next we present some salient ideas of this approach. 
The subcortical objects are considered as closed surfaces in $\real^3$. Each closed surface can be represented in a parametrized form using a smooth map: $ f : \s^2 \rightarrow \real^3$. Let $\mathcal{F}$ denote the space of all such surfaces. If $s=(u,v)$ is a point on the sphere $\s^2$, then the partial derivatives $f_u$ and $f_v$ denote tangent vectors to the surface $f$ at the point $f(s)$. The (unnormalized) normal vector at point $s$ is given by $\bold{n}(s)=f_u \times f_v$, where $\times$ indicates the cross product in $\real^3$. Figure~\ref{Embedding and normal} shows a surface $f$, parametrized by points on a unit sphere $\s^2$, 
with the tangent and normal vectors at point $f(s)$ on the surface.

Let $\Gamma$ be the set of all orientation-preserving diffeomorphisms of $\s^2$; the elements of $\Gamma$ help us re-parameterize surfaces. For any parameterized surface $f\in\mathcal{F}$ and a $\gamma \in \Gamma$, 
the composition $f \circ \gamma$ denotes a re-parameterization of $f$. Equivalently, elements of $\gamma$ also help in a dense registration of points across surfaces. Initially, for any $s \in \s^2$, the point $f_1(s)$ on $f_1$ is said to be registered to the point $f_2(s)$ on $f_2$.
However, if we re-parameterize $f_2$ by $\gamma$, then the point $f_1(s)$ is now registered to the point $f_2(\gamma(s))$ on $f_2$. Thus, $\gamma$ becomes a tool for controlling the registration between $f_1$ and $f_2$. 
The next question is: How can we find the best registration between any two surfaces $f_1$ and $f_2$? A related question is: What should be the objective function for defining and calculating the optimal $\gamma$ that best registers $f_2$ with $f_1$? An obvious choice would be the $\ltwo$ norm, leading to a registration problem of the type: 
$
\argmin_{\gamma \in {\Gamma}} \| f_1 - f_2 \circ \gamma\|^2$.
However, despite being popular, this objective function has some fundamental limitations. It is degenerate and leads to singularities in solutions. While one can impose additional penalties to avoid degeneracy, some basic flaws of this solution remain. Specifically, the registration solution is not inverse symmetric. That is, the registration of surface $f_1$ to $f_2$ may not be consistent with the registration of surface $f_2$ to $f_1$. This is problematic for the interpretation of results. 
From a mathematical perspective, the problems in using the $\ltwo$ norm for registering surfaces stem from the following fact. 
In general, for any $f_1,f_2 \in \mathcal{F}$ and $\gamma \in \Gamma$, we have: 
$
\left\| f_1-f_2 \right\| 
 \ne 
\left\| f_1\circ\gamma-f_2\circ\gamma \right\|$.
In the other words, we lose some information about the shape of the surfaces after re-parameterization if we use the $\mathbb{L}^2$ distance to compare them. 

A significantly better solution for registration and shape analysis comes from an elastic Riemannian metric. 
While this metric's original form is too complex for practical usage, a square-root representation of surfaces simplifies their usage. 
This representation, termed the square root normal field (SRNF), is defined as follows: for $s \in \s^2$, define
\begin{equation} \label{SRNF}
q(s) = \frac{n(s)}{\left| n(s) \right| ^\frac{1}{2}} \ ,
\end{equation}
where $n(s)$ is the normal at a point $f(s)$ as explained earlier. Thus, $q$ is nothing a but a normal vector field on the surface $f$ with the magnitude given by $\sqrt{|n(s)|}$.
If $q$ is the SRNF of a surface $f$, then the SRNF of the re-parameterized surface $f \circ \gamma$ is 
given by:
\begin{equation} \label{SRNF_repara}
(q \star \gamma)(s) \equiv Q(f \circ \gamma)(s) = \sqrt{\mathbf{J} \left[ \gamma(s) \right]} (q \circ \gamma)(s),
\end{equation}
where $\mathbf{J}[\gamma(s)]$ is the Jacobian of $\gamma$ at $s$. One of the fundamental advantages for using SRNF representation is the following. Let $q_1$ and $q_2$ denote SRNFs of two surfaces $f_1$ and $f_2$, respectively. Then, 
 the $\mathbb{L}^2$ norm between SRNF satisfies the condition: 
 $$
 \left\| q_1-q_2 \right\| = \left\| (Oq_1\star \gamma)-(Oq_2 \star \gamma) \right\|\ ,
 $$ 
for all 3D rotations $O \in SO(3)$ and all $\gamma \in \Gamma$. This property allows us to 
define a {\it shape metric} between surfaces $f_1$ and $f_2$ according to:
\begin{equation} \label{SRNF_distance}
d_s(f_1,f_2) = \inf_{(O,\gamma) \in SO(3)\times\Gamma} \left\| q_1 - O(q_2,\gamma) \right\| = \left\| q_1 - O^*(q_2,\gamma^*) \right\|,
\end{equation}
where $q_1$ and $q_2$ are SRNFs of surface $f_1$ and $f_2$. 
In other words, we fix one of the surfaces and find the best rotation and re-parameterization of the second surface so as to minimize this distance. 
Note that if we apply an arbitrary rotation or re-parameterization to either $f_1$ or $f_2$ or both, their shape distance remains unchanged!

The computation of shape metric in Eqn.~\ref{SRNF_distance} requires solving for the optimal $O^*\in SO(3)$ and $\gamma^* \in\Gamma$. We use the Procrustes method to solve for the optimal rotations of $f_1$ while keeping $\gamma\in\Gamma$ fixed, and we use a gradient descent approach to optimize over $\Gamma$, whose details and algorithms are presented in \citet{book_AC3}, 
while keeping the rotation fixed. 
After finding the optimal $O^*\in SO(3)$ and $\gamma^* \in\Gamma$, and applying them to surface $f_2$ constitutes surface registration and alignment. The aligned surface is $f_2^* =  O^*(f_2 \circ \gamma^*)$, and the distance between $f_2^*$ and $f_1$ is the shortest among all $O(f_2 \circ \gamma)$ for any $O \in SO(3)$, $\gamma \in \Gamma$. 
\begin{figure}[!ht]
		\captionsetup[subfloat]{position = top}
		\centering
		\resizebox{.85\columnwidth}{!}{
		\subfloat[\large Original $f_1$]
 		{\includegraphics[height = 4.7cm]{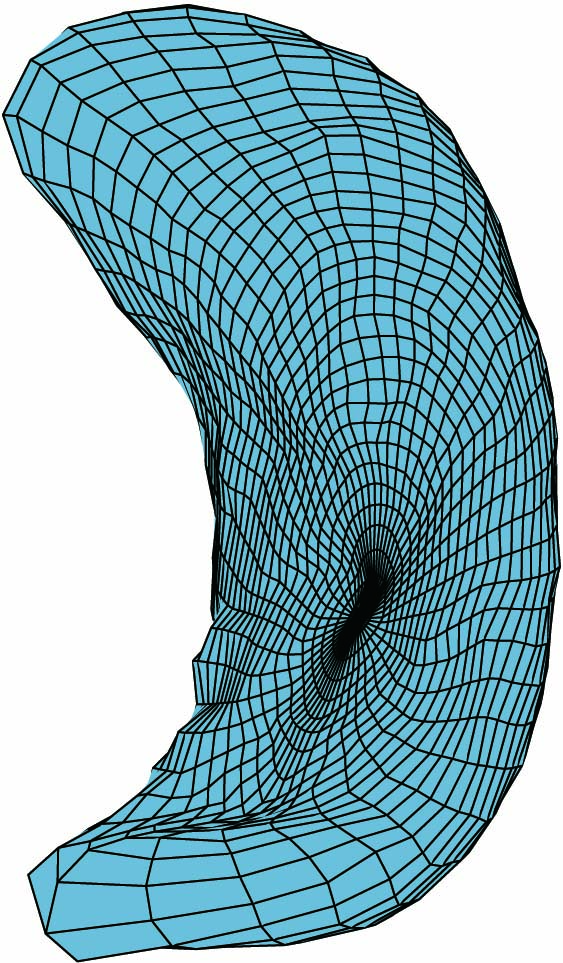}}
		\hspace{0.5cm}      
		\subfloat[\large Original $f_2$]
   	{\includegraphics[height = 4.7cm]{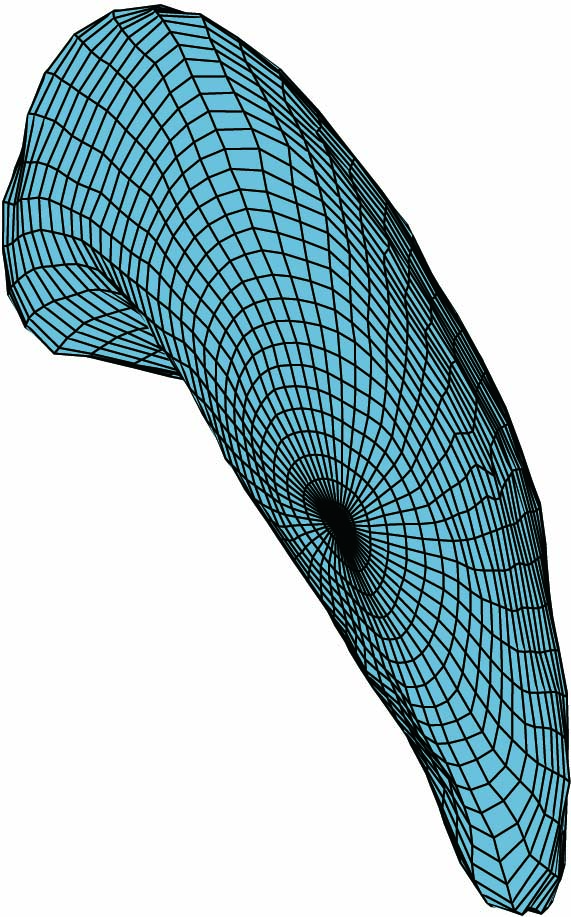}}
		\hspace{1.4cm}     
		\subfloat[\large Aligned $f_2^*$]
 		{\includegraphics[height = 4.7cm]{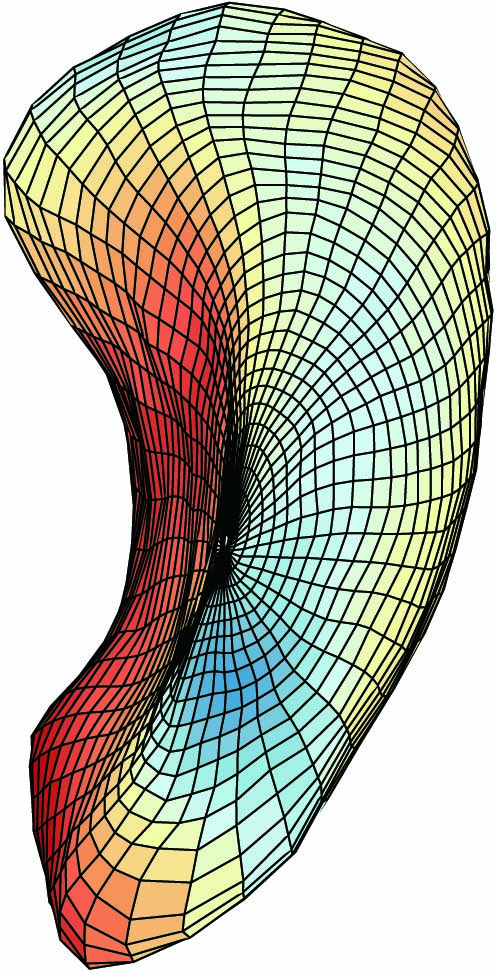}}
		\hspace{0.5cm}      
		\subfloat[\large Optimal $\gamma^*$]
 		{\includegraphics[height = 4cm]{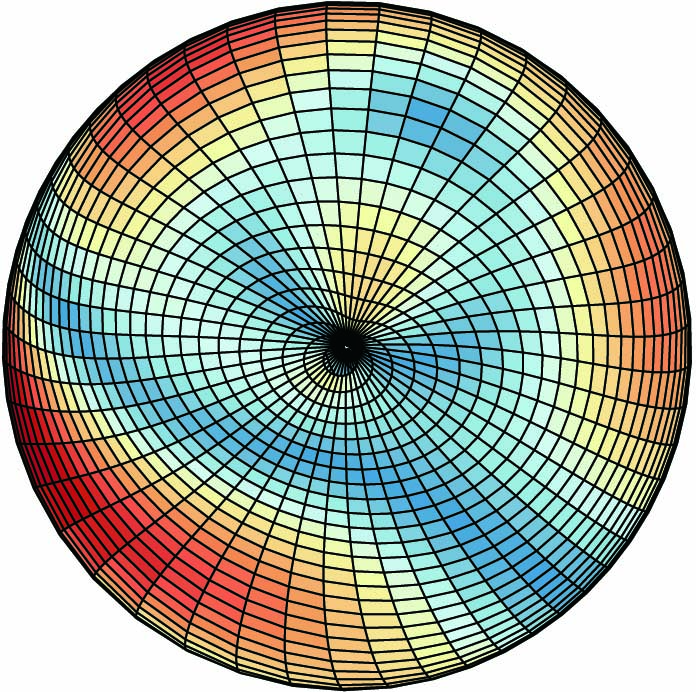}}
      \hspace{0.5cm}         
 	 \subfloat{\includegraphics[height = 4cm]{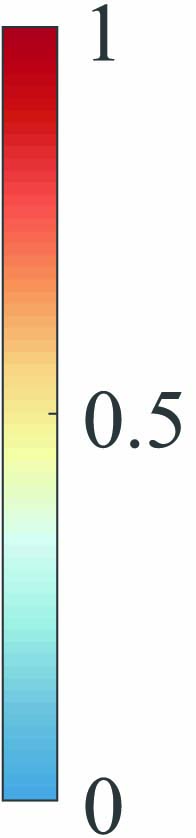}}
	 }
	 	\resizebox{.85\columnwidth}{!}{
		\subfloat{
 		\includegraphics[height = 4.5cm]{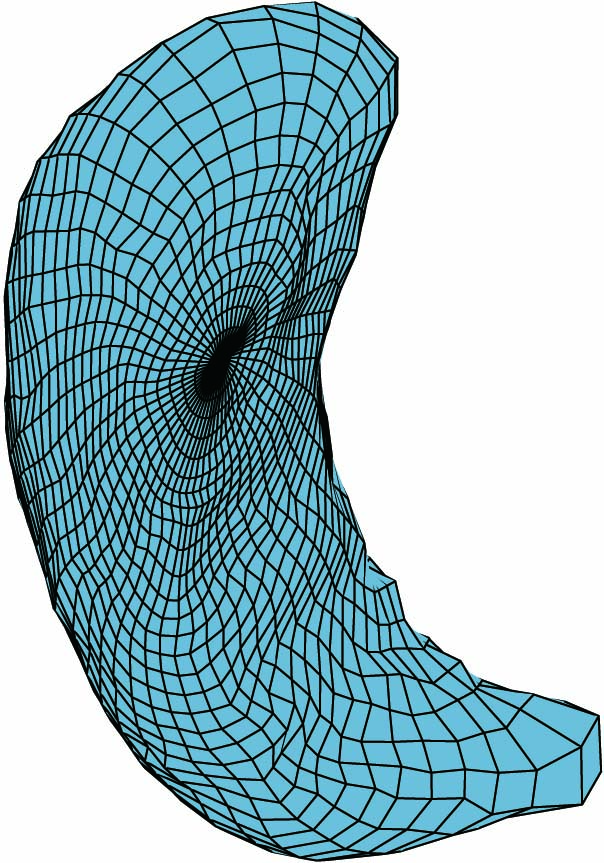}
		\hspace{0.1cm}
 		\includegraphics[height = 4.5cm]{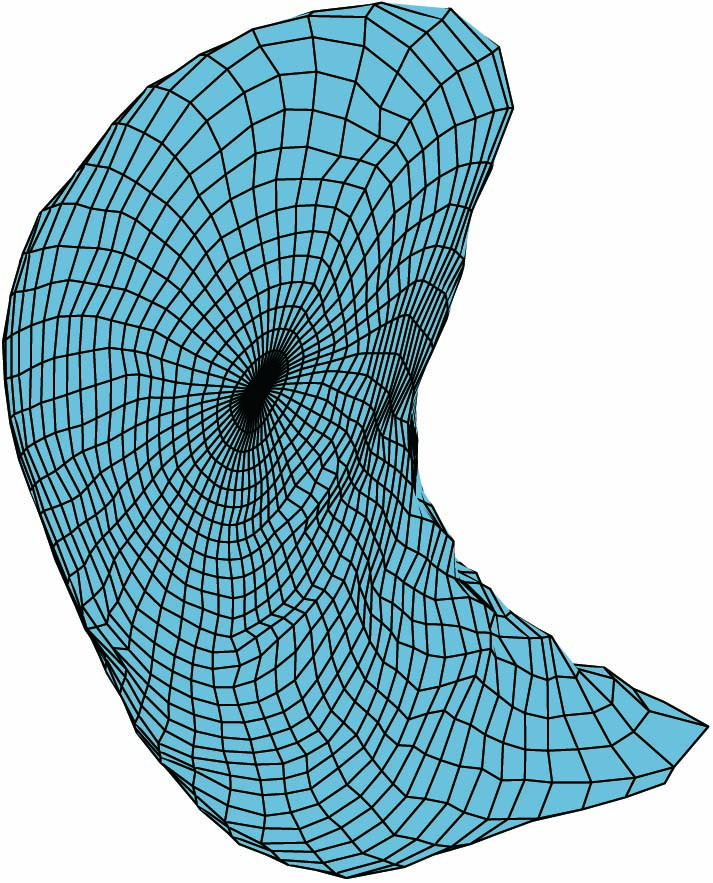}
		\hspace{0.1cm}
 		\includegraphics[height = 4.5cm]{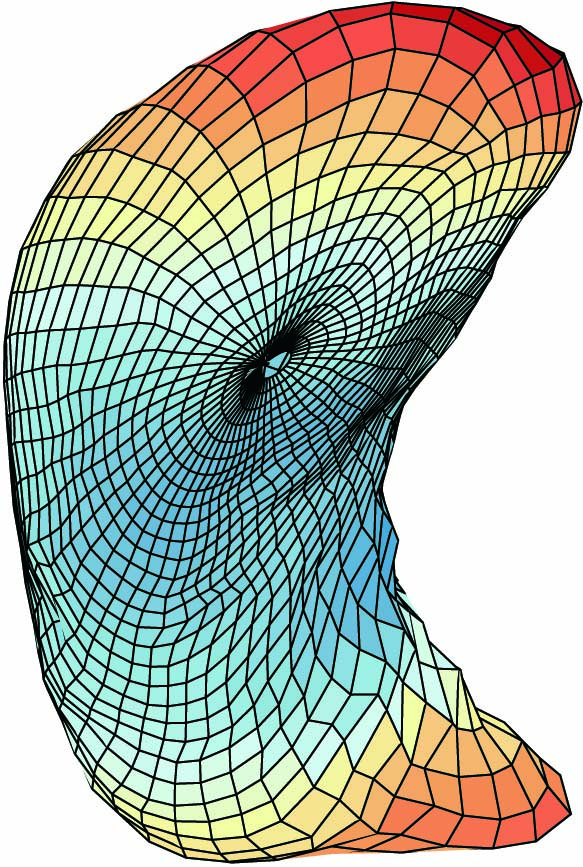}
		\hspace{0.1cm}
 		\includegraphics[height = 4cm]{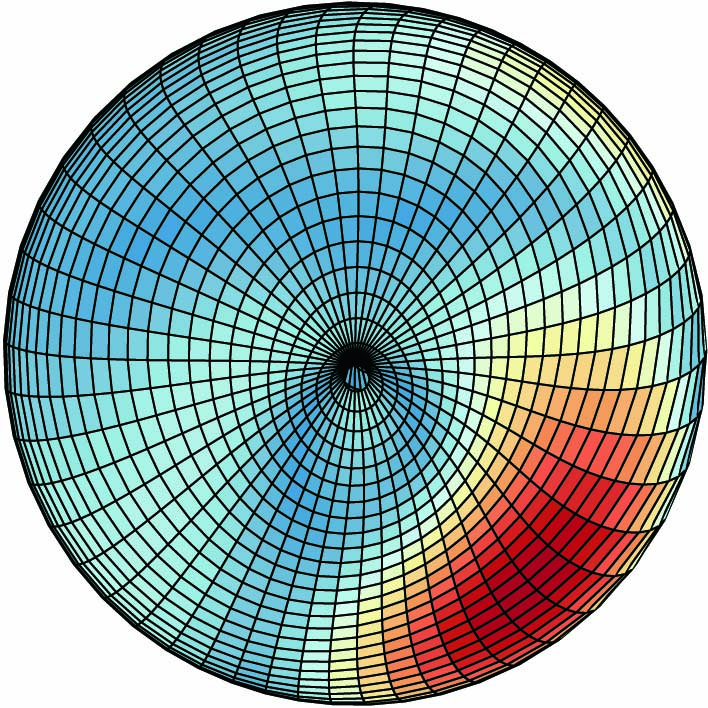}
     \hspace{0.4cm}
 		\includegraphics[height = 4cm]{Colorbar1.jpg}
		}
		}
		\caption{Two examples of elastic surface alignment and registration. In each row: (a) the original template surface $f_1$; (b) the original surface $f_2$ to be aligned; (c) the optimal re-parametrized surface $f_2^*$; and (d) the optimal re-parameterization function $\gamma^*$ shown on the unit sphere. Colors indicate the relative determinant of the Jacobian of re-patameterization function.}
		\label{Alignment_example}
\end{figure}

Figure \ref{Alignment_example} illustrates two examples of pairwise surface registration, where $f_2$ is registered with $f_1$. 
Columns (a) and (b) show the original $f_1$ and $f_2$, and column (c) shows $f_2^*$, the surface after optimal re-parameterization and rotation. The corresponding re-parameterization function $\gamma^*$ is shown in column(d), where the color indicates the relative determinant of the Jacobian $\det \mathbf{J}(\gamma)$ of re-parameterization function $\gamma^*$. The first example is a left putamen surface registered with a left hippocampus surface, and the second example is the pairwise elastic registration of two left hippocampus surfaces. From the first example, we can observe that left putamen's two outermost parts are registered with anterior and posterior ends of left hippocampus respectively. From the second example, we can observe that anterior, middle and posterior part of two surfaces are well-aligned, which makes it possible to compare two surfaces' shape. Since two surfaces are more different in anterior end and posterior end, re-parameterization acts more prominently in these parts and have larger determinant of the Jacobian.

\subsubsection{Elastic Registration: A Simulation Study}
\begin{figure}[!ht]
		\centering
			\fbox{
			\subfloat[\large Heat Map]{
					  	 		\includegraphics[height=4.5cm]{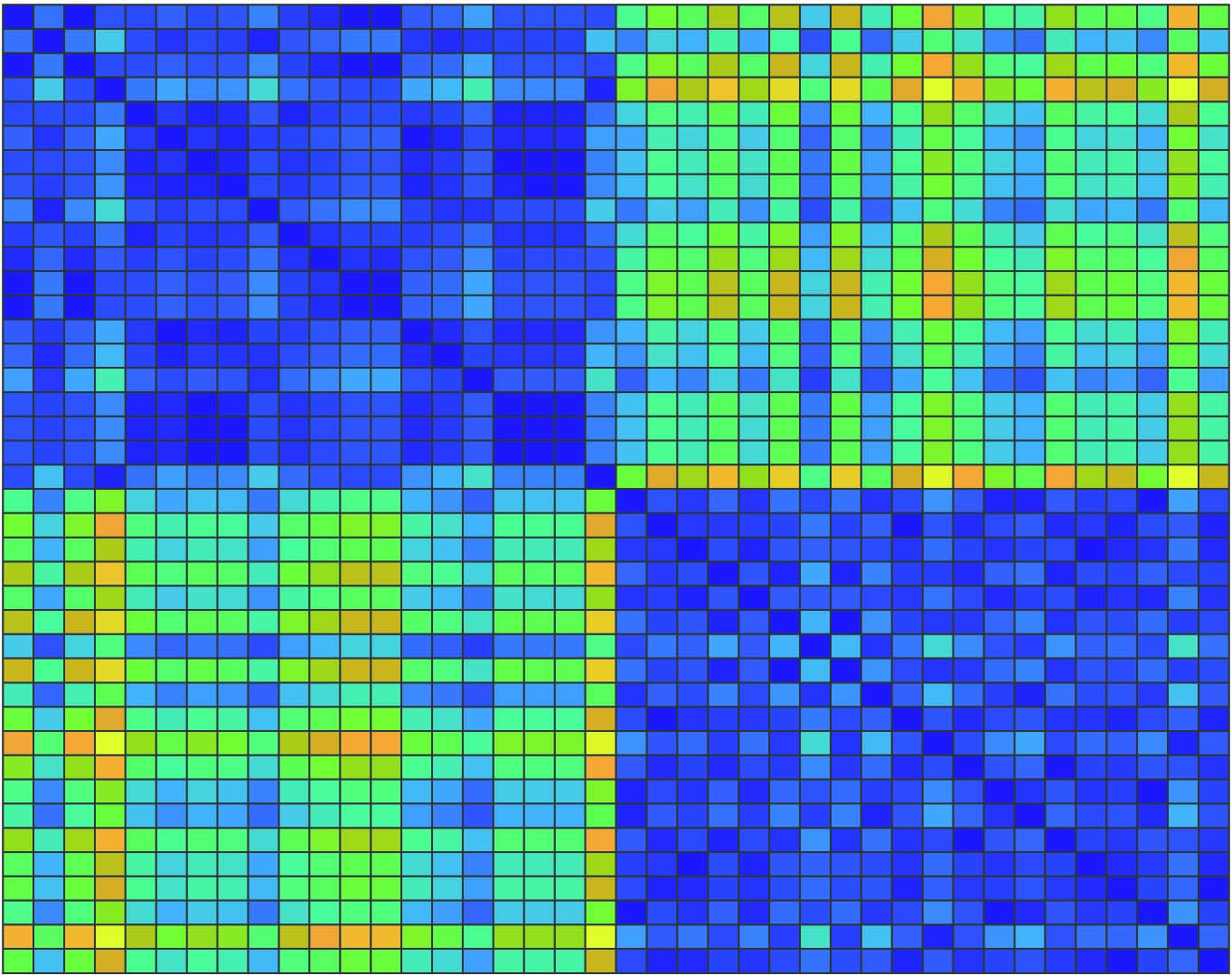}
					  			}}
		  \qquad
			\fbox{
			\subfloat[\large MDS Plot]{
					  	 		\includegraphics[height=4.5cm]{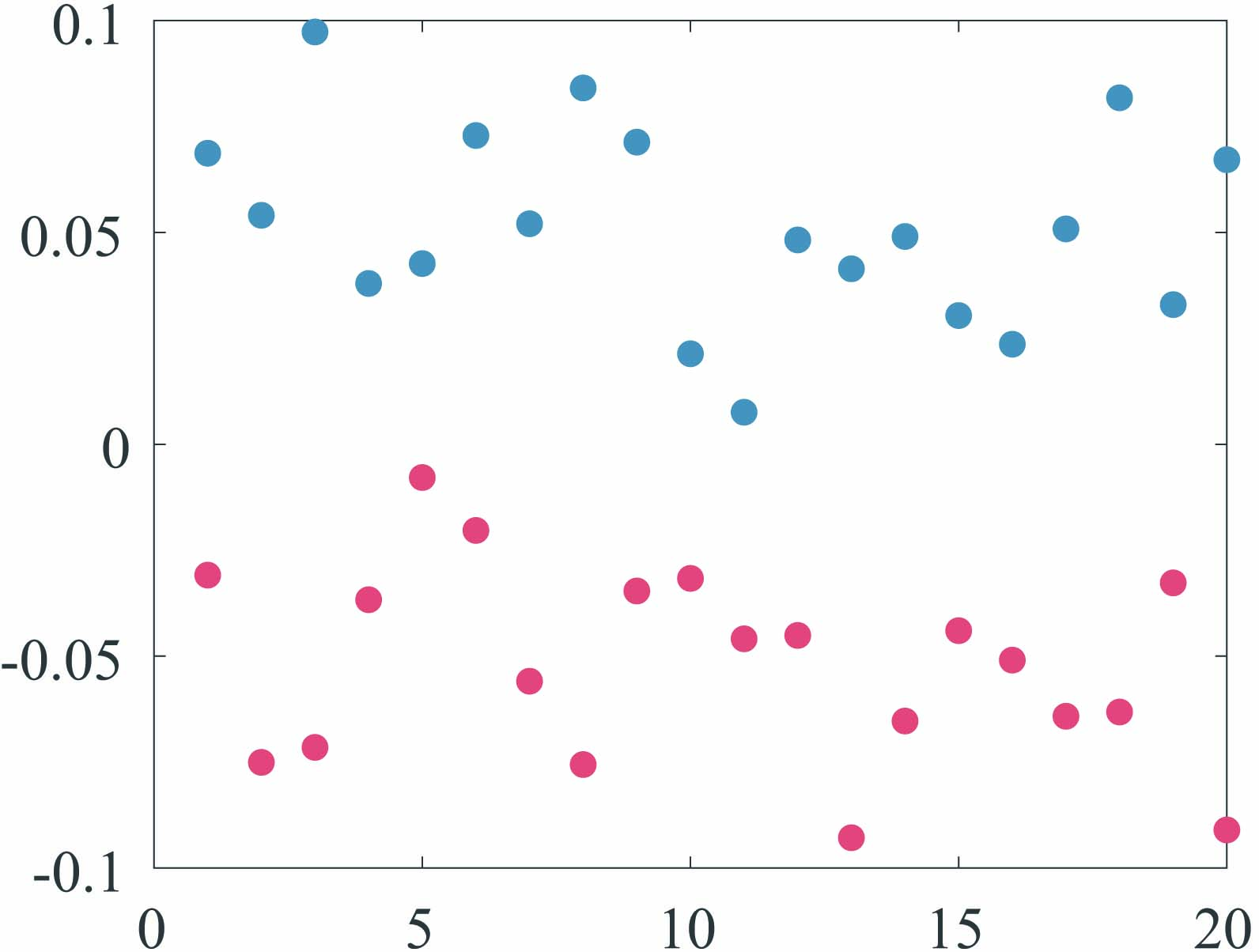}
					  			}}
   \caption{(a) Heat map plot and (b) MDS plot of distances between well aligned and registered surfaces. Colors of the heat map indicates the relative distances between surfaces. Blue dots in MDS plot indicate the first 20 simulation surfaces and red dots indicate the last 20 simulation surfaces.}
   \label{simu_before}
\end{figure}

Here, in order to illustrate and validate the necessity of surface registration, we conduct some simulation studies. We randomly generate 40 simulation surfaces using PCA representations of shapes of left hippocampus (the use of PCA is detailed later in this paper). More
specifically, we use only the first principal direction in this experiment and generate 20 surfaces each on either side of mean along that direction. That is, we generate ${f}_i = \mu + {x}_iv_1$, $i=1,2,\dots, 40$, where  
${x}_i \in (0,1]$ for $I \leq 20$ amd ${x}_i \in [-1,0)$ for $i > 20$. 
%
%
Since PCA is performed after surface registration, these simulated surfaces can be considered well aligned and registered. We calculate the distances between simulation surfaces $d_{ij} =  \left\| {f}_i -{f}_j \right\|$, where $i, j = 1, 2, ... , 40$. Figure \ref{simu_before} shows the heat map and multidimensional scaling (MDS) plot of the distances between surfaces. First 20 surfaces are presented by blue dots and last 20 are presented by red dots. This figure illustrates that with surface registration, the surfaces with similar shapes that are on the same direction of principal shape component have relatively small distance, and the similar surfaces are correctly clustered into the same class. 

\fboxsep=4pt
\fboxrule=1pt

\begin{figure}[!ht]
		\centering
			\fbox{
			\subfloat[\large Heat Map]{
					  	 		\includegraphics[height=4.5cm]{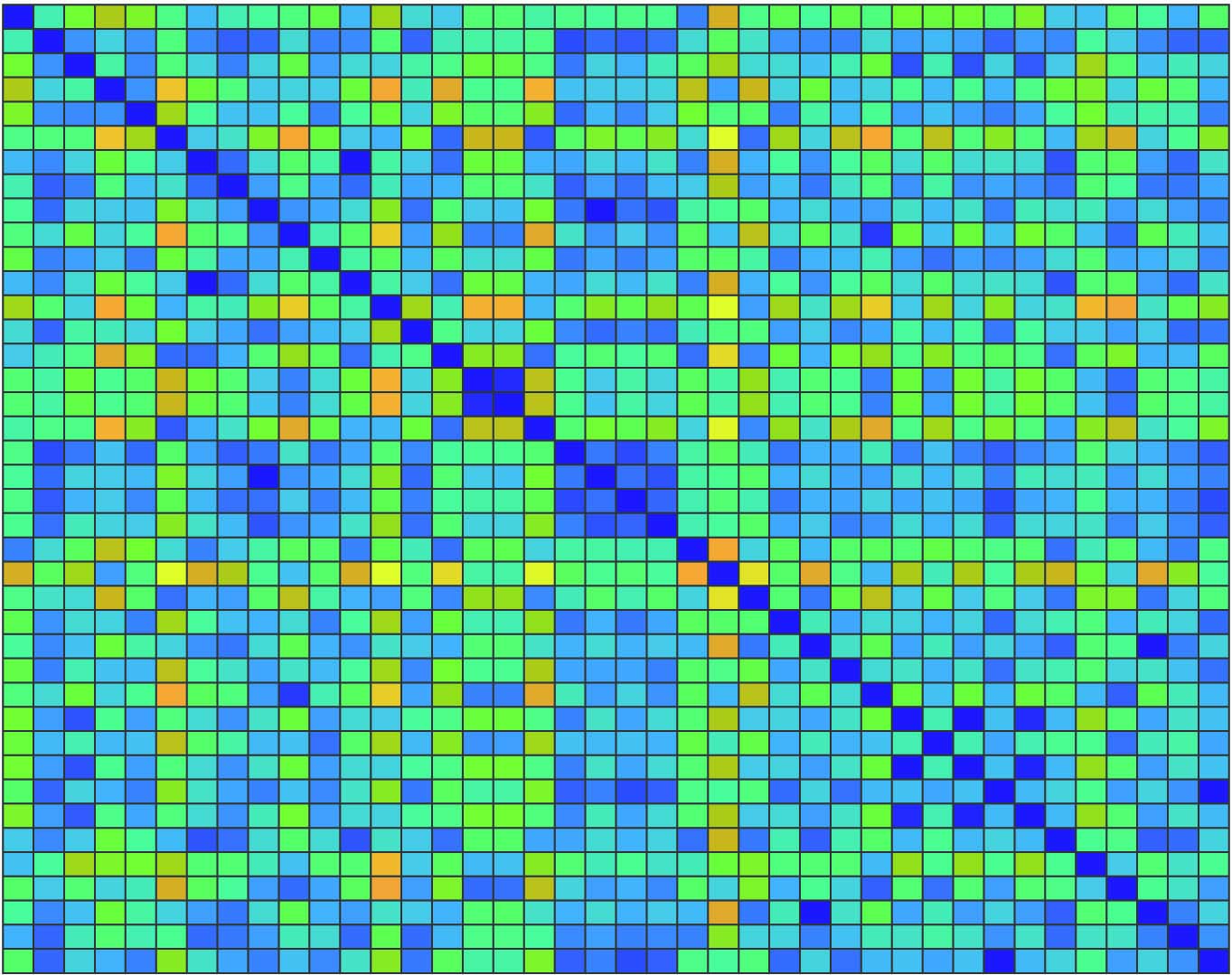}
					  			}}
		  \qquad
			\fbox{
			\subfloat[\large MDS Plot]{
					  	 		\includegraphics[height=4.5cm]{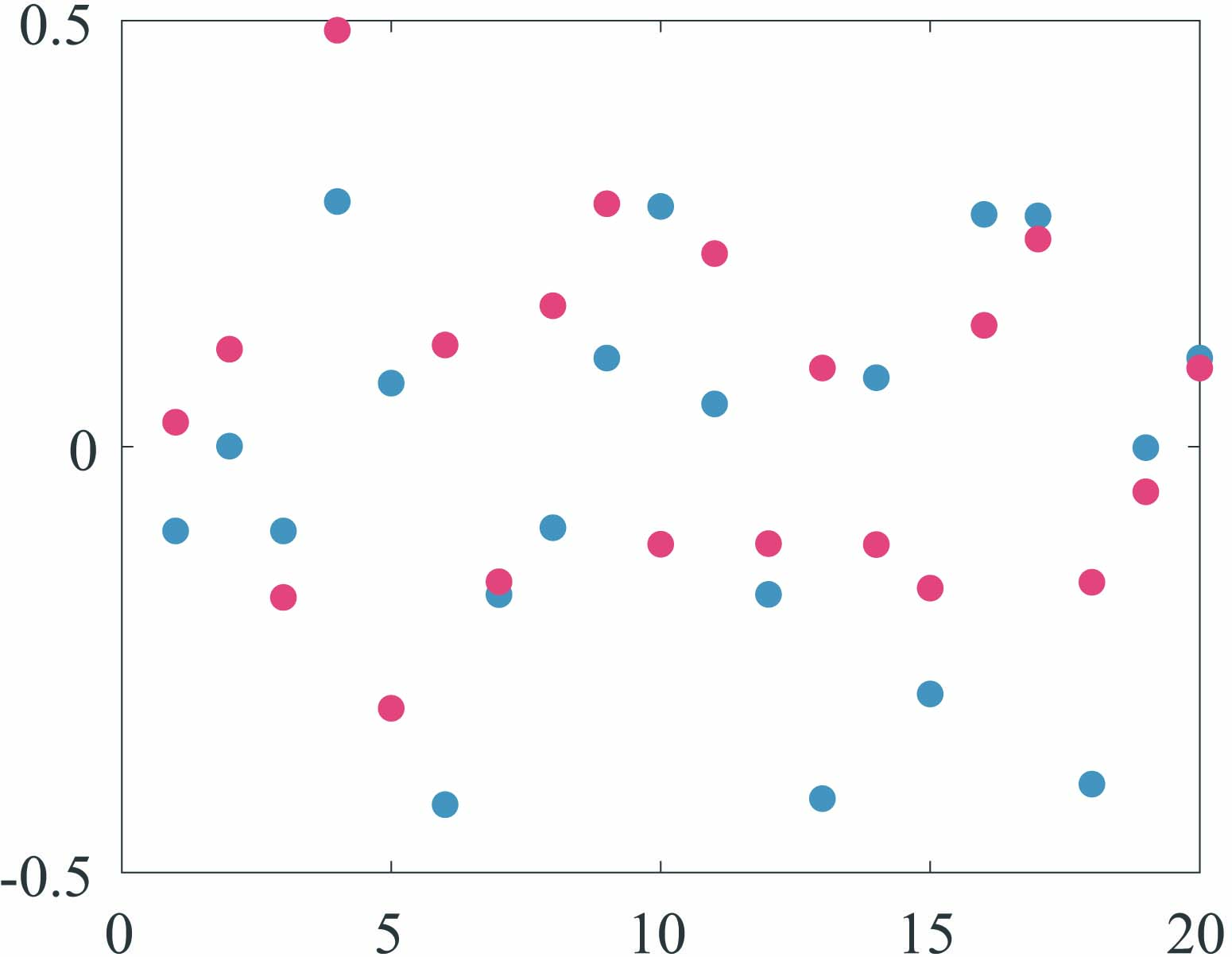}
					  			}}
   \caption{Heat map plot and MDS plot of distances between random parameterized surfaces (not well registered). Colors of the heat map indicates the relative distances between surfaces. Blue dots in MDS plot indicate the first 20 simulation surfaces and red dots indicate the last 20 simulation surfaces.}
   \label{simu_after}
\end{figure}

Next, we introduce random parameterization functions ${\gamma}_i \in \Gamma$ and apply ${\gamma}_i$'s to ${f}_i$'s to simulate randomly parameterized surfaces. 
For each $i$, the surface $\tilde{f}_i = {f}_i \circ {\gamma}_i$ has the same shape as ${f}_i$, but a different parameterization. The distances between unregistered surfaces are also calculated by $\tilde{d}_{ij} =  \left\| \tilde{f}_i - \tilde{f}_j \right\|$. Figure \ref{simu_after} shows the heat map and MDS plot of the distances between randomly parameterized surfaces. We see that distances between surfaces that have similar shapes are not smaller anymore and MDS plot shows that unregistered surfaces are not effectively clustered.

\fboxsep=4pt
\fboxrule=1pt

\subsubsection{Shape Analysis Tools: Geodesics, Mean, and PCA}
The framework developed so far allows for representing and registering anatomical surfaces and comparing shapes of these surfaces in a pairwise fashion using a proper shape metric. This metric is used to develop some additional statistical tools, which, in turn, lead to a compact way of representing shapes. These tools include finding geodesics between shapes, computing means of shapes of surfaces, and discovering principal modes of shape variation in a given set of shapes.
\\

\begin{figure}[!ht]
		\centering
		\resizebox{.9\columnwidth}{!}{
		\fbox{\begin{minipage}{\dimexpr\textwidth-2\fboxsep-2\fboxrule\relax}
		\centering
		\subfloat{
 		\includegraphics[width=.1\linewidth]{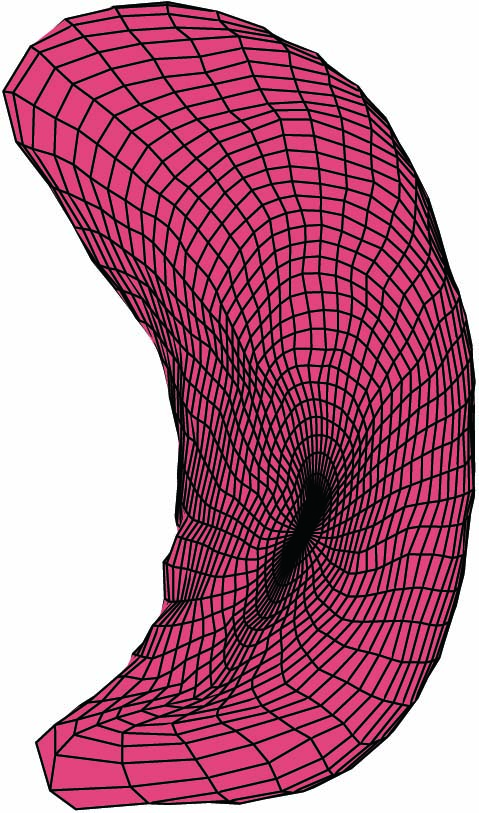}
 		\includegraphics[width=.1\linewidth]{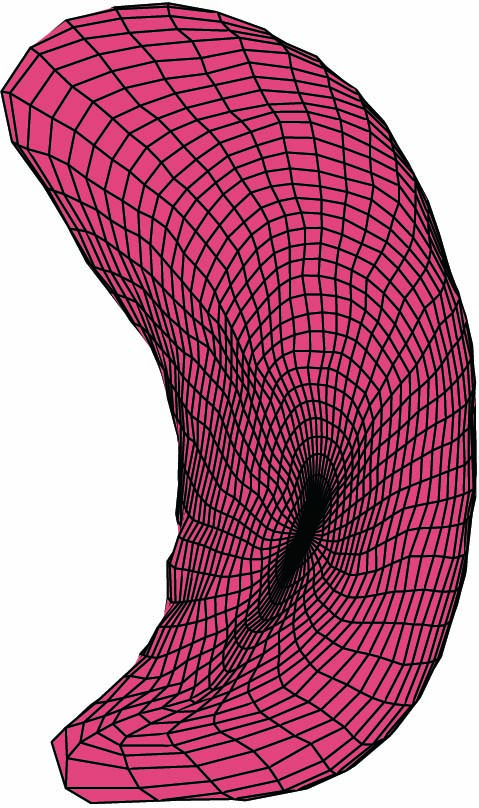}
 		\includegraphics[width=.1\linewidth]{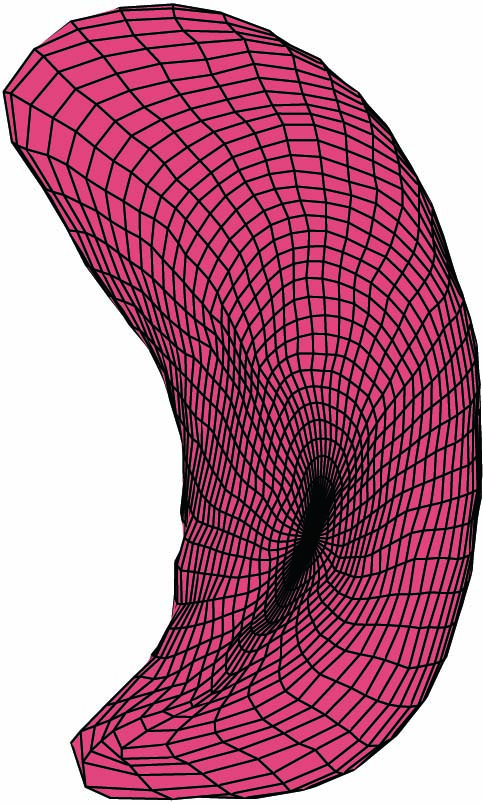}
 		\includegraphics[width=.1\linewidth]{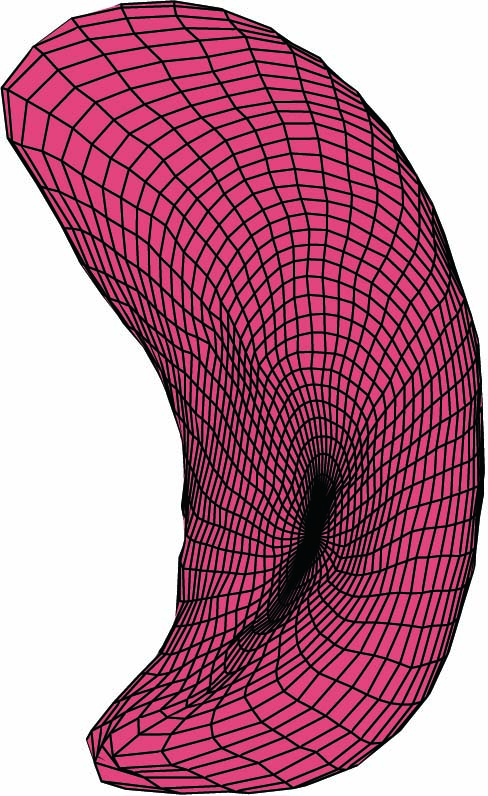}
 		\includegraphics[width=.1\linewidth]{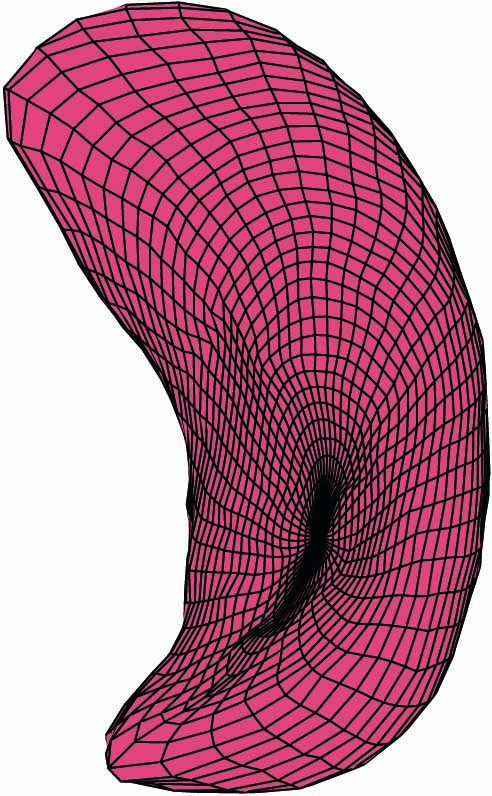}
 		\includegraphics[width=.1\linewidth]{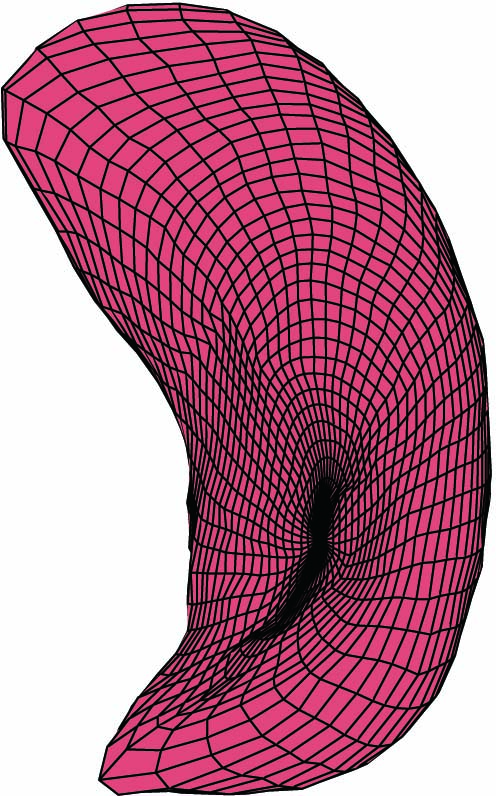}		
 		\includegraphics[width=.1\linewidth]{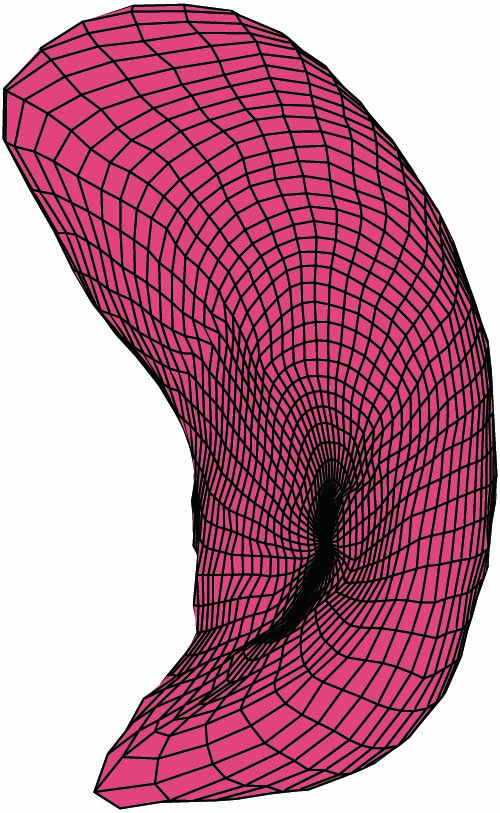}
 		\includegraphics[width=.1\linewidth]{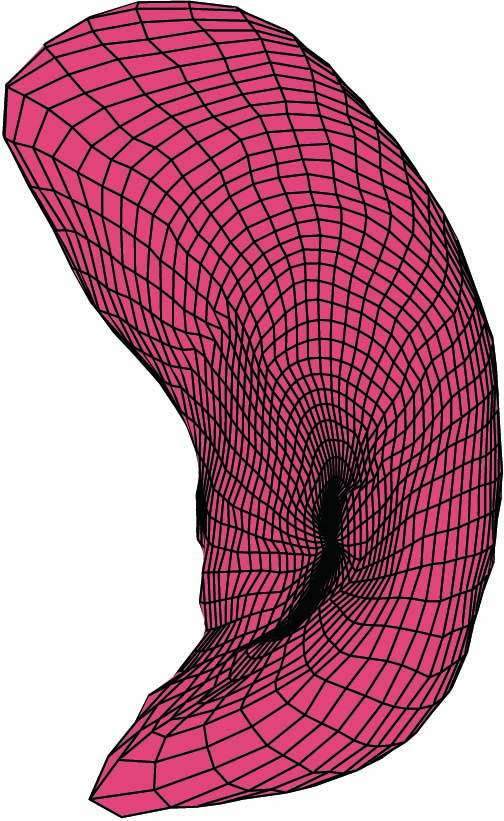}
		\includegraphics[width=.1\linewidth]{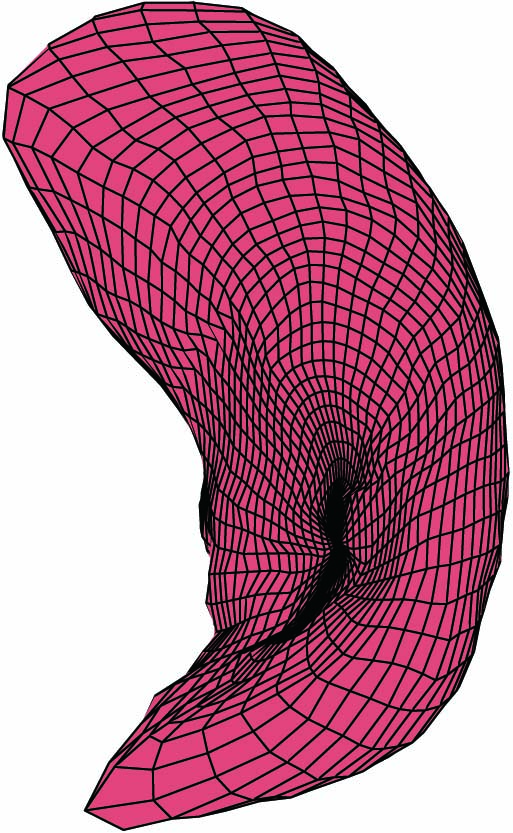}
		}
		\caption*{\Large $f_1 \xrightarrow{\hspace*{13.5cm}} f_2$}
		\end{minipage}}
		}
		\resizebox{.9\columnwidth}{!}{
		\fbox{\begin{minipage}{\dimexpr\textwidth-2\fboxsep-2\fboxrule\relax}
		\centering
		\subfloat{
 		\includegraphics[width=.1\linewidth]{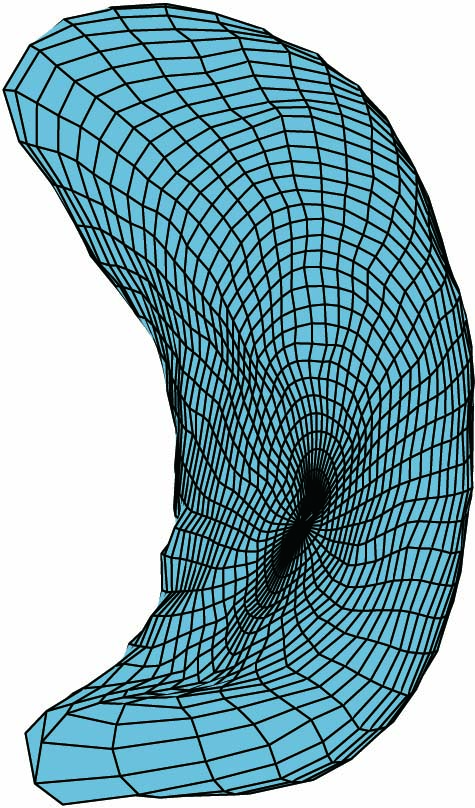}
 		\includegraphics[width=.1\linewidth]{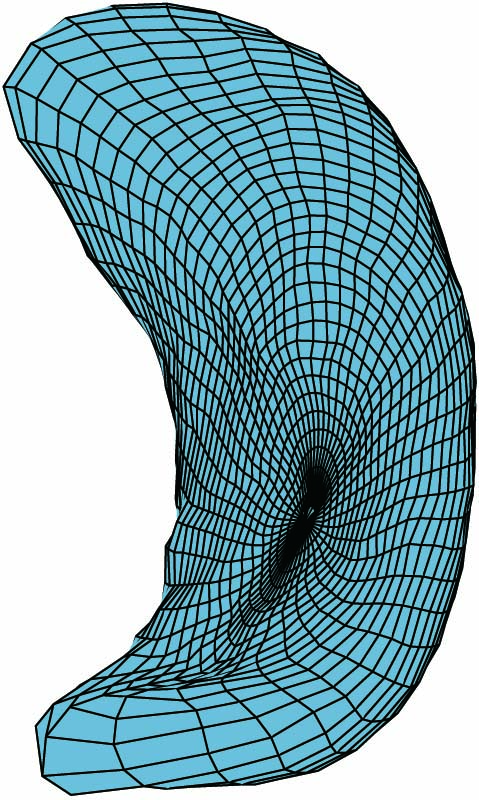}
 		\includegraphics[width=.1\linewidth]{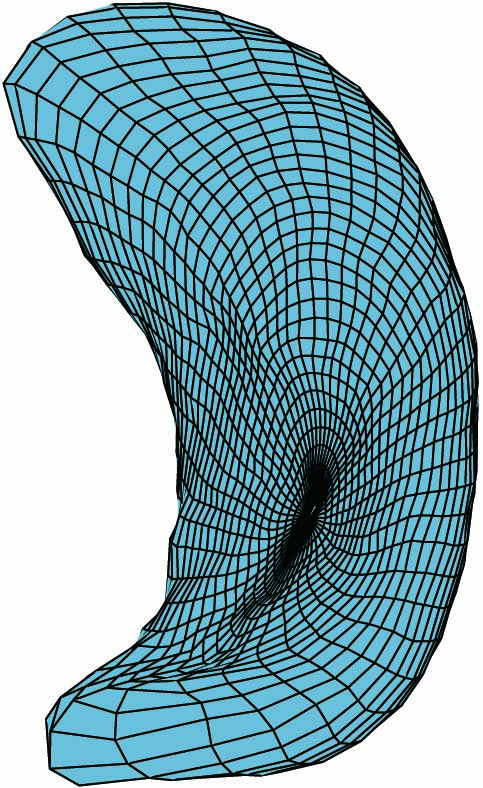}
 		\includegraphics[width=.1\linewidth]{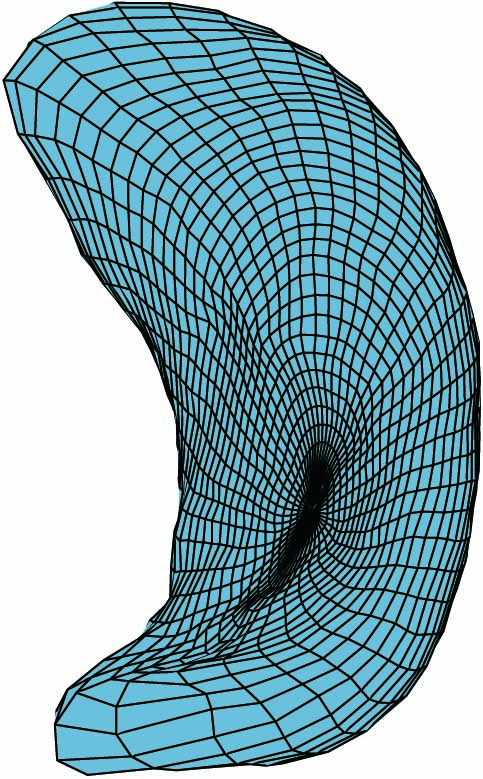}
 		\includegraphics[width=.1\linewidth]{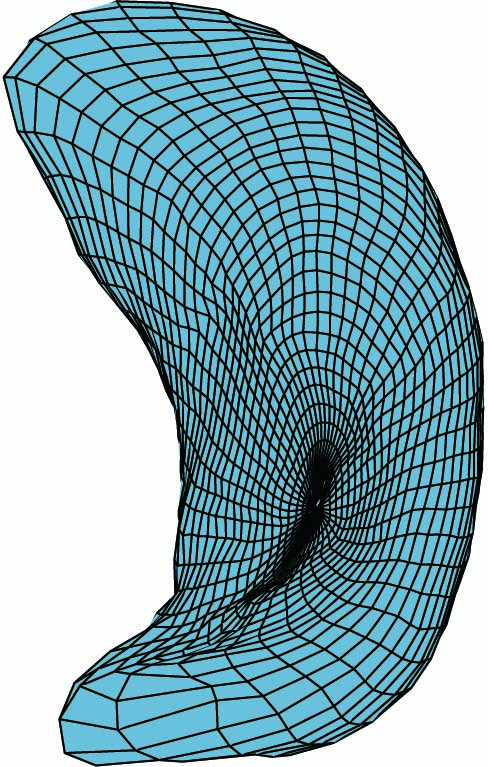}
		\includegraphics[width=.1\linewidth]{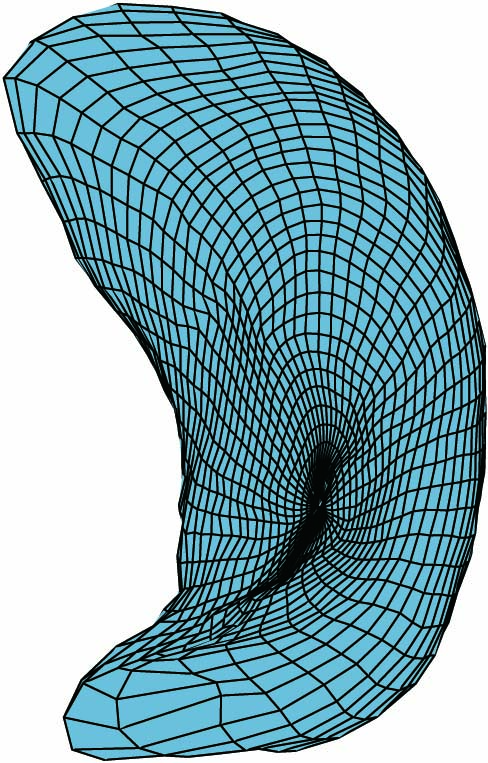}			
		\includegraphics[width=.1\linewidth]{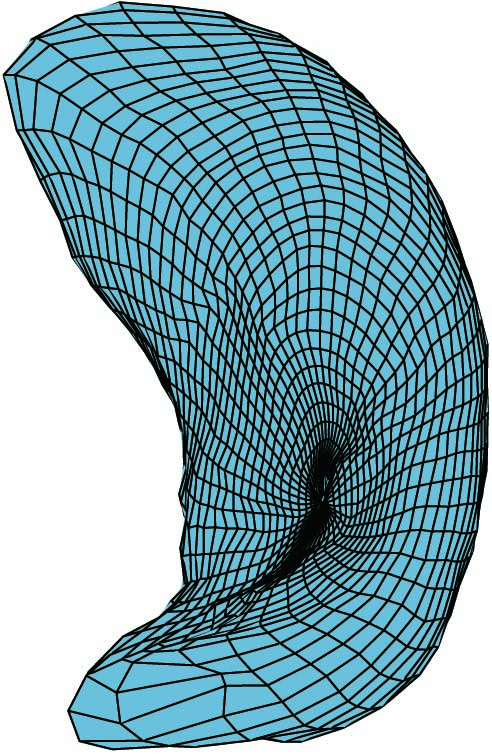}
 		\includegraphics[width=.1\linewidth]{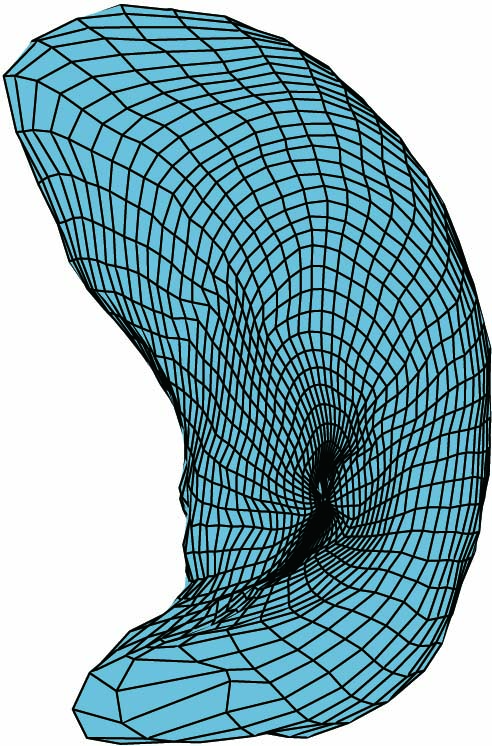}
 		\includegraphics[width=.1\linewidth]{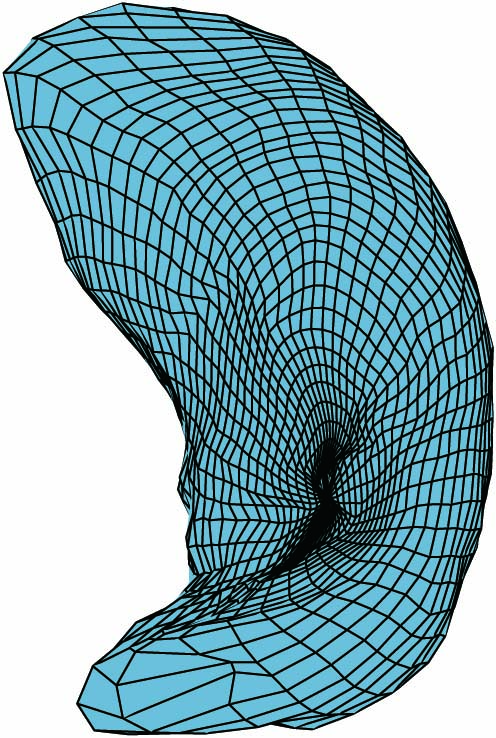}
		}
		\caption*{\Large $f_1 \xrightarrow{\hspace*{13.5cm}} f_2^*$}
		\end{minipage}}
		}
		\caption{Two examples of geodesic between left hippocampus surfaces. Upper row shows the geodesic between unregistered surfaces, and lower row shows the geodesic between surfaces that are elastic registered.}
		\label{Geodesic_example}
\end{figure}

\noindent {\bf Shape Geodesic}: 
Given two surfaces, $f_1$ and $f_2$, a geodesic between their shapes is a visualization of the optimal deformation from one to the other. Although there are more sophisticated ways to compute exact geodesics, we use a simple linear interpolation to approximate this deformation according to: 
\begin{equation} \label{Geodesic}
\alpha_\tau^*(s) = (1-\tau) f_1(s) + \tau f_2^*(s),\ \ s \in \s^2\ ,
\end{equation}
where $\tau \in [0,1]$ is the time index of the geodesic. At $\tau = 0$ we have $f_1$ and at $\tau=1$ we have $f_2^*$. Figure~\ref{Geodesic_example} shows two examples of geodesics, where the upper one is the geodesics between unregistered surfaces and the lower one is between elastic registered surfaces. Due to the misalignment of points between surfaces, the hippocampus's posterior end ''degenerates'' on the upper geodesic at the midway point. In contrast, the anatomical features of hippocampus surfaces are preserved with elastic registration, making the midway surfaces along the geodesic more interpretable.

\fboxsep=5pt
\fboxrule=2pt
Other example geodesics between surfaces of the same subcortical structure can be found in Supplementary Fig. 1. We provide geodesic paths also as GIF files in the Supplementary Materials.
\\

\noindent {\bf Mean Shape}: 
The next concept we introduce is the mean of a shape of shapes of surfaces $f_1, f_2, \dots, f_n$. 
We define an intrinsic mean, or the Karcher mean as the shape that $\mu$ that minimizes the sum of square of distances 
(under the shape metric) to the given shapes. That is: 
$$
\mu - \argmin_{f \in {\cal F}} \sum_{i=1}^n d_s(f, f_i)^2\ .
$$
We use an iterative algorithm (Algorithm~\ref{mean_algorithm}) to compute this mean shape. Here we start by selecting an arbitrary surface as the initial guess for $\mu$. Then, in each iteration, we register each $f_i$ with the current mean and compute the mean of these
registered $f_i$s. Once the algorithm has converged, we obtain the mean shape $\mu$. We outline these steps using Algorithm~\ref{mean_algorithm}. 
\\

\vspace{\baselineskip}
\begin{algorithm}[H]
\caption{Computing Karcher mean shape of surfaces}
\label{mean_algorithm}
\SetAlgoLined
\KwData{Surfaces $f_i \in \mathcal{F}$, $i=1,2,...,n$}
\KwResult{Karcher mean $\mu$ and registered surfaces $f_i^*$, $i=1,2,...,n$}
Find random $f_i$ as $\mu_0$ \;
\For{$j = 1$ to $5$}{
	\For{$i = 1$ to $n$}{
	$O_i^*,\gamma_i^* = \underset{O \in SO(3),\gamma \in \Gamma}{\arg\min}d_s(\mu_{j-1},f_i)$\;
	$f_i^*= O^*(f_i \circ \gamma^*)$\;
	}
	$\mu_i = \frac{1}{n}\sum_{i=1}^{n} f_i^*$\;
}
$O_i^*,\gamma_i^* = \underset{O \in SO(3),\gamma \in \Gamma}{\arg\min}d_s(\mu,f_i)$\;
$f_i^*= O^*(f_i \circ \gamma^*)$\;
\end{algorithm}	
\vspace{\baselineskip}

\noindent {\bf Shape PCA}: 
Next, we perform Principal Component Analysis (PCA) to capture essential shape variability in a given set of surfaces. We start by computing the covariance matrix $C$ for surfaces:
\begin{equation} \label{Covariance}
C = \sum_{i=1}^{n} V_i V_i^T,\ \ \ \mbox{where}\ \ V_i = f_i - \mu\ .
\end{equation}
By performing singular value decomposition (SVD) on the covariance matrix $C$, we obtain the left singular vectors as the columns of the unitary matrix $U$. These columns form the principal directions of shape variability the data: The first column is called the 1st principal component, the second column the 2nd principal component, etc. This decomposition also results in singular values that indicate the variance of the shape variability among each of the principal directions. 
\\

\noindent {\bf Low-Dimensional Shape Representations}: We use PCA to derive low-dimensional representations of shapes of objects for using in statistical models and regressions. 
During experiments we divide all surfaces (of a specific type, say hippocampus) randomly into a training group and test groups. and compute the principal components of shape variation using only the training surfaces. Then, we compute the principal score for all test surfaces. 
For a test surface $f_i \in \mathcal{F}$, $i=1,2,...,n$ and principal directions $U(:,d),\ d=1,2,...,n$, the principal score is computed by:
$z_{i,d} = \inner{f_i - \mu}{U(:,d)}$. 
Thus, a high-dimensional object $f_i$ is not represented by a $d$-dimensional vector $z_i \in \real^d$. 
It is important to note that this representation is invertible and we can reconstruct test surfaces according to:
$\hat{f}_i = \mu + \sum_{d=1}^{n} z_{i,d}U(:,d)$. 
We validate this representation by examining the difference between the reconstructed surface $\hat{f}_i$ and the original surface $f_i$.

\subsection{Regression Model Design}
Having obtained low-dimension mathematical representations of subcortical structures' shapes, we can study their roles in statistical models for diagnosing PTSD. Specifically, 
we train ten linear models to test the significance of shape differences between subcortical surfaces in different childhood traumatic experience levels and PTSD symptom scales. In model 1 and model 5, we include the interactions between shape (for $d= 5$ principal scores) and confounding variables (age and BDI). We design models 3, 4, 7, and 8 to test the interpretation power of shape in PTSD and experience levels. We use models 9 and 10 that include the intracranial volume (ICV) as a control covariate. Table \ref{Model design} shows predictors and responses for eight different models. To capture the shape differences while minimizing noise, we take only the first 15 principal scores (PS) for each surface. These represent the 15 most dominant modes of shape variation to train the model.

\begin{table}[!ht]\label{Model design}
		\begin{center}
				\caption{Model Design}
				\label{Model design}
				\begin{tabular}{c|c|c}
				\textbf{Model} & \textbf{Predictors} & \textbf{Responses}\\
				\hline
				1 & Age + BDI + PS + Interactions & PSS\\
				2 & Age + BDI + PS & PSS\\
				3 & Age + BDI & PSS\\
				4 & PS & PSS\\
				\hline
				5 & Age + BDI + PS + Interactions & CTQTOT\\
				6 & Age + BDI + PS & CTQTOT\\
				7 & Age + BDI & CTQTOT\\
				8 & PS & CTQTOT\\
				\hline
				9 & Age+ BDI + ICV + PS + Interactions & PSS\\
				10 & Age+ BDI + ICV + PS + Interactions & CTQTOT\\
				\end{tabular}
		\end{center}
\end{table}

In these models, we select the most significant predictors using bidirectional stepwise regression. 

\subsection{Comparison with Vertex-wise Analysis}
To verify the effectiveness and ability of elastic shape analysis in identifying shape differences attributed to PTSD disease, we compare it with vertex-wise analysis, a widely applied shape analysis method in neuroimaging. We perform this comparison by applying a similar pipeline to vertex-wise surfaces, as our elastic approach, and comparing the results. 

We start by representing surfaces using sets of vertices or point clouds. Next, we apply state-of-art point cloud registration algorithms, iterative closest point \citep{besl1992method} (ICP), to register individual surfaces. This step is analogous to the elastic registration step. After registration, we compute mean, covariance and PCA, in the same way as elastic shape analysis.
We compare the two methods -- elastic shape analysis and vertex-wise analysis -- using (i) average inter- and intra-class total point-wise distances; and (ii) regression model results with the same model design following Table~\ref{Model design}, just with different principal scores of each analysis method. Here we define PTSD diagnosis, a binary (0 and 1) label included in the original data, as the class label for a subject. Although this PTSD diagnosis data is obtained using thresholding and cannot be considered the original medical diagnosis, it is still acceptable to use it as a label for comparing methods. We compute average inter- and intra-class total point-wise distance according to:
\begin{equation} \label{inter_distance}
d_{inter} = \frac
{\sum\limits_{i=1}^{90} \hspace{0.2cm} \sum\limits_{j=1,j > i, l_j \ne l_i}^{90} \hspace{0.2cm} \sum\limits_{k=1}^{K} \sqrt{(x_{j,k}-x_{i,k})^2+(y_{j,k}-y_{i,k})^2+(z_{j,k}-z_{i,k})^2}}
{\#\{(i,j)|i = 1,...,90, j = 1,...,90, j > i, l_j \ne l_i\}},
\end{equation}

\begin{equation} \label{intra_distance}
d_{intra} = \frac
{\sum\limits_{i=1}^{90} \hspace{0.2cm} \sum\limits_{j=1,j > i, l_j = l_i}^{90} \hspace{0.2cm} \sum\limits_{k=1}^{K} \sqrt{(x_{j,k}-x_{i,k})^2+(y_{j,k}-y_{i,k})^2+(z_{j,k}-z_{i,k})^2}}
{\#\{(i,j)|i = 1,...,90, j = 1,...,90, j > i, l_j = l_i\}},
\end{equation}
where $(x_{i,k},y_{i,k},z_{i,k})$ and $(x_{j,k},y_{j,k},z_{j,k})$ are $x$, $y$ and $z$ coordinates of the $k$\textit{th} point (with $K$ points in total) of the $i$\textit{th} and $j$\textit{th} surfaces. Recall that we have 90 surfaces in total for each subcortical structure. Also, here $l_i$ and $l_j$ are the surface class labels (0 or 1) for the $i$\textit{th} and $j$\textit{th} surfaces separately. We compute both average inter- and intra-class total point-wise distances for the two methods after their respective registrations.

\section{Results}
This paper focuses on discovering the impact of traumatic experiences and PTSD disease on three subcortical brain structures surface: left hippocampus, left amygdala, and left putamen. Thus, we present the shape analysis results for three groups of surfaces and their connections with traumas and PTSD disease.

\subsection{Shape Statistics of Surfaces}
First, we show the shape statistics, including geodesics, Karcher mean, principal components and reconstruction accuracy for three subcortical structures separately.

\subsubsection{Karcher Mean}
\begin{figure}[!ht]
			\centering
			\resizebox{.9\linewidth}{!}{           
			\fbox{
			\hfill
    		\subfloat[\large Left Hippocampus]{
    				\begin{tikzpicture}
    							\node at (1.2,3.5) {\includegraphics[height=4.3cm]{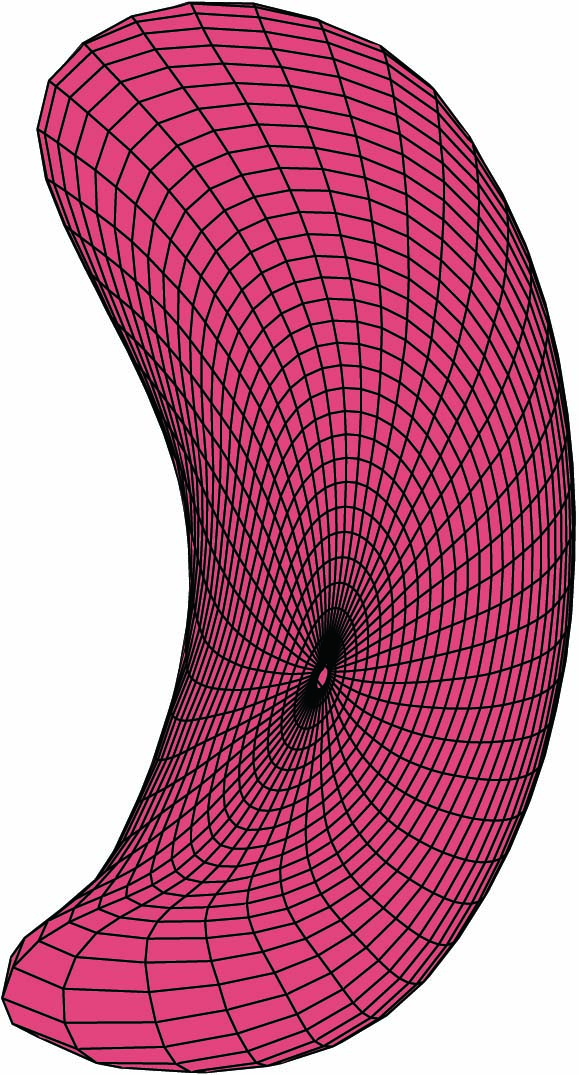}};
							 	\node at (0,0) {\includegraphics[height=2.5cm]{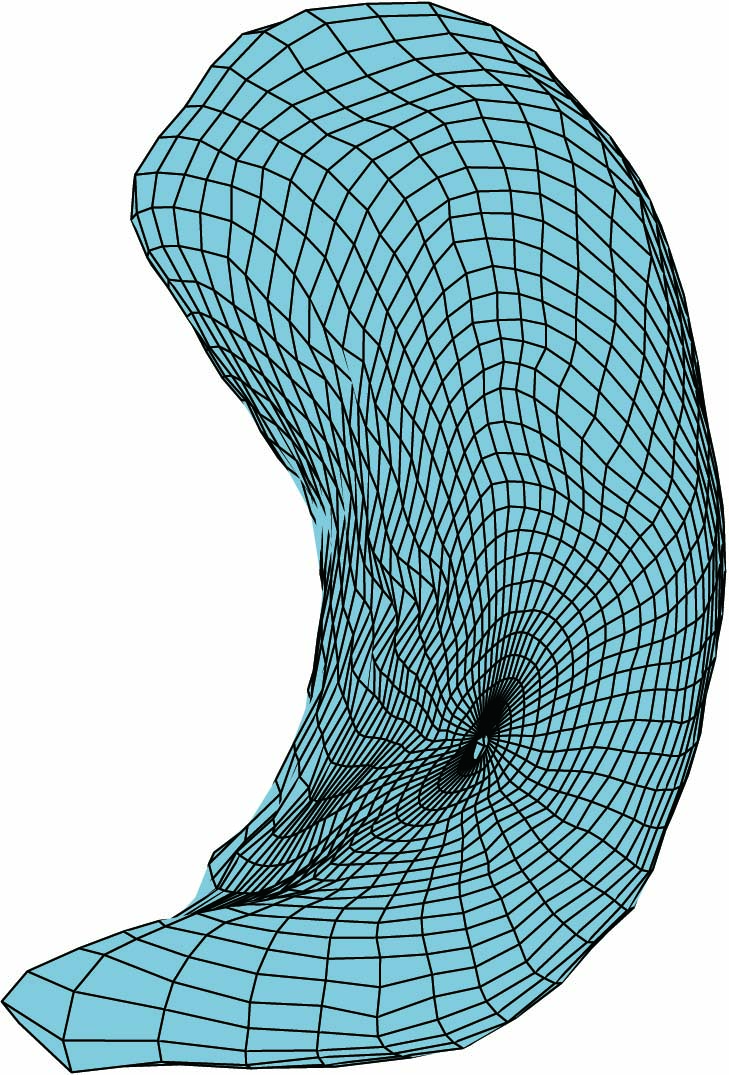}};
							 	\node at (0,7) {\includegraphics[height=2.5cm]{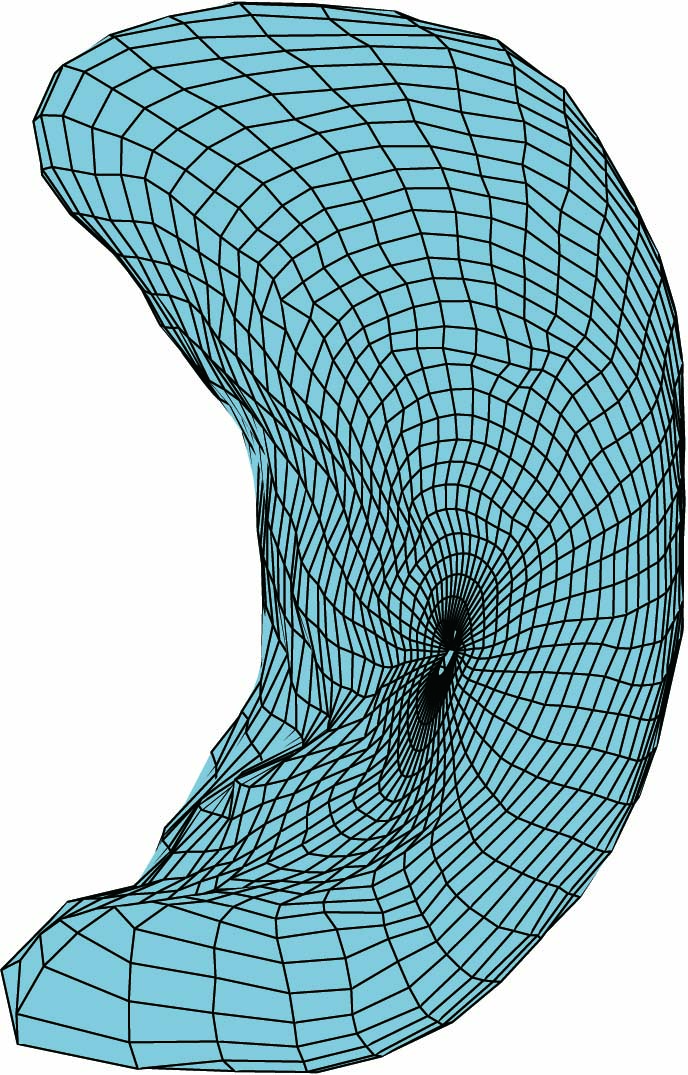}};
							 	\node at (2.5,7) {\includegraphics[height=2.5cm]{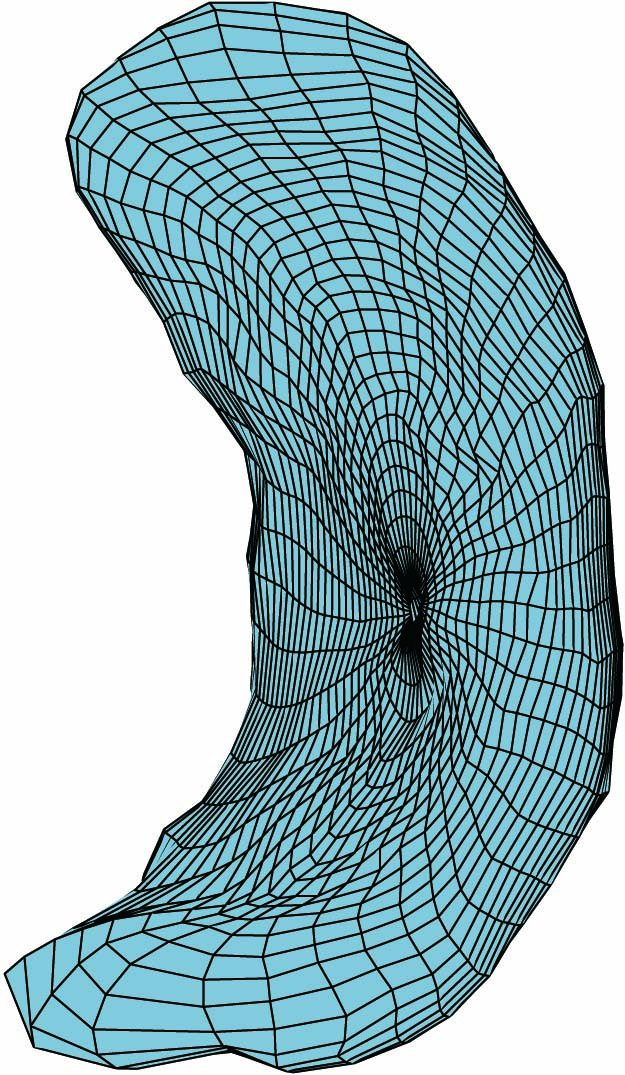}};
							 	\node at (2.5,0) {\includegraphics[height=2.5cm]{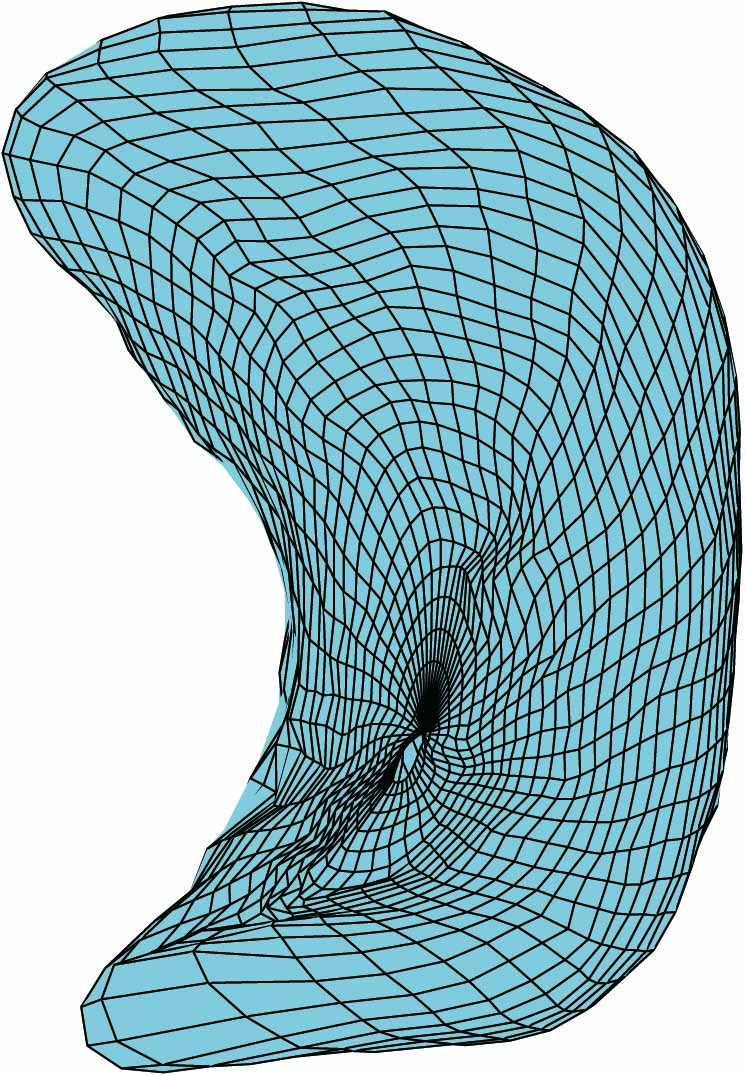}};
						\end{tikzpicture}		
			}}
			\fbox{
			\hfill
			\subfloat[\large Left Amygdala]{
    				\begin{tikzpicture}
    				 			\node at (1.2,3.5) {\includegraphics[height=4.3cm]{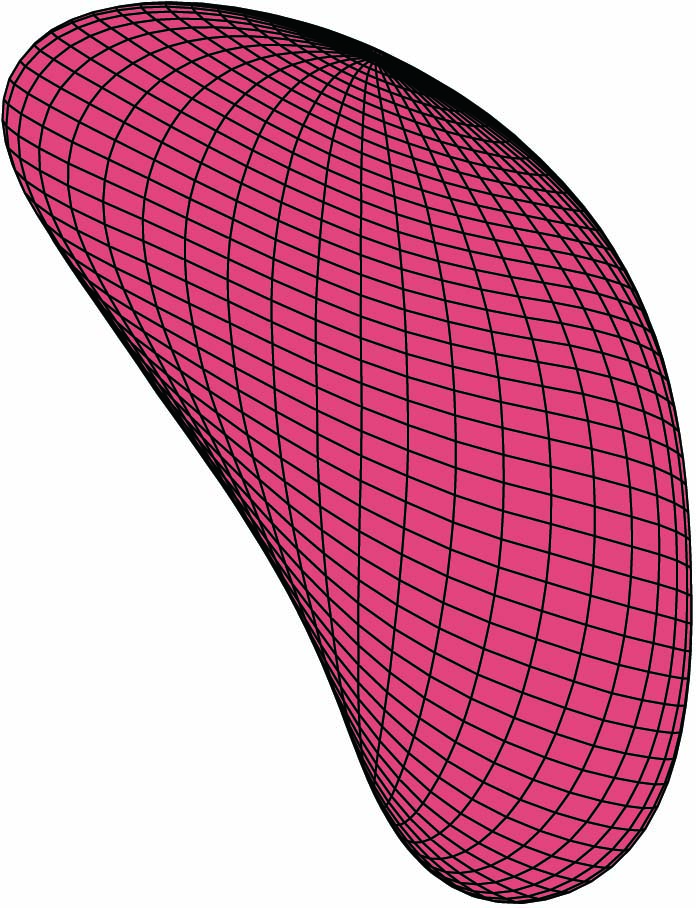}};
							 	\node at (0,0) {\includegraphics[height=2.5cm]{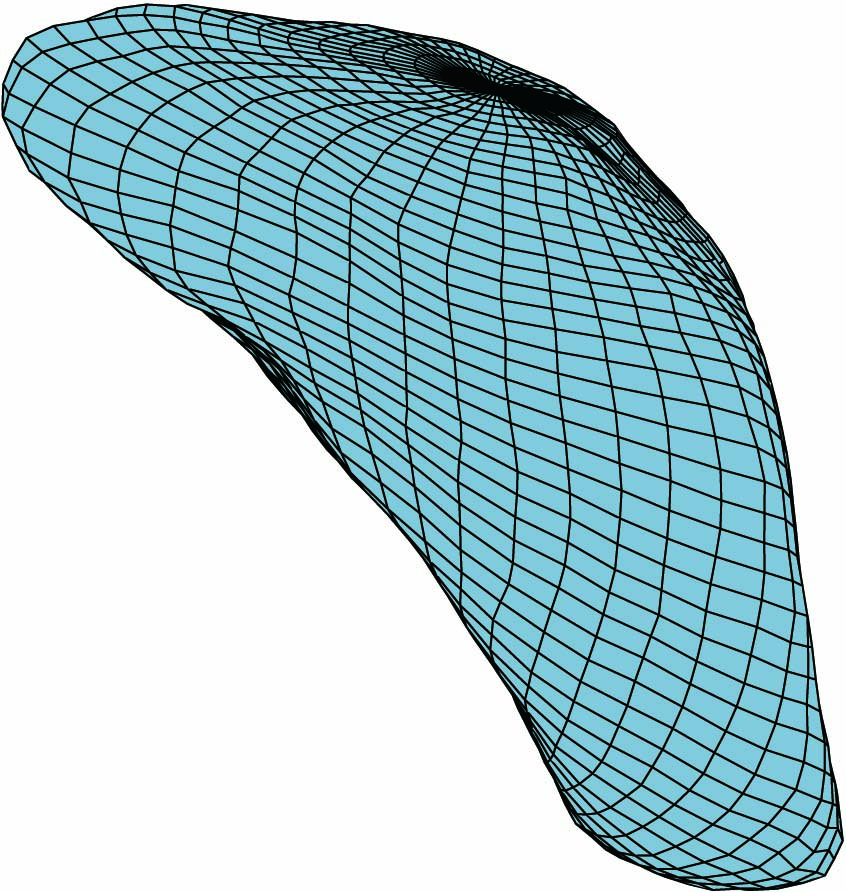}};
							 	\node at (0,7) {\includegraphics[height=2.5cm]{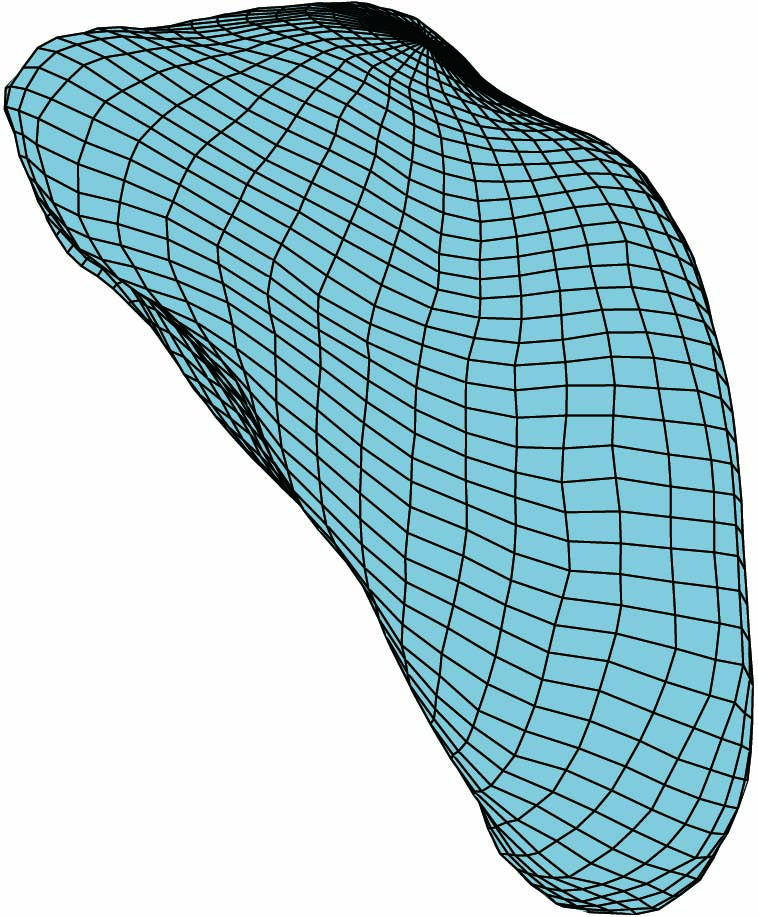}};
							 	\node at (2.5,7) {\includegraphics[height=2.5cm]{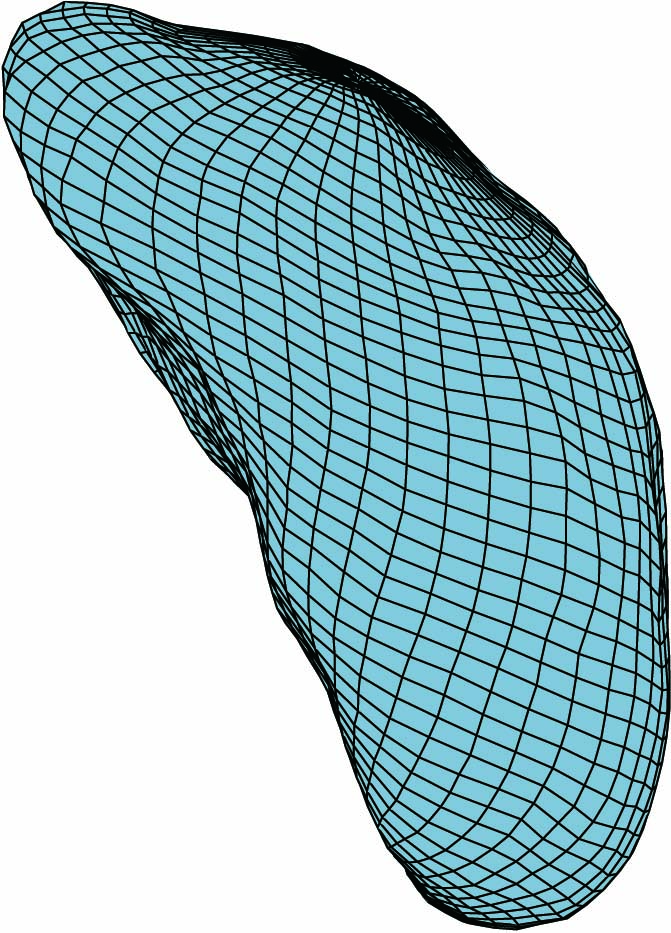}};
							 	\node at (2.5,0) {\includegraphics[height=2.5cm]{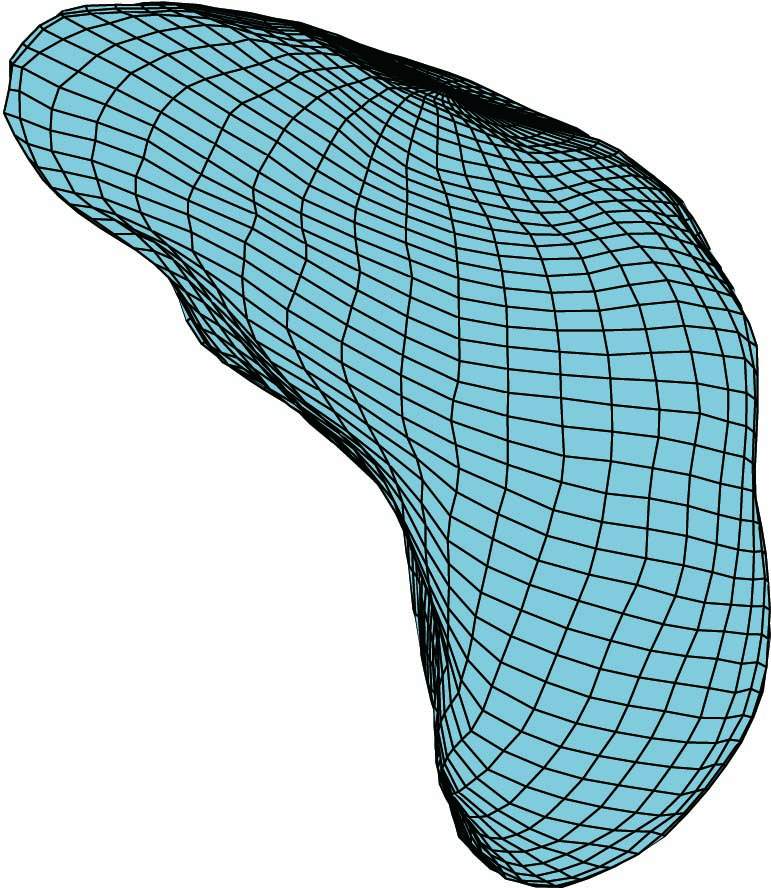}};
						\end{tikzpicture}		
			}}
			\fbox{
			\hfill
			\subfloat[\large Left Putamen]{
    				\begin{tikzpicture}
    				 			\node at (1.2,3.5) {\includegraphics[height=4.3cm]{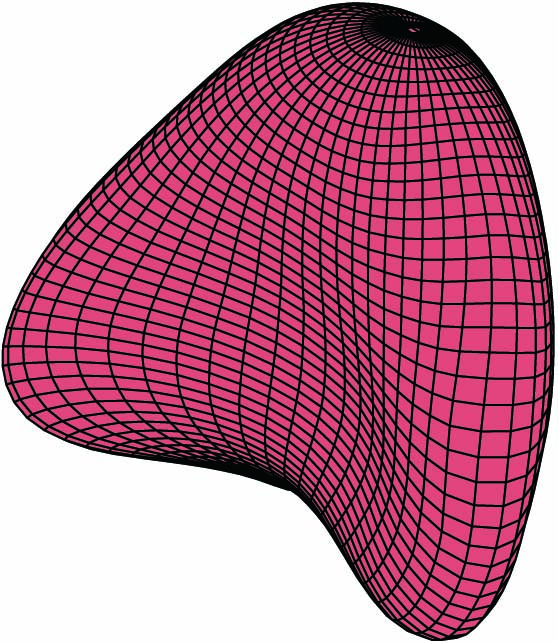}};
							 	\node at (0,0) {\includegraphics[height=2.5cm]{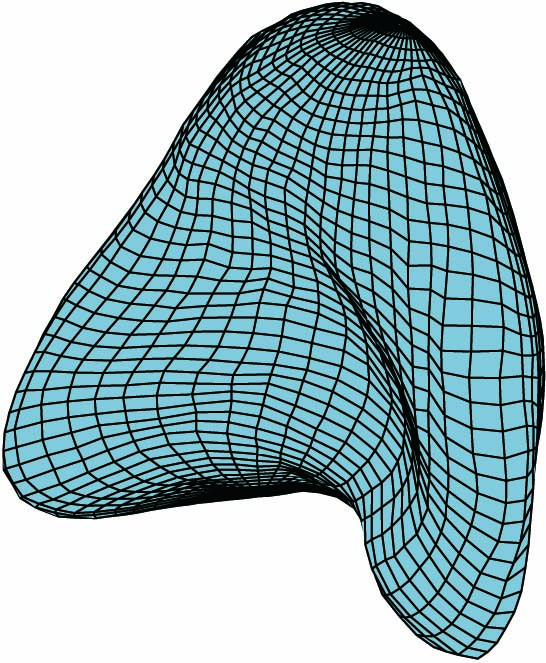}};
							 	\node at (0,7) {\includegraphics[height=2.5cm]{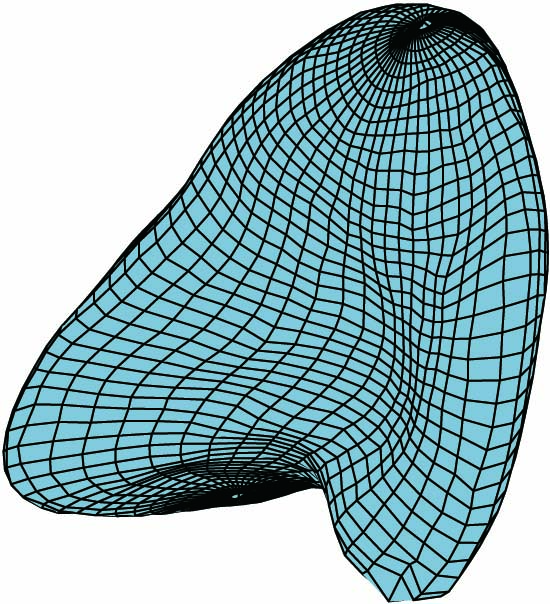}};
							 	\node at (2.5,7) {\includegraphics[height=2.5cm]{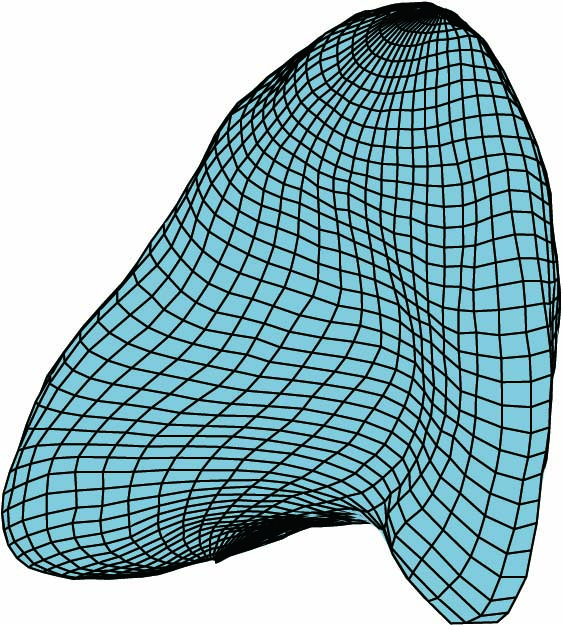}};
							 	\node at (2.5,0) {\includegraphics[height=2.5cm]{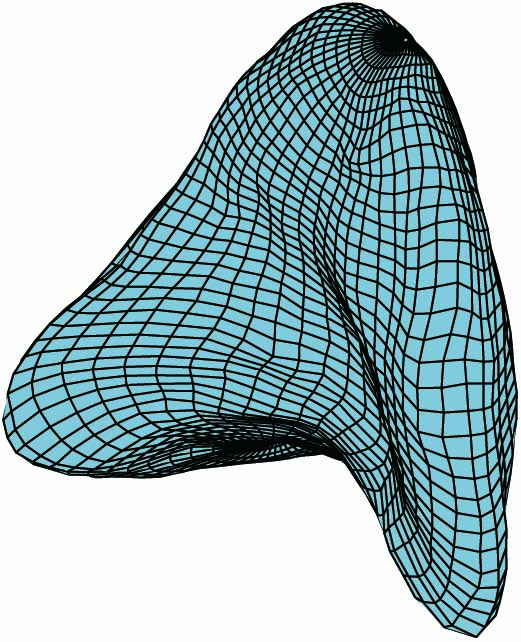}};
						\end{tikzpicture}		
			}}}
    \caption{Karcher mean surfaces and sample individual surfaces of three subcortical structures. Surrounding blue ones are randomly selected sample individual surfaces, and the middle red ones are Karcher mean surfaces computed using Algorithm~\ref{mean_algorithm}.}
    \label{Mean_surface}
\end{figure}

First, we present outputs of statistical averaging of shapes in the form of the Karcher mean surfaces $\mu$ as computed with Algorithm~\ref{mean_algorithm}. Figure \ref{Mean_surface} shows some registered individual surfaces $f_i^*$ (drawn around the mean) and their Karcher mean surfaces $\mu$ (drawn in the middle). Column (a) illustrates the shapes for the left hippocampus, and column (b), (c) illustrate those for the left amygdala and left putamen. We see that the Karcher mean surfaces capture salient anatomical shape features among the groups while reducing the individual noise.

\fboxsep=1.5pt
\fboxrule=1pt

\begin{figure}[!ht]
		\centering
			\fbox{
    				\subfloat[\large Left Hippocampus]{
					  	 		\includegraphics[height=3.5cm]{Mean_hipp.jpg}
					  	 		\includegraphics[height=3.5cm]{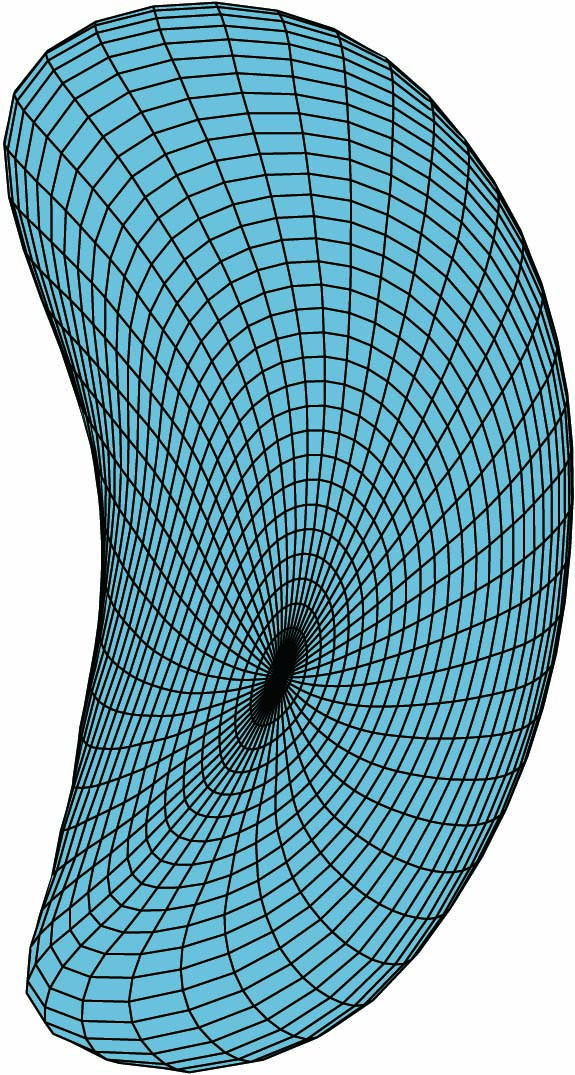}
					 }}
			\fbox{
    				\subfloat[\large Left Amygdala]{
					  	 		\includegraphics[height=3.5cm]{Mean_amyg.jpg}
					  	 		\includegraphics[height=3.5cm]{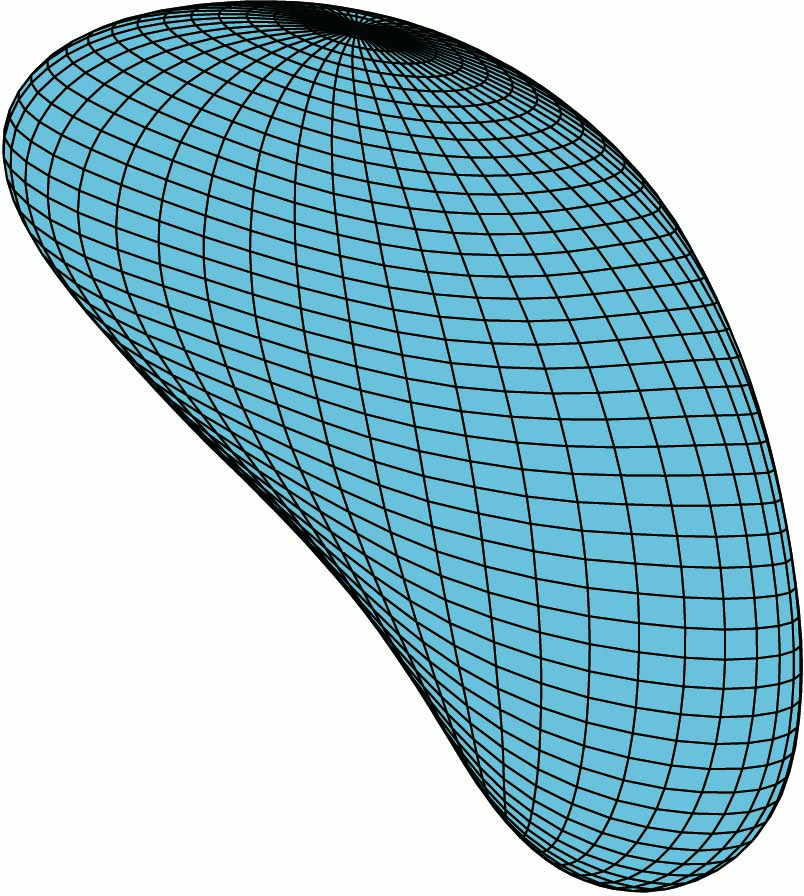}
					 }}
		\hfill
		\fbox{
    				\subfloat[\large Left Putamen]{
					  	 		\includegraphics[height=3.5cm]{Mean_puta.jpg}
					  	 		\includegraphics[height=3.5cm]{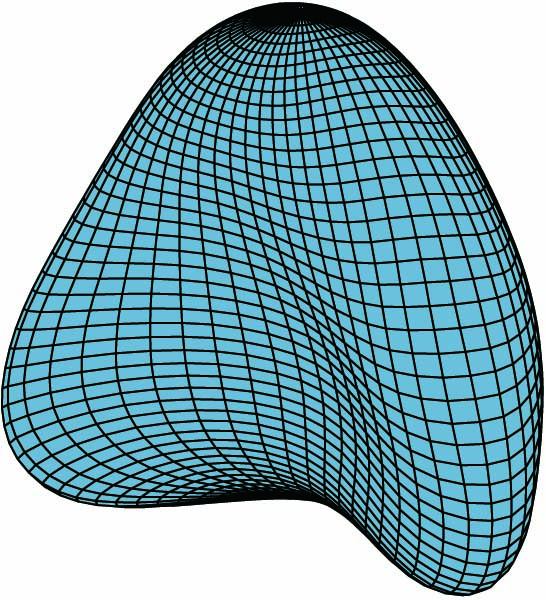}
					 }}
   \caption{Comparison between mean surfaces of three subcortical structures computed with and without elastic registration. Red ones are mean surfaces computed using Algorithm~\ref{mean_algorithm} with surface registration, while blue ones are computed without surface registration.}
   \label{Mean_compare}
\end{figure}

Figure~\ref{Mean_compare} compares shapes of mean surface computed with and without elastic registration. The left red shapes are the elastic mean surfaces computed with Algorithm~\ref{mean_algorithm}, and the right blue ones are the mean surfaces computed without surface registration. We can see that elastic mean captures more anatomical shape features, especially at the corner parts of subcortical structures. On the contrary, some shape information is ''averaged out'' when computing the mean surface without surface registration. For example, the posterior end of the hippocampus surfaces blurs when computing the mean this way.  

\fboxsep=3pt
\fboxrule=1pt

\subsubsection{Principal Components}
Using the mean surfaces $\mu$, we can now compute covariance matrix and perform PCA using Eqn.~\ref{Covariance} and SVD. We present the PCA results for three subcortical structures separately. 

\begin{figure}[!ht]
		\centering
		\resizebox{.9\linewidth}{!}{
		\fbox{\begin{minipage}{\dimexpr\textwidth-2\fboxsep-2\fboxrule\relax}
		\subfloat[\Large 1st Principal Component of Left Hippocampus]{
		\qquad
 		\includegraphics[height = 3.5cm]{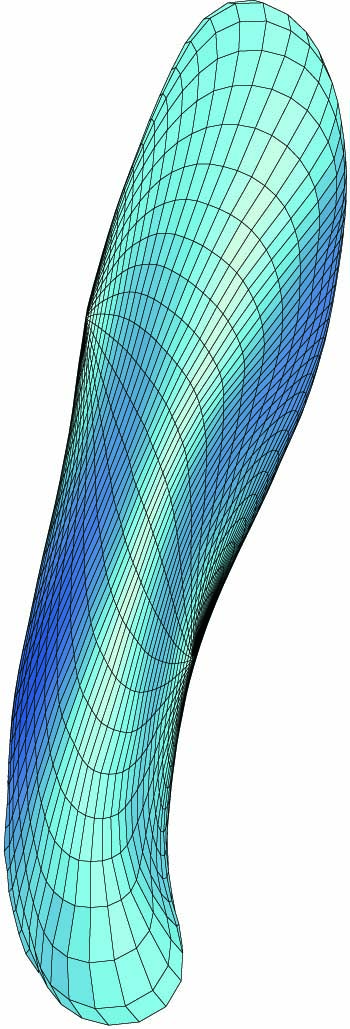}\hspace{1.8cm} 
 		\includegraphics[height = 3.5cm]{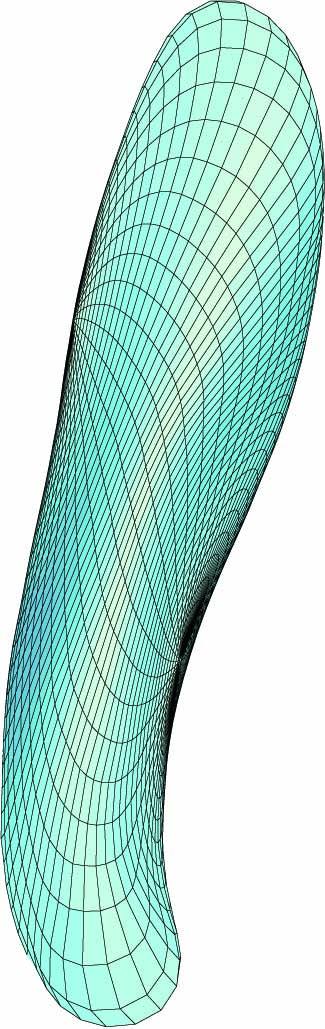}\hspace{1.8cm} 
 		\includegraphics[height = 3.5cm]{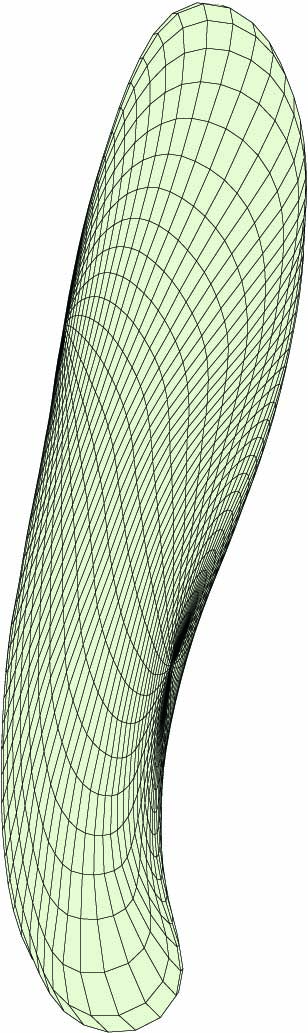}\hspace{1.8cm} 
 		\includegraphics[height = 3.5cm]{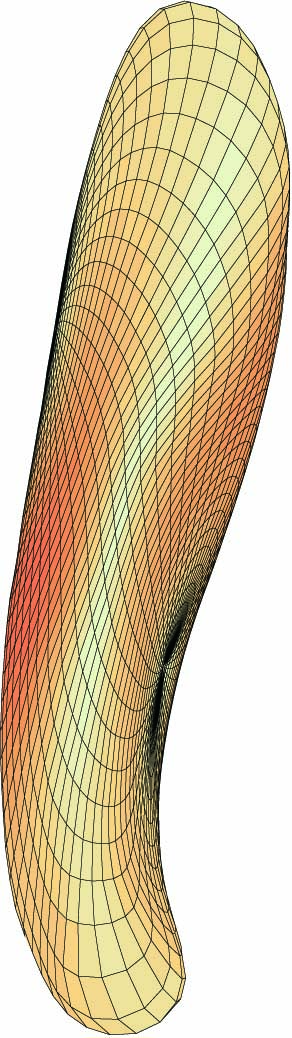}\hspace{1.8cm} 
 		\includegraphics[height = 3.5cm]{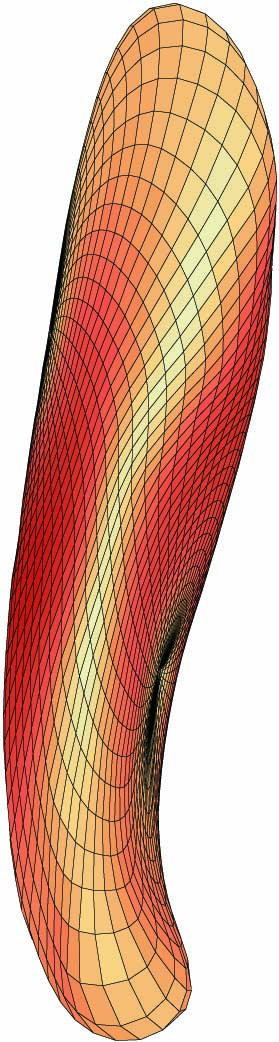}
		\hspace{1.2cm}
 		\includegraphics[height = 3cm]{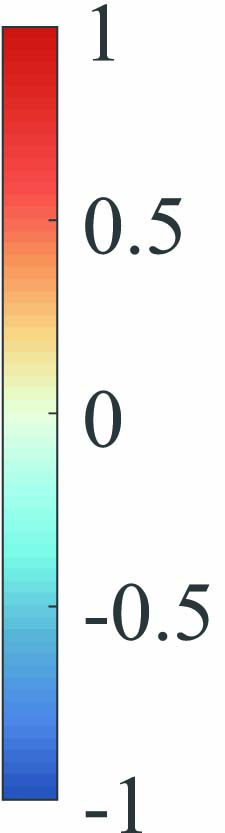}
		}
		\caption*{\Large $\mu-\sigma \rightarrow \mu+\sigma$}
		\end{minipage}}}

		\resizebox{.9\linewidth}{!}{
		\fbox{\begin{minipage}{\dimexpr\textwidth-2\fboxsep-2\fboxrule\relax}
		\subfloat[\Large 1st Principal Component of Left Amygdala]{
 		\includegraphics[width = .14\linewidth]{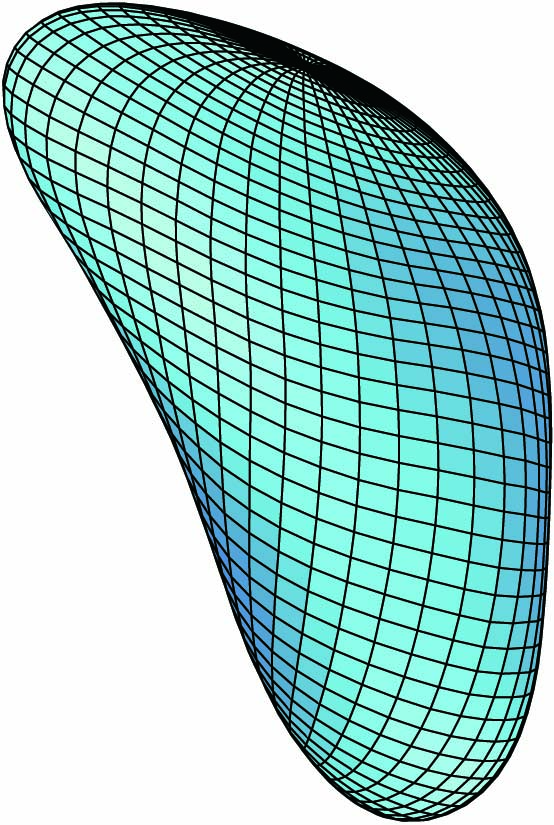}\qquad
 		\includegraphics[width = .14\linewidth]{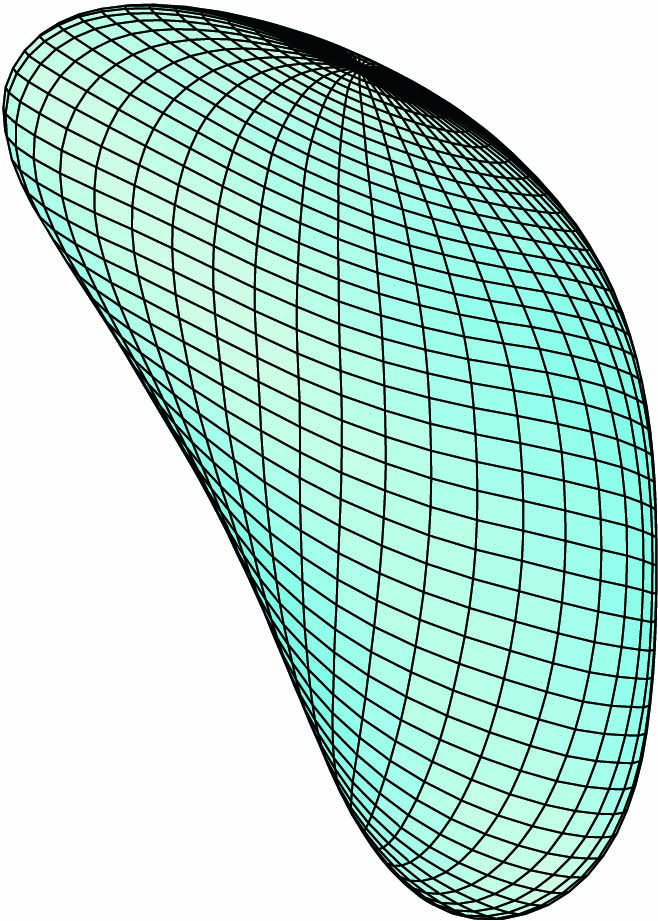}\qquad
 		\includegraphics[width = .14\linewidth]{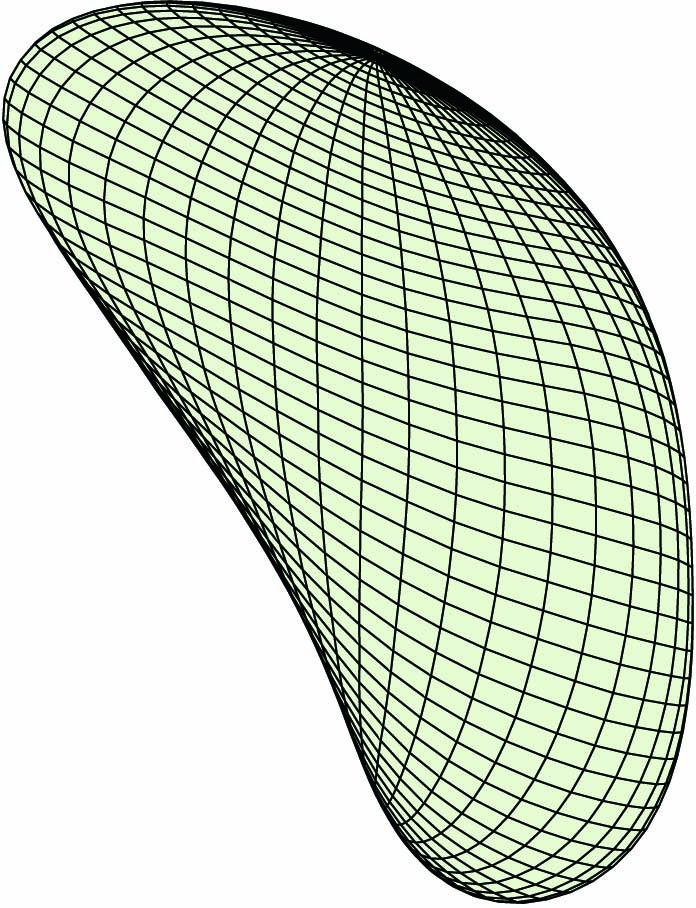}\qquad
 		\includegraphics[width = .14\linewidth]{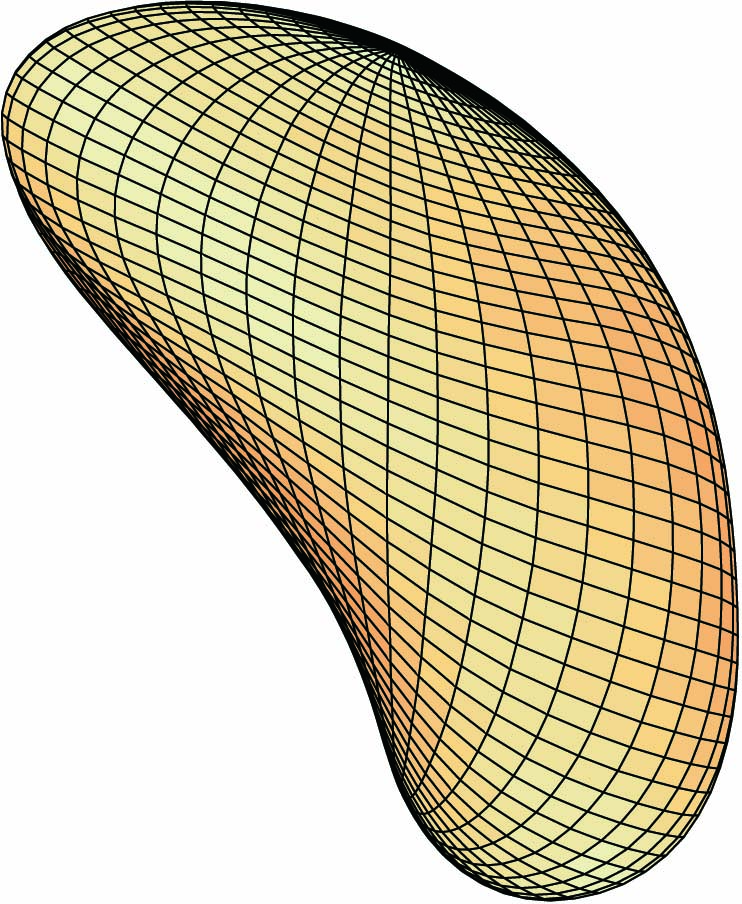}\qquad
 		\includegraphics[width = .14\linewidth]{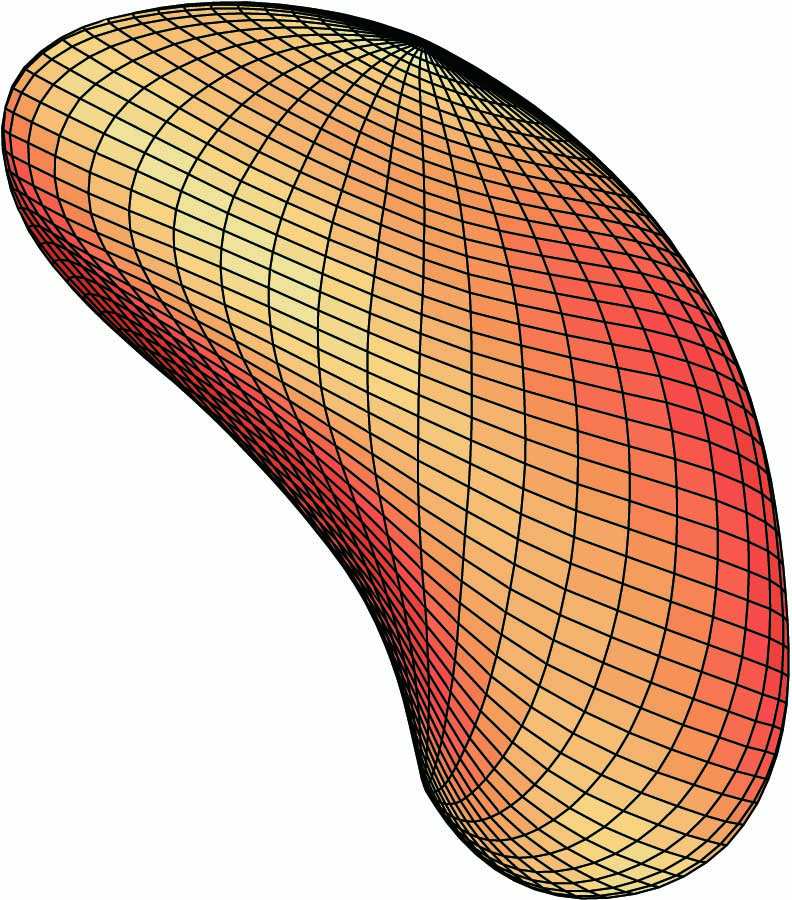}
		\qquad
 		\includegraphics[height = 3cm]{Colorbar2-1-1.jpg}
		}
		\caption*{\Large $\mu-\sigma \rightarrow \mu+\sigma$}
		\end{minipage}}}

		\resizebox{.9\linewidth}{!}{
		\fbox{\begin{minipage}{\dimexpr\textwidth-2\fboxsep-2\fboxrule\relax}
		\subfloat[\Large 1st Principal Component of Left Putamen]{
 		\includegraphics[width = .14\linewidth]{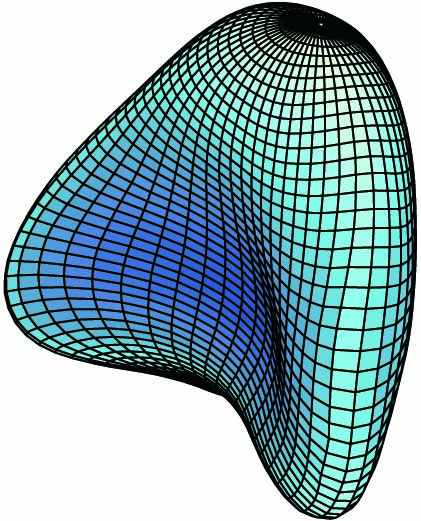}\qquad
 		\includegraphics[width = .14\linewidth]{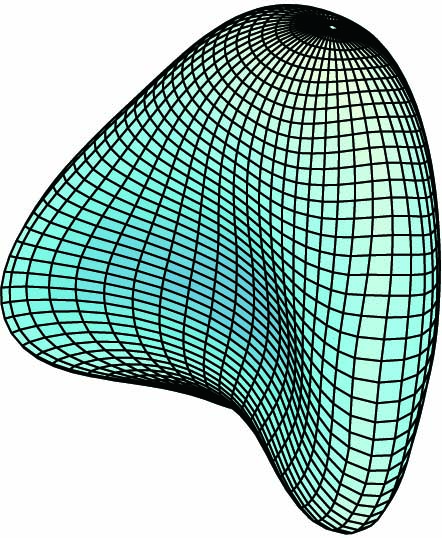}\qquad
 		\includegraphics[width = .14\linewidth]{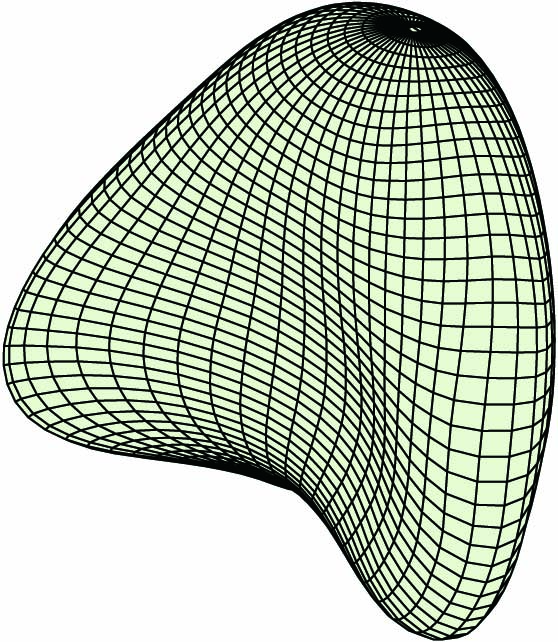}\qquad
 		\includegraphics[width = .14\linewidth]{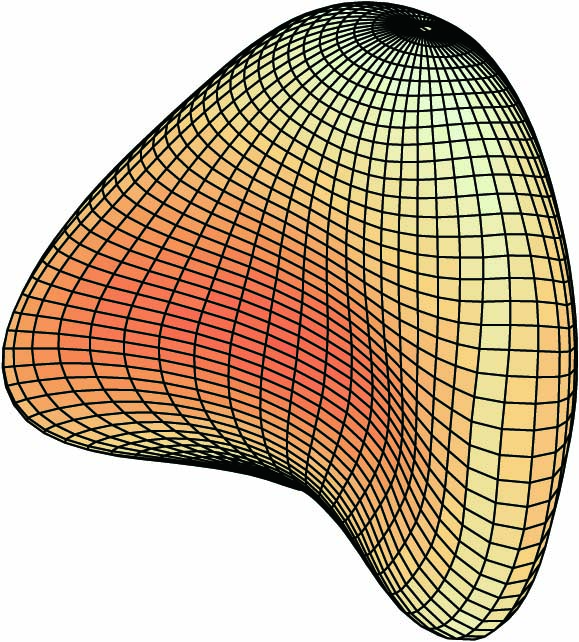}\qquad
 		\includegraphics[width = .14\linewidth]{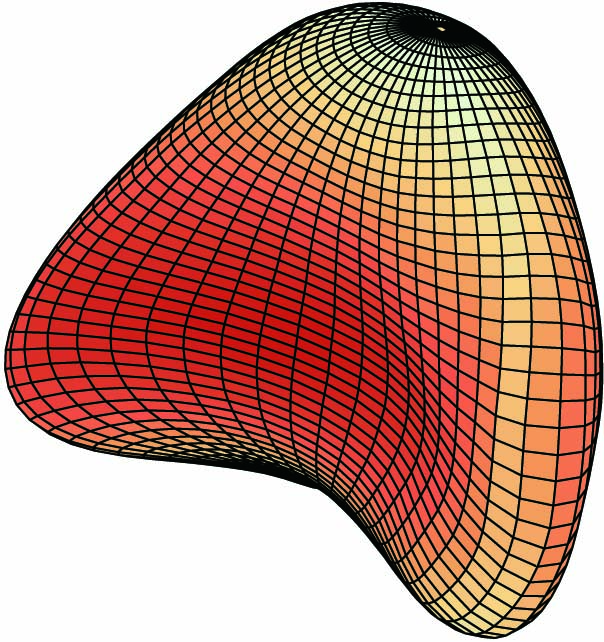}
		\qquad
 		\includegraphics[height = 3cm]{Colorbar2-1-1.jpg}
		}
		\caption*{\Large $\mu-\sigma \rightarrow \mu+\sigma$}
		\end{minipage}}}
		\caption{1st principal components of three subcortical structures. The figures show the deformation along the path of $\mu-\sigma \rightarrow \mu+\sigma$ following 1st principal direction. Color indicates the small patch's relative shape difference (deformation level) compared with the mean surface.}
		\label{1stPCs}
\end{figure}

Figure \ref{1stPCs} illustrates the 1st principal component of left hippocampus, left amygdala and left putamen surface shape. We show a principal direction using the elastic deformation path $\mu-\sigma \longrightarrow \mu \longrightarrow \mu+\sigma$.
Colors on a surface indicate the patch-wise shape differences of that surface when compared with the mean surface. Note that: (a) for left hippocampus, the largest shape variability is in the form of changes the angle of the posterior ends; (b) for left amygdala, the surfaces ''bends'' more towards the ''tail'' end along the 1st principal component; and (c) for left putamen, the curvature of the middle part changes along the 1st principal component.

\fboxsep=5pt
\fboxrule=2pt

The 2nd and 3rd principal components of three subcortical structure surfaces are shown in Supplementary Figs. 2, 3, and 4.  There we also provide interactive slider graphs to help visualize changes along different principal components for these subcortical structures.

\begin{figure}[!ht]
		\centering
		\resizebox{.45\linewidth}{!}{
			\fbox{
    				\subfloat[\large Left Hippocampus]{
					  	 		\includegraphics[width=.45\linewidth]{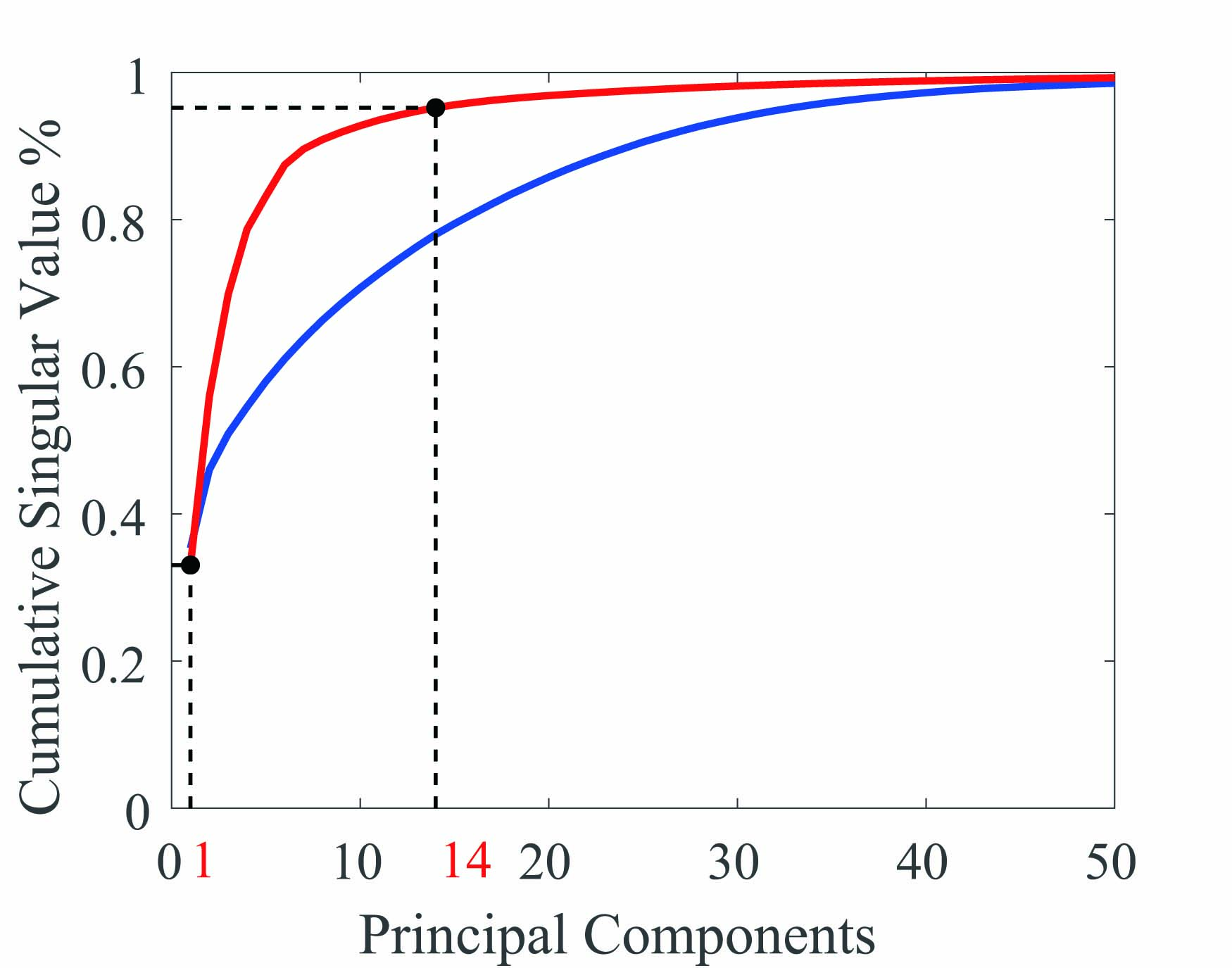}
					 }}
					 }
		 \qquad \qquad
    \resizebox{.45\linewidth}{!}{
			\fbox{
    				\subfloat[\large Left Amygdala]{
					  	 		\includegraphics[width=.45\linewidth]{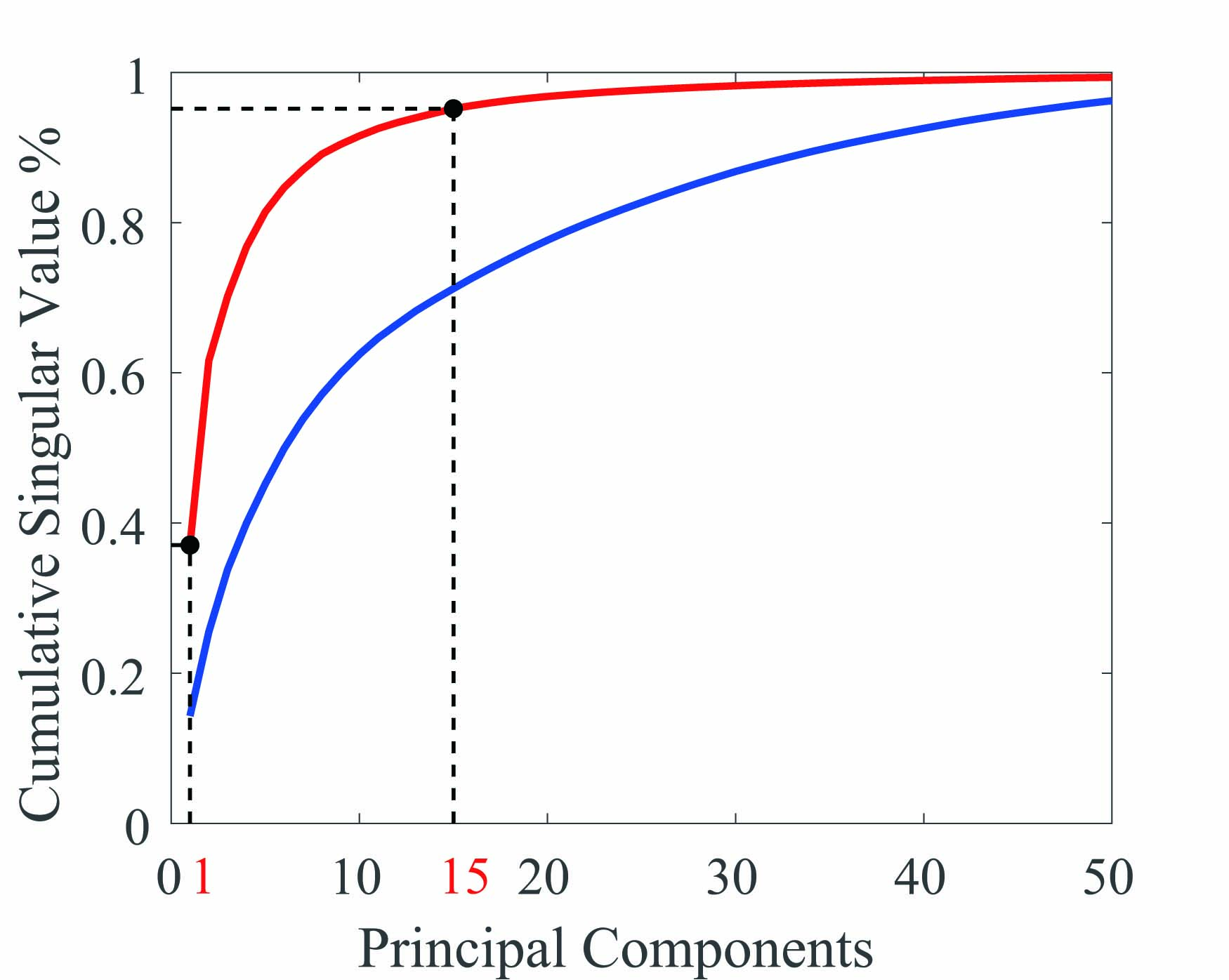}
					 }}
					 }
		 \resizebox{.45\linewidth}{!}{
		 \fbox{
    				\subfloat[\large Left Putamen]{
					  	 		\includegraphics[width=.45\linewidth]{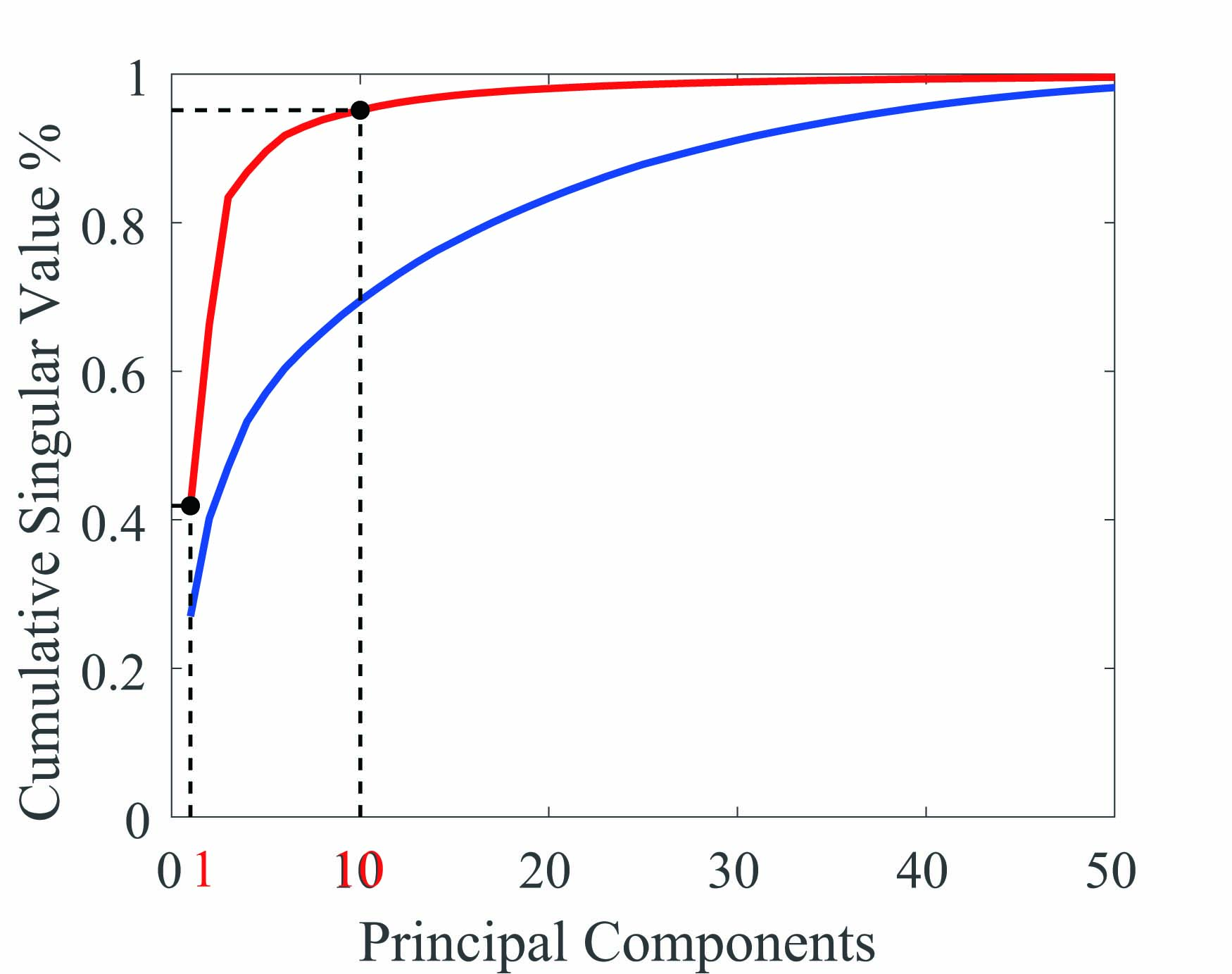}
					 }}
					 }
   \caption{The cumulative proportion of total singular values for two methods. Red lines stand for elastic shape analysis, and blue lines stand for vertex-wise analysis. Under elastic shape analysis, the 1st principal components can explain about 33\%, 37\% and 42\% variability of surface shape for left hippocampus, left amygdala and left putamen. The cumulative singular values take up more than 95\% of total singular values with 14,15 and 10 principal components respectively.}
   \label{PCvariability}
\end{figure}

We can quantify the level of shape variability explained by the principal components using the cumulative proportion of total singular values, as shown in Fig.~\ref{PCvariability} for both vertex-wise analysis and elastic shape analysis. Under elastic shape analysis, the 1st principal component explains about 33\%, 37\%, and 42\% variability for the left hippocampus, left amygdala, and left putamen, respectively. For these three structures, we can explain over 95\% variability of the surface shape difference with $14$, $15$, and $10$ principal components in total, respectively. Therefore, we will use the first 15 principal components to represent a shape in the subsequent regression analysis. Furthermore, by comparing elastic shape analysis (red lines) with vertex-wise analysis (blue lines), we conclude that elastic shape analysis explains more shape variability with the same number of principal components.

\fboxsep=4pt
\fboxrule=1pt

\begin{figure}[!ht]
		\centering
		\captionsetup[subfloat]{position = bottom}
		\resizebox{.9\linewidth}{!}{      
		\subfloat[\large Reconstructed Surfaces]{
 		\includegraphics[width = .12\linewidth]{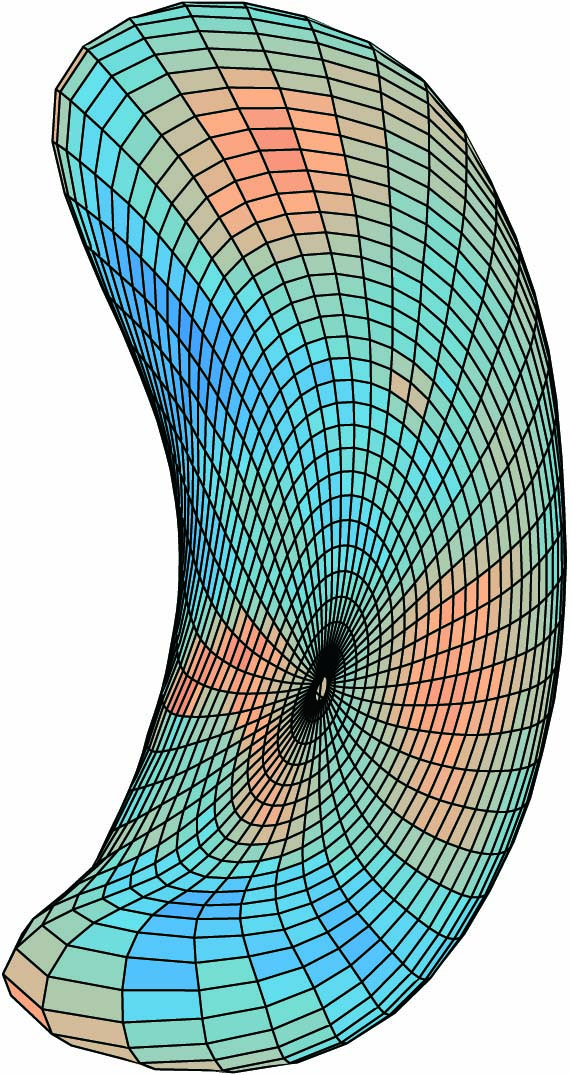}
 		\includegraphics[width = .12\linewidth]{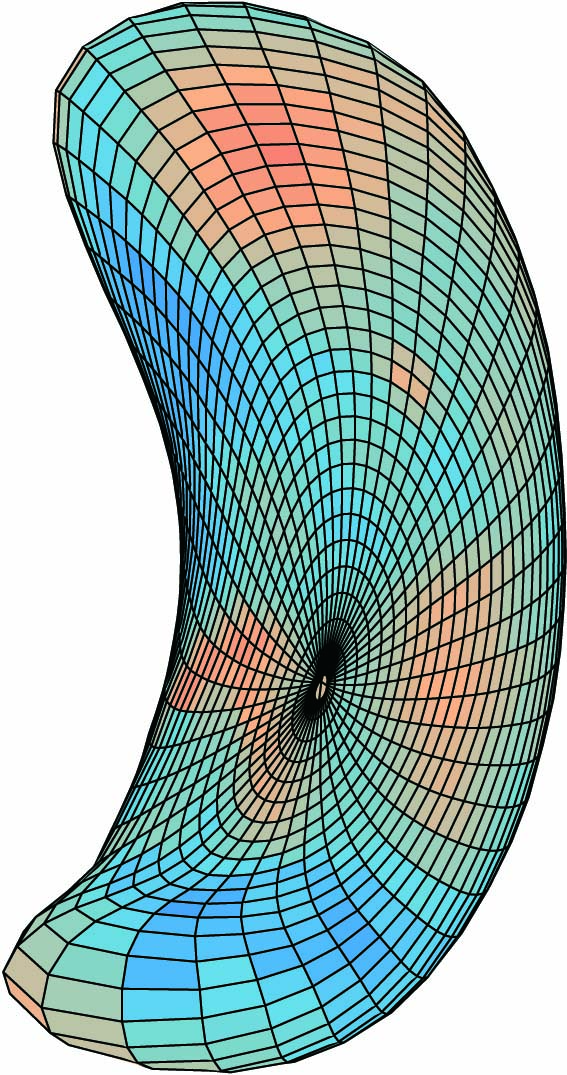}
 		\includegraphics[width = .15\linewidth]{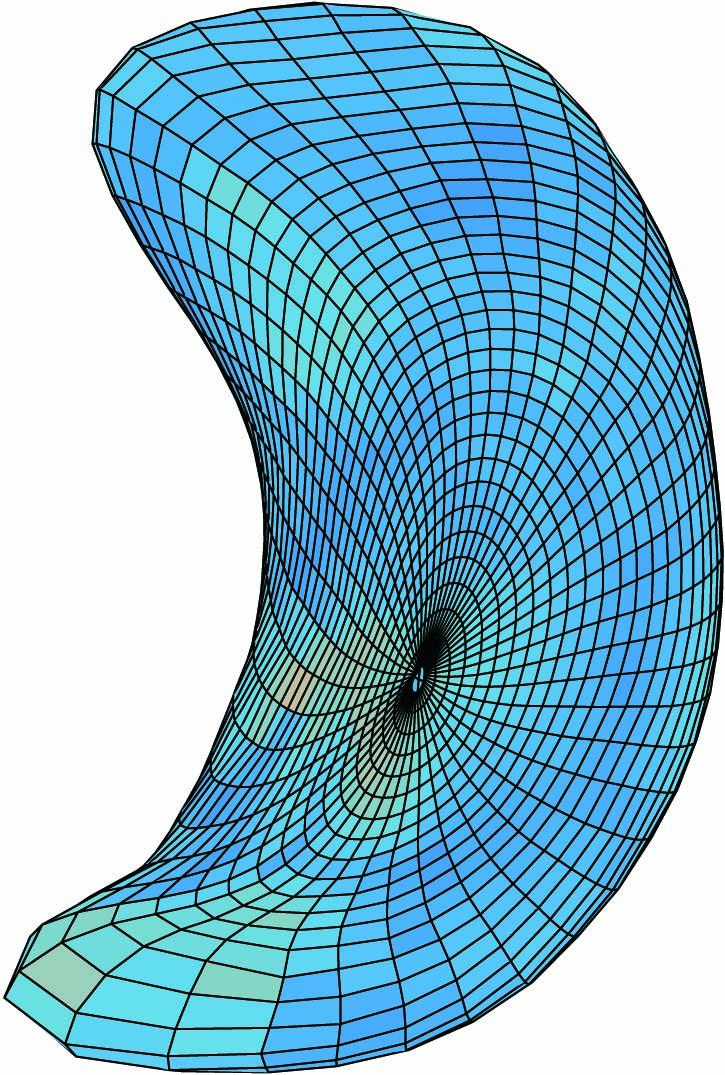}
 		\includegraphics[width = .15\linewidth]{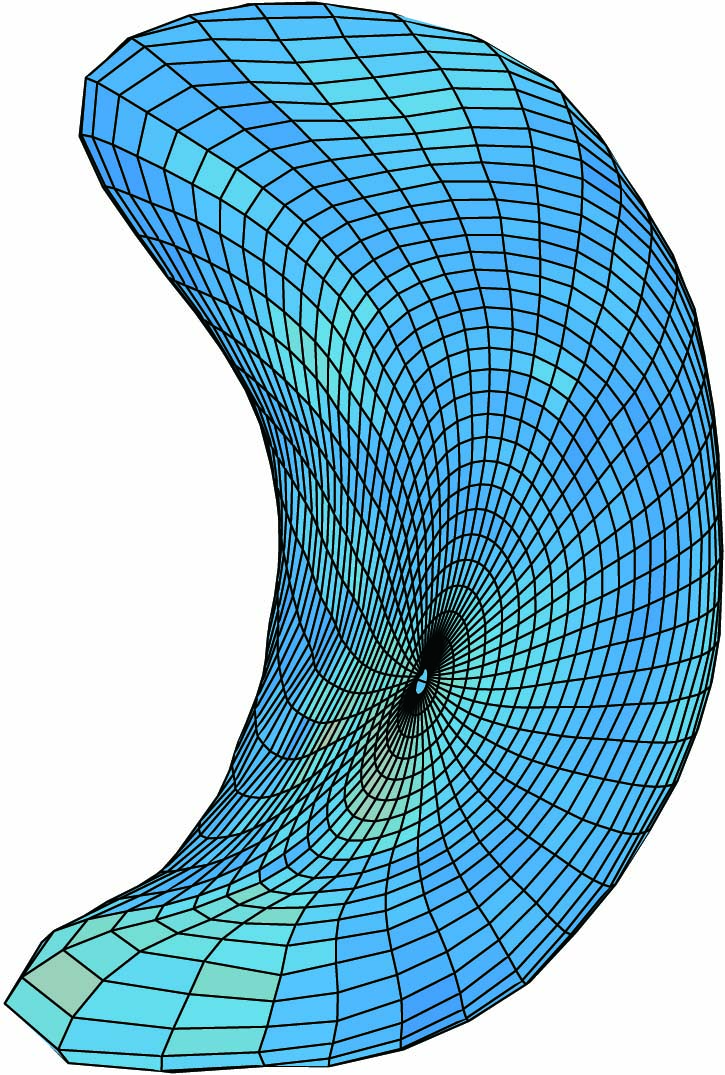}
 		\includegraphics[width = .15\linewidth]{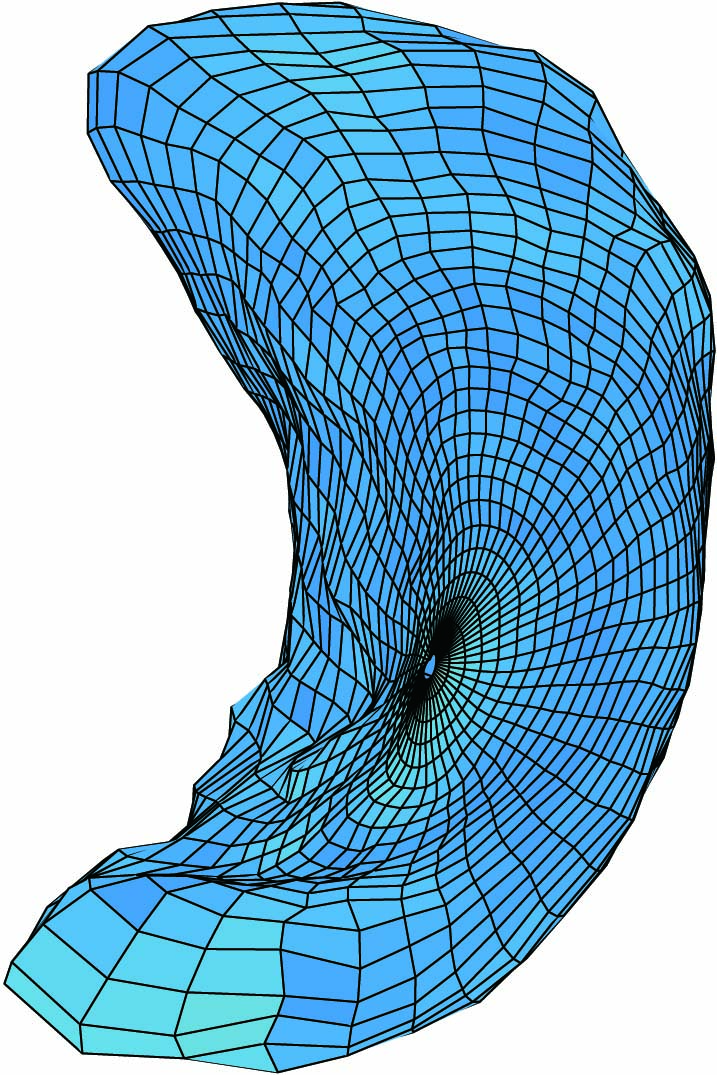}
		\hspace{0.1cm}
 		\includegraphics[height = 3cm]{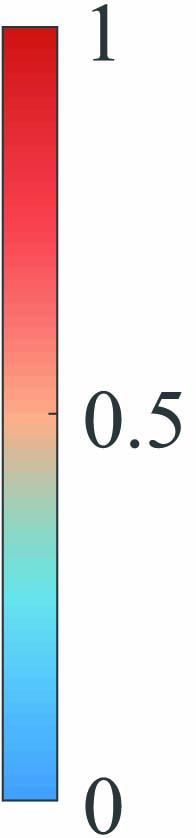}
		}
		\hfill
		\fbox{
		\subfloat[\large Target]{
 		\includegraphics[width = .14\linewidth]{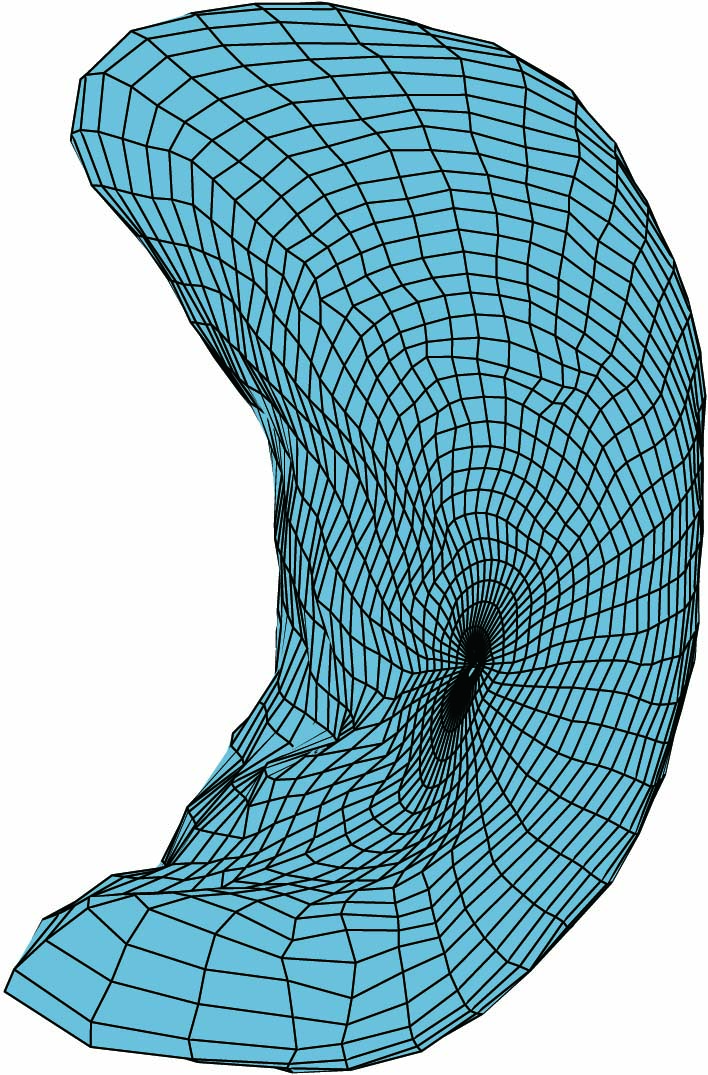}
		}}
		}
		\resizebox{.9\linewidth}{!}{     
		\subfloat[\large Reconstructed Surfaces]{
 		\includegraphics[width = .13\linewidth]{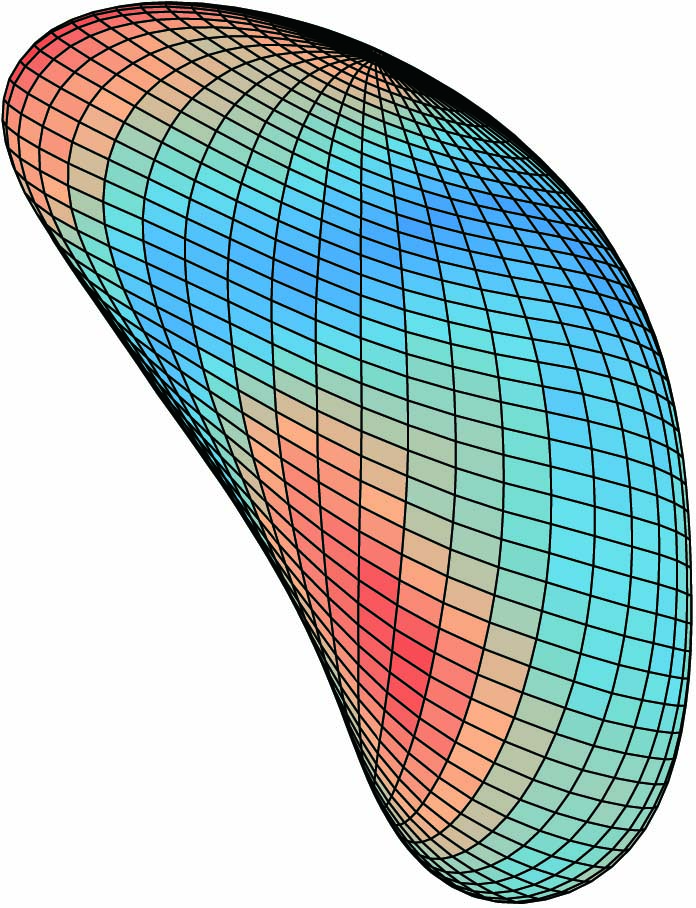}
 		\includegraphics[width = .13\linewidth]{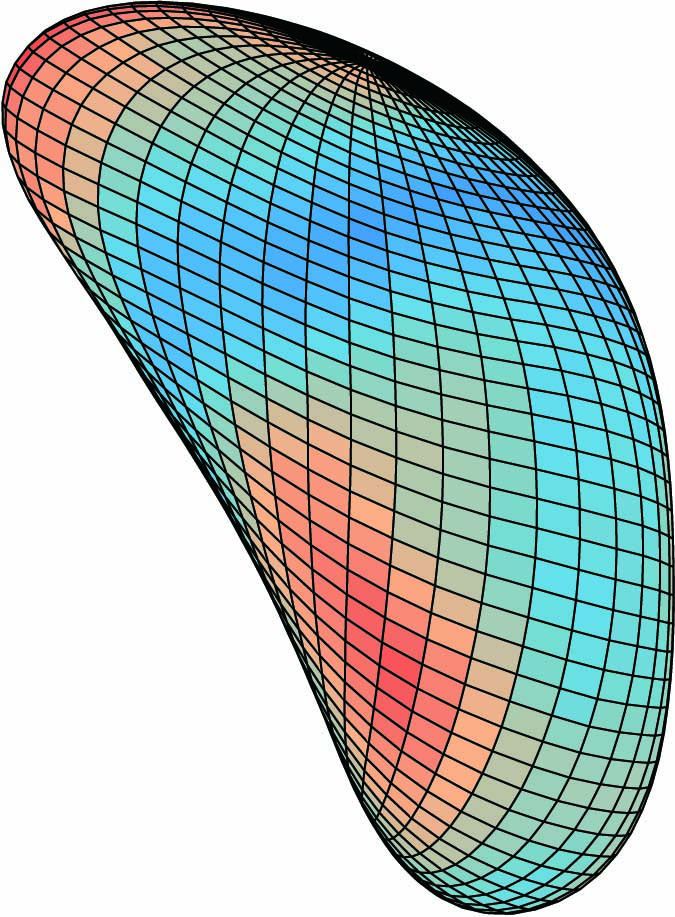}
 		\includegraphics[width = .14\linewidth]{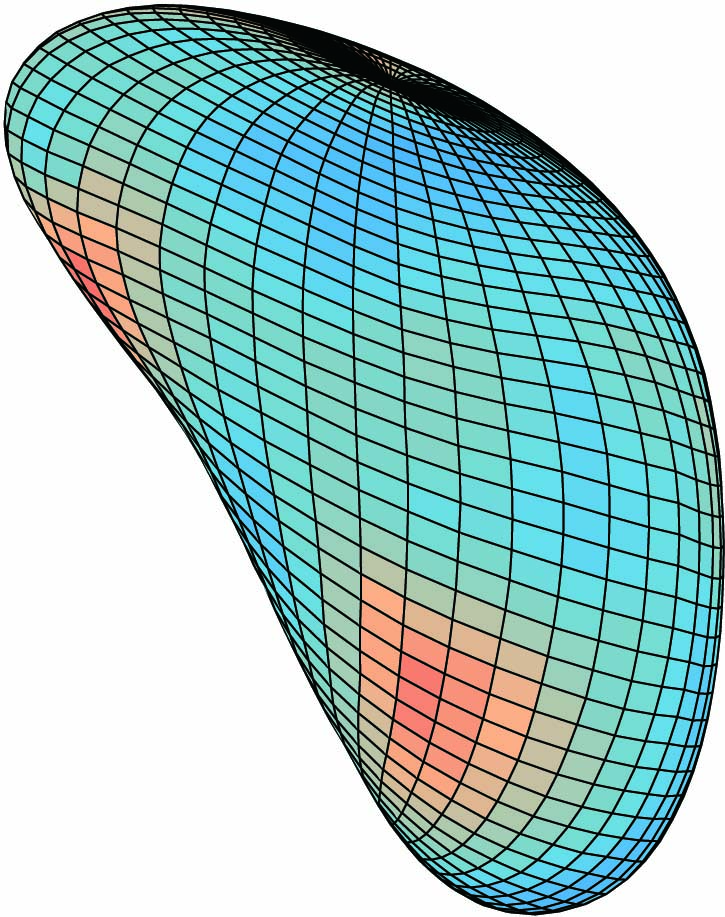}
 		\includegraphics[width = .14\linewidth]{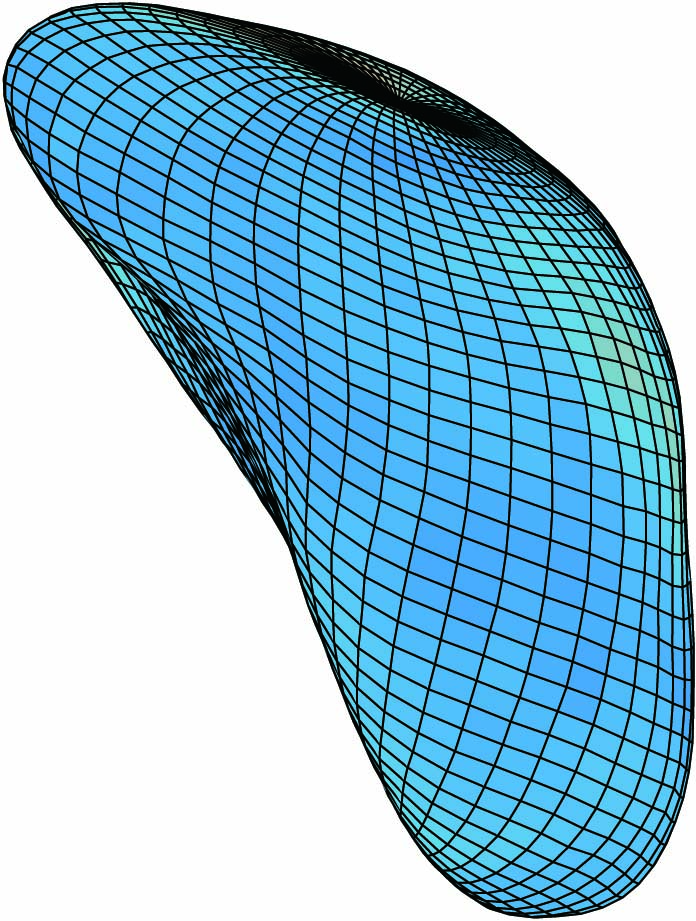}
 		\includegraphics[width = .14\linewidth]{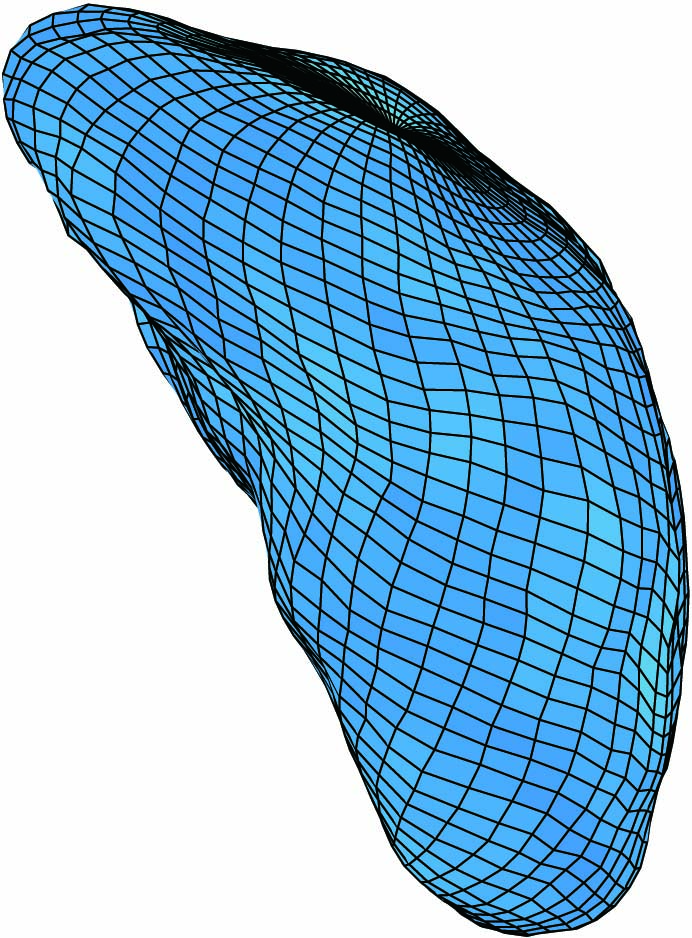}
		\hspace{0.1cm}
 		\includegraphics[height = 3cm]{Colorbar3.jpg}
   }
   \hfill
		\fbox{
		\subfloat[\large Target]{
 		\includegraphics[width = .14\linewidth]{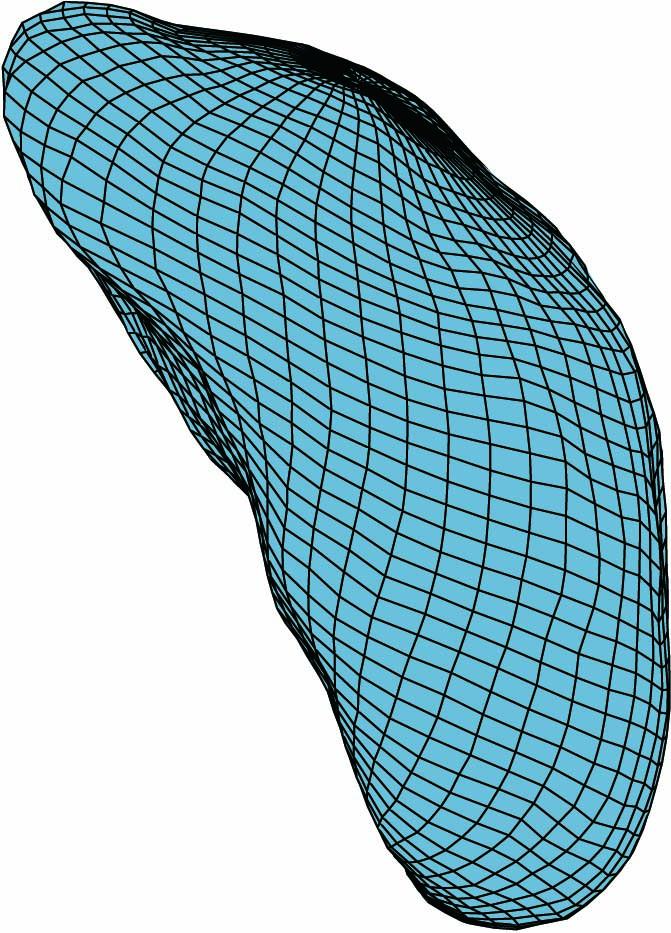}
		}}
		}
		\resizebox{.9\linewidth}{!}{ 
		\subfloat[\large Reconstructed Surfaces]{
 		\includegraphics[width = .13\linewidth]{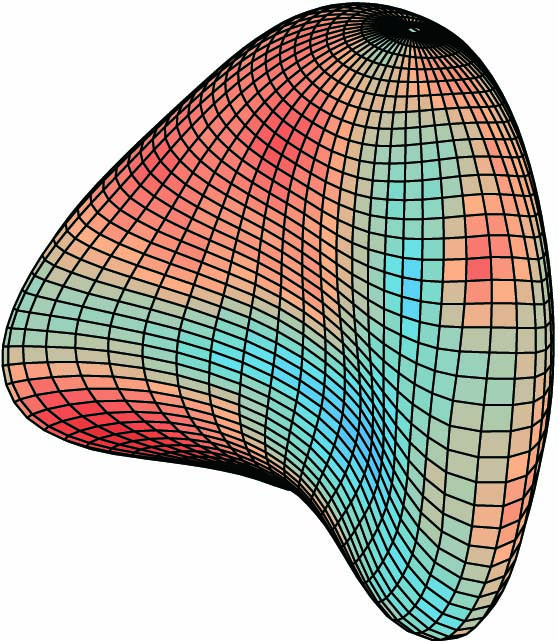}
 		\includegraphics[width = .13\linewidth]{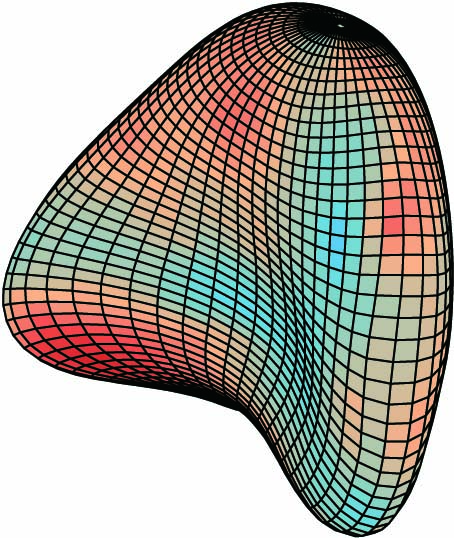}
 		\includegraphics[width = .14\linewidth]{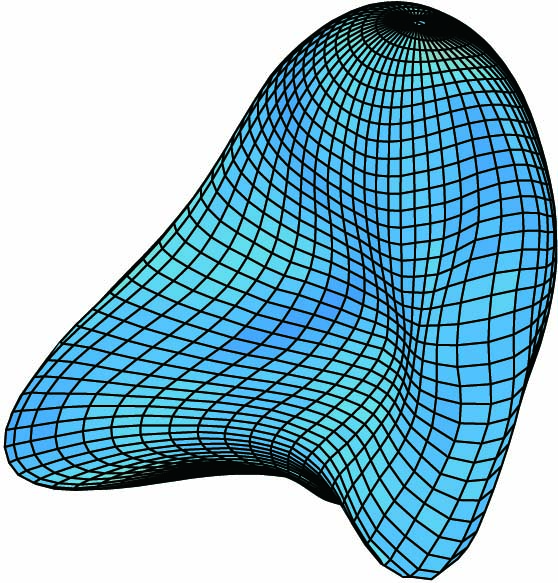}
 		\includegraphics[width = .14\linewidth]{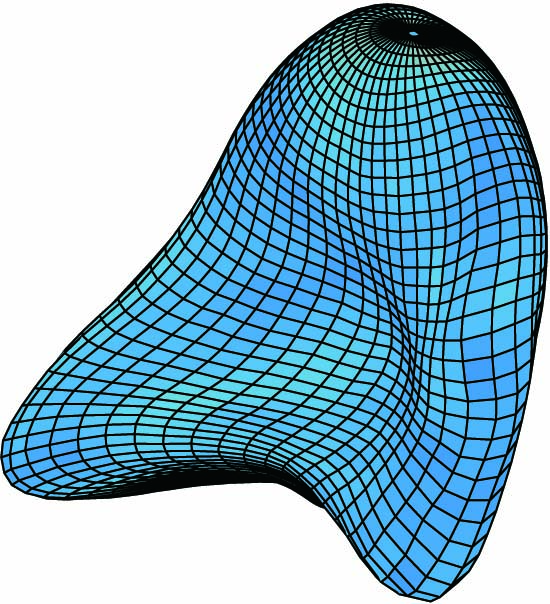}
 		\includegraphics[width = .14\linewidth]{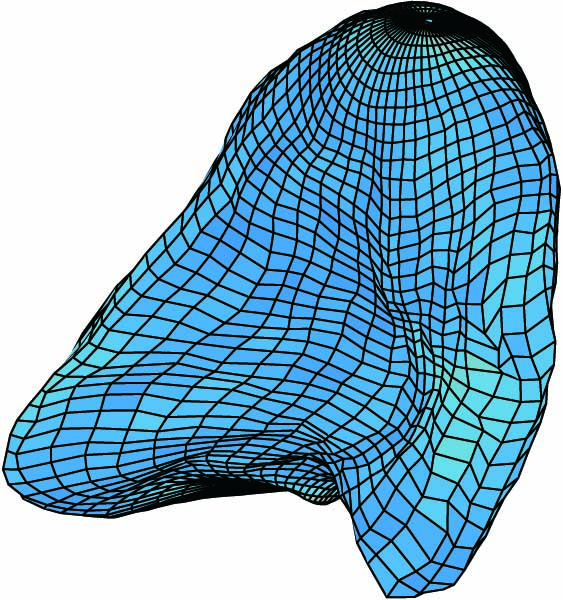}
		\hspace{0.1cm}
 		\includegraphics[height = 3cm]{Colorbar3.jpg}
   }
   \hfill
		\fbox{
		\subfloat[\large Target]{
 		\includegraphics[width = .14\linewidth]{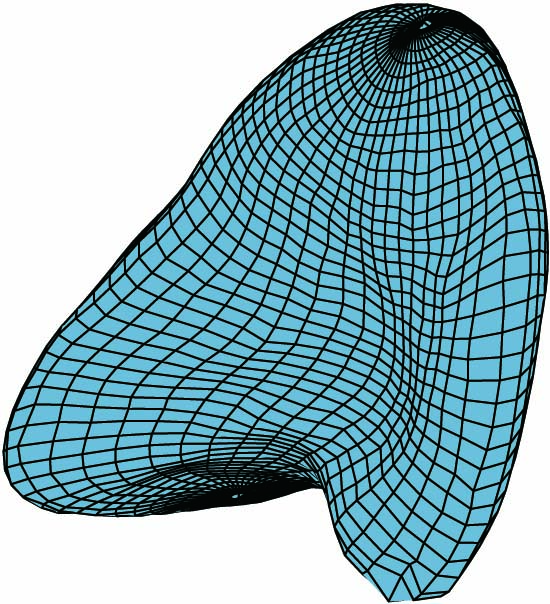}
		}}
		}
		\caption{Sample surface reconstructions of three subcortical structures. (a), (c) and (e) are reconstructed surfaces with first 0, 1, 5, 15 and 72 principal components. (b), (d) and (f) are target surfaces. Color indicates the small patch's relative shape difference compared with the target surface.} 
		\label{Reconresult}
\end{figure}

An important strength of this shape representation is that we can map these features back to the object space. 
Figure~\ref{Reconresult} presents some examples of such reconstructions. We use $90 \times 0.8=72$ surfaces to compute principal components and the other 18 surfaces to reconstruct and test. The right column shows sample individual surfaces for each subcortical structure, and the left side shows surfaces reconstructed using ${f}_i = \mu + \sum_{j=1}^d{x}_{i,j} v_j$. The surfaces are reconstructed with $d=0, 1, 5, 15$, and $72$ principal components added to the mean surface respectively. Color indicates the relative shape difference between reconstructed surface and example surface $|\hat{f}_i - f_i |$. As more and more principal components added, the patches change color from red to blue, and the shapes of reconstructed surfaces look similar to the sample surfaces. When $d=72$ principal components (all of the principal components) are used, the reconstructed is almost identical to the original surface. The reconstructed surfaces are mostly blue, which means the difference between reconstructed surface and example surface is relatively very small. This result illustrates that elastic mean and PCA successfully capture the modes of shape variations in subcortical structure surfaces, and represent individual shapes using a small number of PCA coefficients. 
We provide GIF examples of the surface reconstruction in the Supplementary Material.

\subsection{Regression Models}
Next we use these compact shape representations as predictors in regression models. 
Specifically, we study the ten linear regression models listed in Table~\ref{Model design}. In this study, we use bidirectional stepwise regression to select most significant principal components. Table~\ref{Model result R} shows the adjusted $R^2$ for the fitted models and Tables~\ref{Model result pss},  \ref{Model result ctqtot}, and \ref{Model result inter}
list the significant principal components with their signs of regression coefficients and the corresponding $p$-values, and significant interaction terms for each model.

\begin{table}[h] 
				\centering
				\caption{Adjusted $R^2$ of Regression Models}
				\label{Model result R}
				\begin{threeparttable}
				\centering
				\begin{tabular}{|c|l|c|c|l|c|}
				\hline
				\textbf{\#} & \textbf{Model} & \textbf{$R^2$} & \textbf{\#} & \textbf{Model} & \textbf{$R^2$}\\
				\hline
				1 & PSS $\sim$ Age+BDI+PS+Interactions & 0.64 & 5 & CTQTOT $\sim$ Age+BDI+PS+Interactions & 0.70\\
				2 & PSS $\sim$ Age+BDI+PS & 0.48 & 6 & CTQTOT $\sim$ Age+BDI+PS & 0.39\\
				3 & PSS $\sim$ Age+BDI & 0.19 & 7 & CTQTOT $\sim$ Age+BDI & 0.11\\
				4 & PSS $\sim$ PS & 0.31 & 8 & CTQTOT $\sim$ PS & 0.29\\
				\hline
				9 & PSS $\sim$ Model 1 predictors+ICV & 0.64 & 10 & CTQTOT $\sim$ Model 1 predictors+ICV & 0.70\\
				\hline
				\end{tabular}
				\begin{tablenotes}
      \small
      \item PS: First 15 principal scores; Interactions: age $\times$ first 5 PS and BDI $\times$ first 5 PS.
    \end{tablenotes}
		\end{threeparttable}
\end{table}

From Table~\ref{Model result R}, we can see that the models including interactions between shape (PS) and confounding variables (age and BDI) have higher adjusted-$R^2$ value than the models excluding interactions. Both Models 1 and 5 achieve large $R^2$ values.   Although BDI is highly correlated with PTSD symptoms, when we compare Models 2, 4, 6, and 8 with Models 3 and 5 that only includes age and BDI, we reach a higher $R^2$. This implies that shape explains more variability in the responses (PSS and CTQTOT) than BDI. Next, we focus on the significant shape principal components of each subcortical structure.

\begin{table}[h] 
				\centering
				\caption{Significant Principal Components for PTSD Symptom}
				\label{Model result pss}
				\begin{threeparttable}
				\centering
				\begin{tabular}{c|c|c|c}
				\textbf{} & \textbf{Predictor} & \textbf{Sign of Coefficient} & \textbf{p-value}\\
				\hline
				1 & amygdala 4\textit{th} & + & 0.0045\\
				2 & amygdala 5\textit{th} & + & 0.0005\\
				3 & amygdala 6\textit{th} & + & 0.006\\
				4 & hippocampus 2\textit{nd} & + & 0.001\\
				5 & putamen 6\textit{th} & + & 0.004\\
				\end{tabular}
				\begin{tablenotes}
      \small
      \item Predictors are significant under significance level 0.01.
    \end{tablenotes}
		\end{threeparttable}
\end{table}

\begin{table}[h] 
				\centering
				\caption{Significant Principal Components for Childhood Traumatic Experience}
				\label{Model result ctqtot}
				\begin{threeparttable}
				\centering
				\begin{tabular}{c|c|c|c}
				\textbf{} & \textbf{Predictor} & \textbf{Sign of Coefficient} & \textbf{p-value}\\
				\hline
				1 & amygdala 2\textit{nd} & + & 0.02\\
				2 & amygdala 11\textit{th} & - & 0.0008\\
				3 & hippocampus 5\textit{th} & - & 0.04\\
				4 & putamen 2\textit{nd} & - & 0.01\\
				5 & putamen 4\textit{th} & - & 0.03\\
				\end{tabular}
				\begin{tablenotes}
      \small
      \item Predictors are significant under significance level 0.05.
    \end{tablenotes}
		\end{threeparttable}
\end{table}

From Table~\ref{Model result pss}, we see that PTSD symptoms are most correlated with subcortical shapes of the following principal components: amygdala 4\textit{th}, amygdala 5\textit{th}, amygdala 6\textit{th}, hippocampus 2\textit{nd} and putamen 6\textit{th}. Similarly, from Table~\ref{Model result ctqtot}, the most significant subcortical shape changes associated with childhood traumatic experience lie in the principal directions: amygdala 2\textit{nd}, amygdala 11\textit{th}, hippocampus 5\textit{th}, putamen 2\textit{nd} and putamen 4\textit{th}. After controlling for ICV, the principal shape components of elastic shape analysis are still found to be statistically significant.
From Table~\ref{Model result inter}, we discover the significant interactions between shape and confounding patterns correlated with PTSD and traumatic experience.

\begin{table}[h] 
				\centering
				\caption{Significant Interactions of Model 1 and 5}
				\label{Model result inter}
				\begin{threeparttable}
				\centering
				\begin{tabular}{c|c|c|c}
				\textbf{Predictor for PSS} & \textbf{p-value} & \textbf{Predictor for CTQTOT} & \textbf{p-value}\\
				\hline
				age $\times$ amygdala 4\textit{th} & 0.002 & age $\times$ amygdala 1\textit{st} & $2*10^{-5}$\\
				BDI $\times$ amygdala 4\textit{th} & 0.005 & age $\times$ amygdala 2\textit{nd} & 0.0007\\
				age $\times$ hippocampus 3\textit{rd} & 0.0005 & age $\times$ amygdala 4\textit{th} & $3*10^{-6}$\\
				BDI $\times$ hippocampus 3\textit{rd} & $5*10^{-5}$ & BDI $\times$ amygdala 3\textit{rd} & $7*10^{-6}$\\
				BDI $\times$ hippocampus 4\textit{th} & 0.002 & BDI $\times$ amygdala 4\textit{th} & $1*10^{-5}$\\
				age $\times$ putamen 1\textit{st}  & 0.001 & age $\times$ hippocampus 4\textit{th} & 0.001\\
				age $\times$ putamen 2\textit{nd}  & 0.005 & BDI $\times$ hippocampus 5\textit{th} & $8*10^{-5}$\\
				age $\times$ putamen 4\textit{th}  & 0.01 & age $\times$ putamen 1\textit{st} & 0.01\\
				age $\times$ putamen 5\textit{th}  & $8*10^{-5}$ & age $\times$ putamen 2\textit{nd} & $3*10^{-5}$\\
				BDI $\times$ putamen 1\textit{st}  & 0.0007 & age $\times$ putamen 3\textit{rd} & $3*10^{-6}$\\
				                  & & BDI $\times$ putamen 2\textit{nd} & 0.005\\
				                  & & BDI $\times$ putamen 3\textit{rd} & 0.002\\
				                  & & BDI $\times$ putamen 4\textit{th} & $8*10^{-6}$\\
				\end{tabular}
				\begin{tablenotes}
      \small
      \item Predictors are significant under significance level 0.01.
    \end{tablenotes}
		\end{threeparttable}
\end{table}

\subsection{Shape Pattern}
\begin{figure}[!ht]
		\centering
		\resizebox{.9\linewidth}{!}{
		\fbox{\begin{minipage}{\dimexpr\textwidth-2\fboxsep-2\fboxrule\relax}
		\subfloat[\Large 2nd Principal Component of Left Hippocampus]{
		\qquad
 		\includegraphics[height = 4cm]{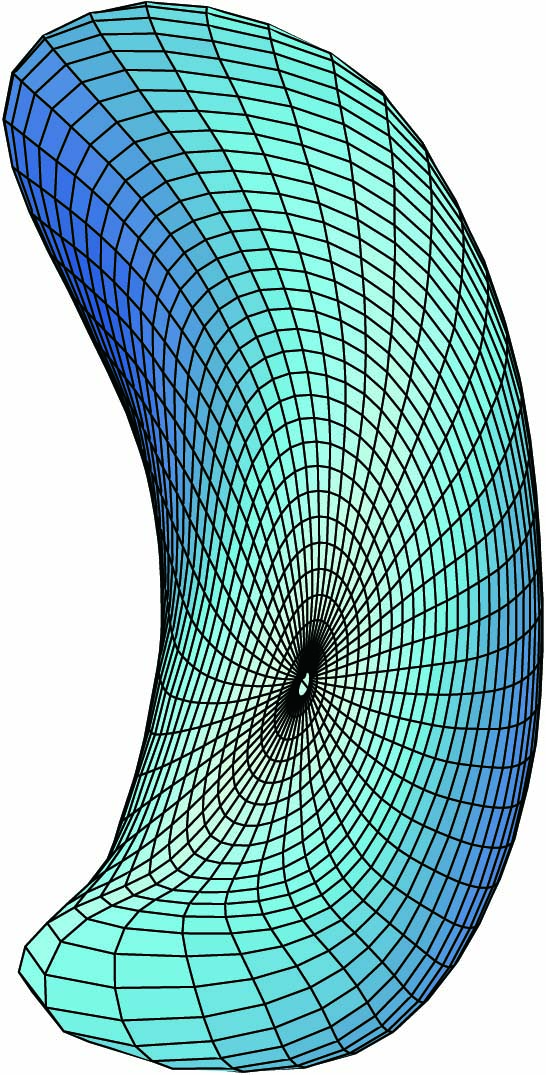}\hspace{0.55cm}
 		\includegraphics[height = 4cm]{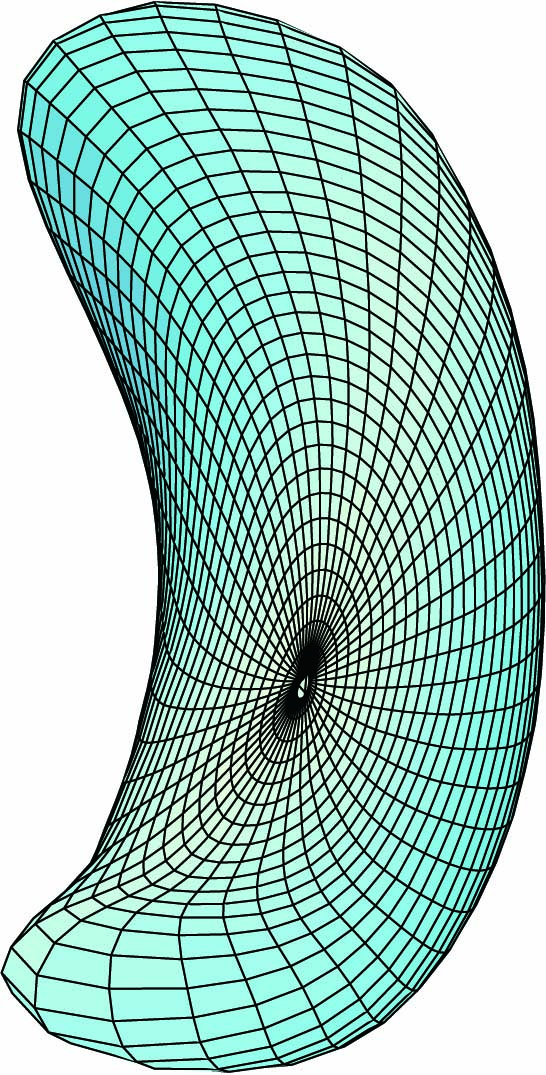}\hspace{0.55cm}
 		\includegraphics[height = 4cm]{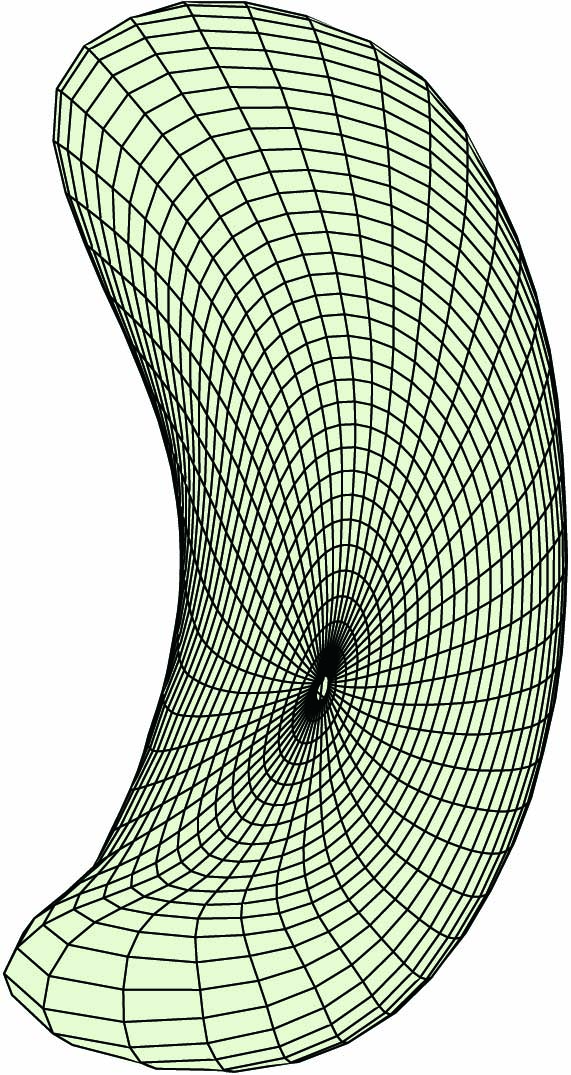}\hspace{0.55cm}
 		\includegraphics[height = 4cm]{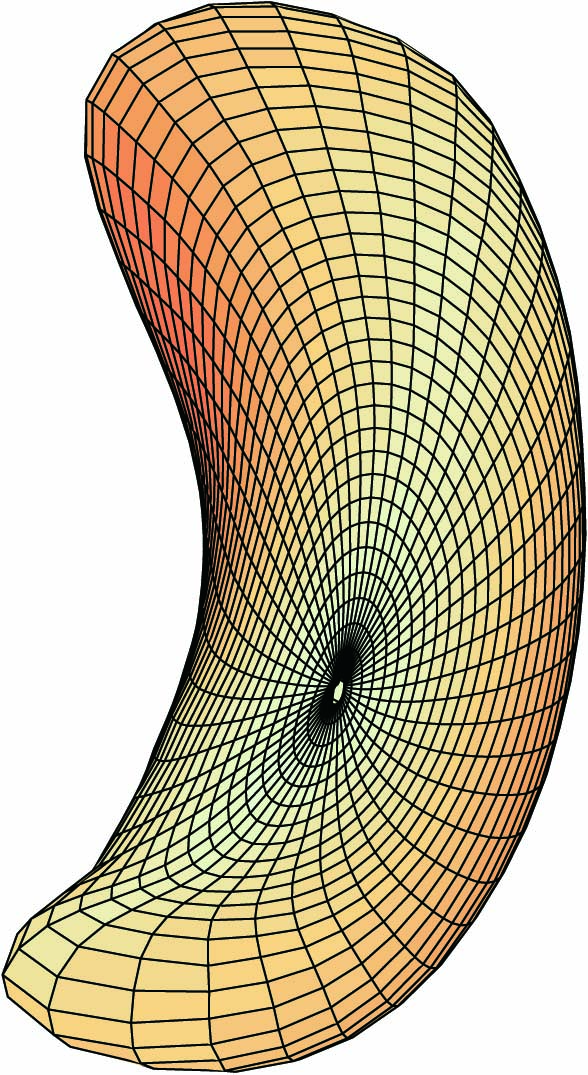}\hspace{0.55cm}
 		\includegraphics[height = 4cm]{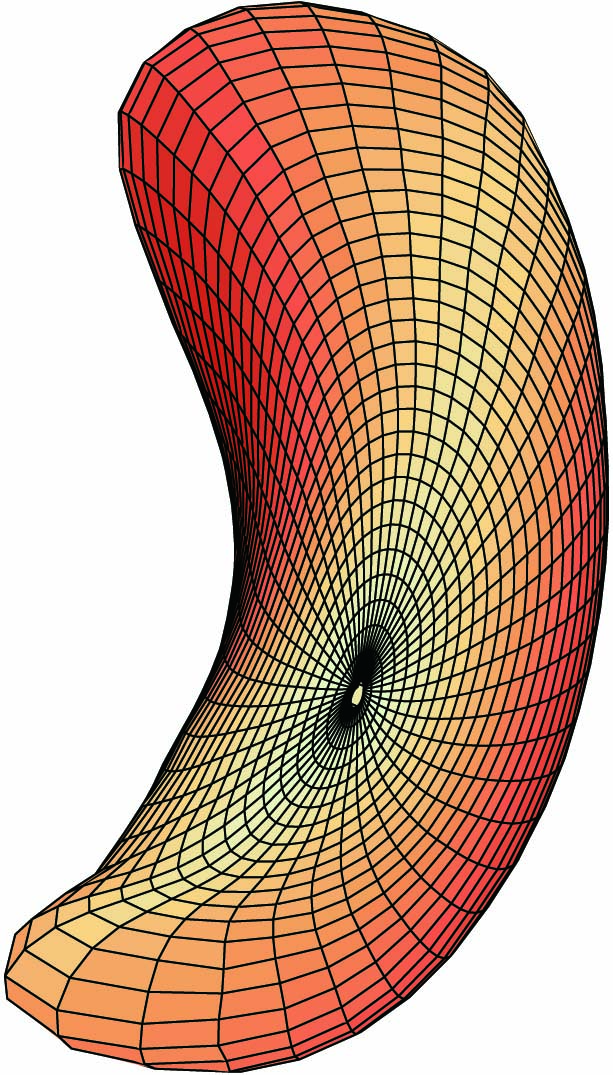}
		\qquad
 		\includegraphics[height = 4cm]{Colorbar2-1-1.jpg}
		}
		\caption*{\Large lower PSS $\leftarrow \rightarrow$ higher PSS}
		\end{minipage}}
		}
		
		\resizebox{.9\linewidth}{!}{
		\fbox{\begin{minipage}{\dimexpr\textwidth-2\fboxsep-2\fboxrule\relax}
		\subfloat[\Large 4th, 5th and 6th Principal Component of Left Amygdala]{
		\qquad
 		\includegraphics[width = .16\linewidth]{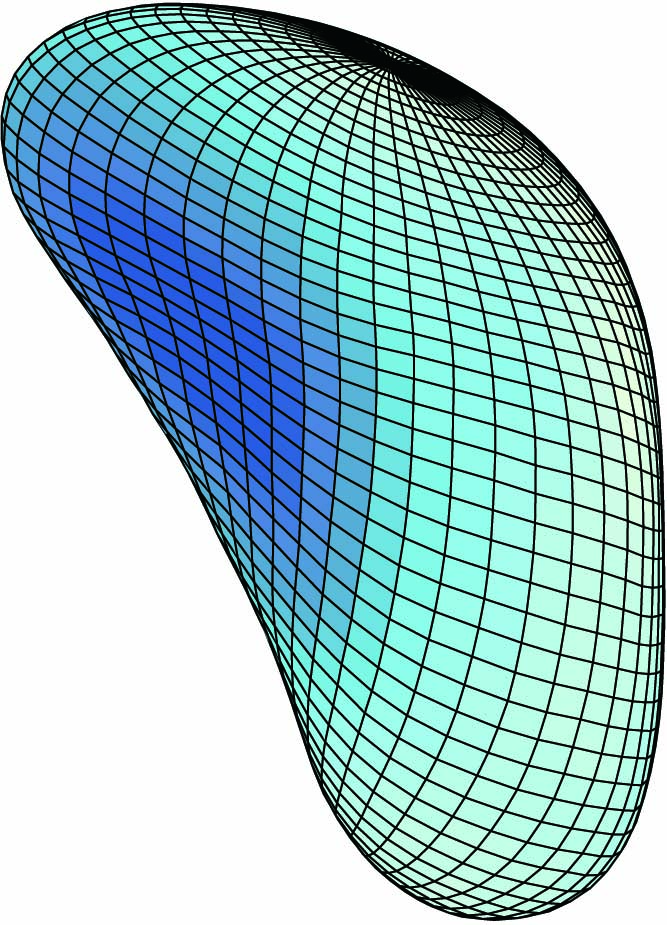}
 		\includegraphics[width = .16\linewidth]{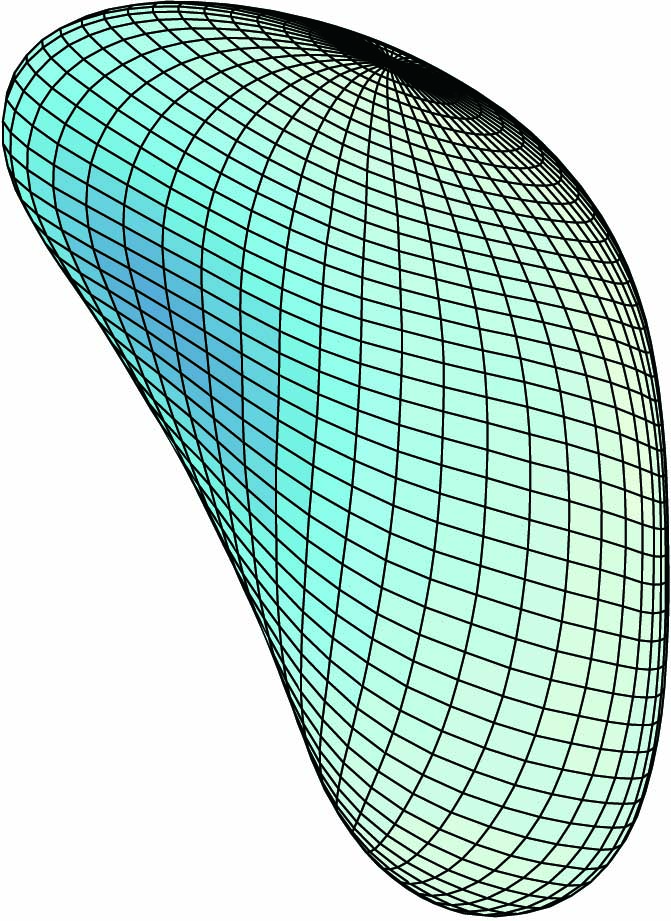}
 		\includegraphics[width = .16\linewidth]{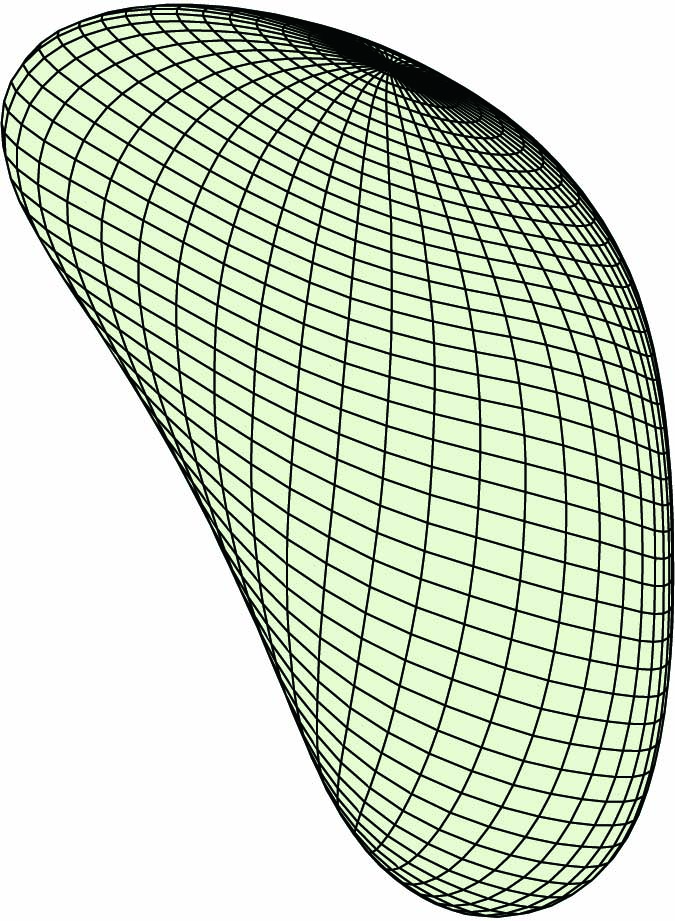}
 		\includegraphics[width = .16\linewidth]{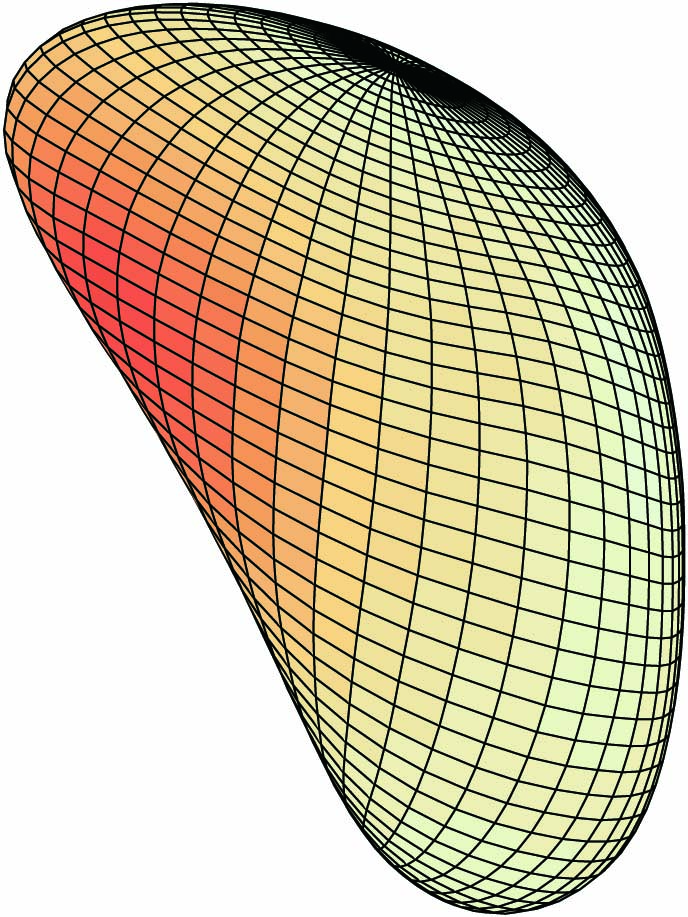}
 		\includegraphics[width = .16\linewidth]{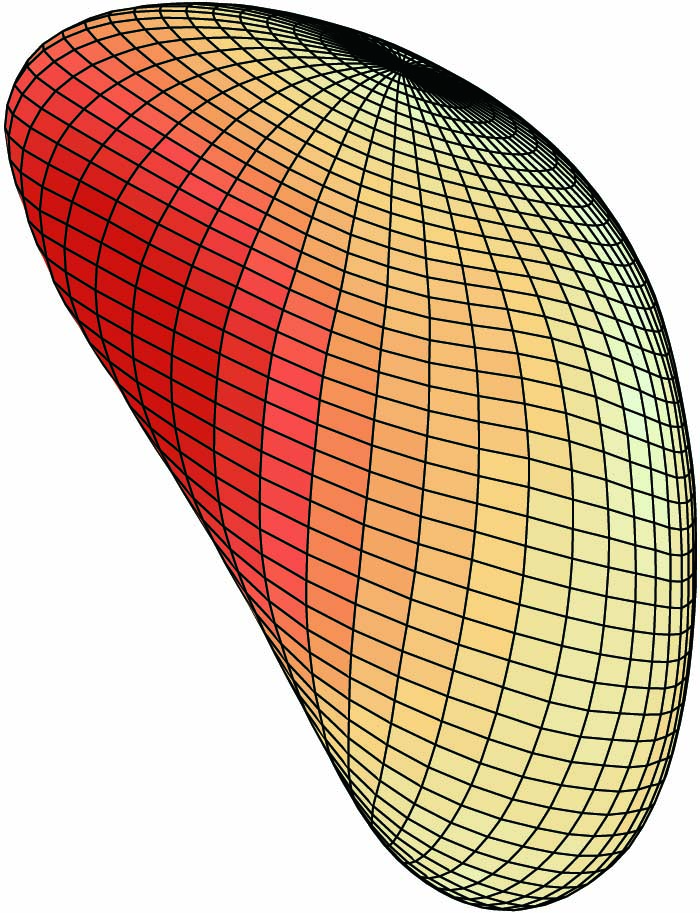}
		\qquad
 		\includegraphics[height = 3.5cm]{Colorbar2-1-1.jpg}
		}
		\caption*{\Large lower PSS $\leftarrow \rightarrow$ higher PSS}
		\end{minipage}}
		}
		
		\resizebox{.9\linewidth}{!}{
		\fbox{\begin{minipage}{\dimexpr\textwidth-2\fboxsep-2\fboxrule\relax}
		\subfloat[\Large 6th Principal Component of Left Putamen]{
		\qquad
 		\includegraphics[width = .16\linewidth]{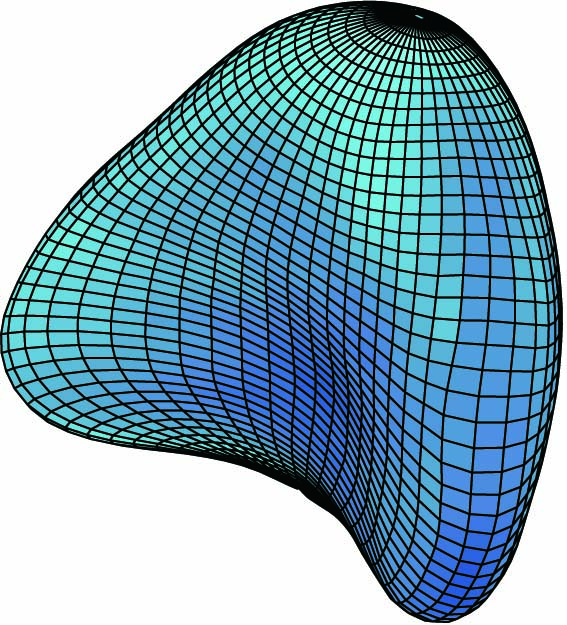}
 		\includegraphics[width = .16\linewidth]{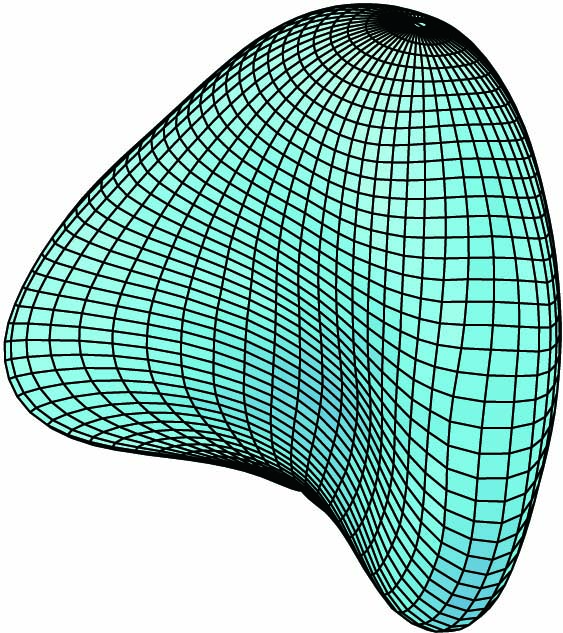}
 		\includegraphics[width = .16\linewidth]{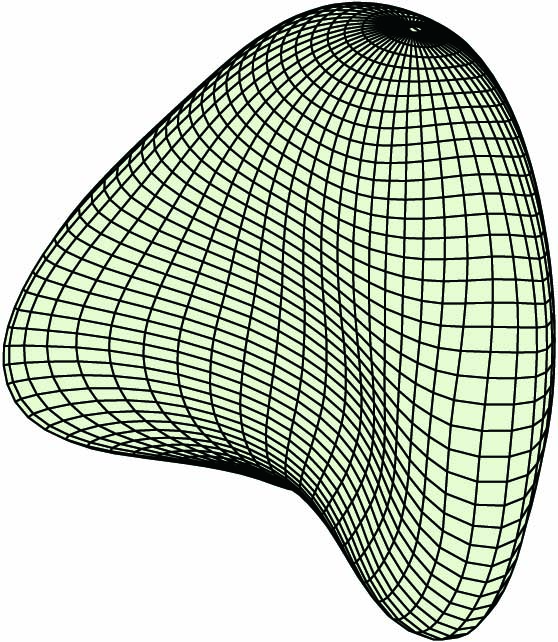}
 		\includegraphics[width = .16\linewidth]{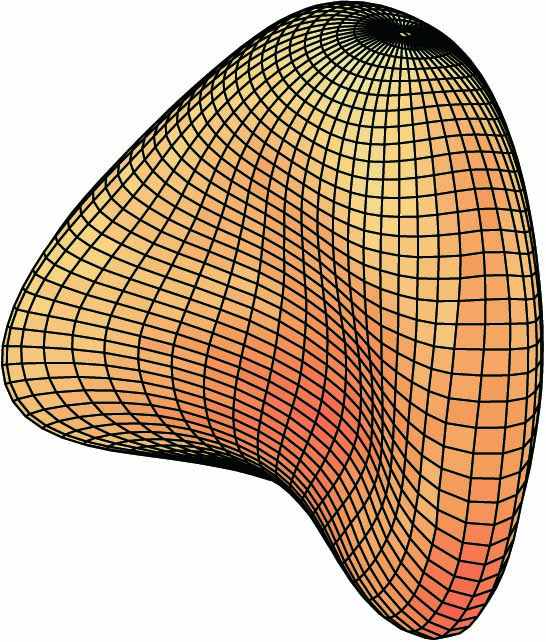}
 		\includegraphics[width = .16\linewidth]{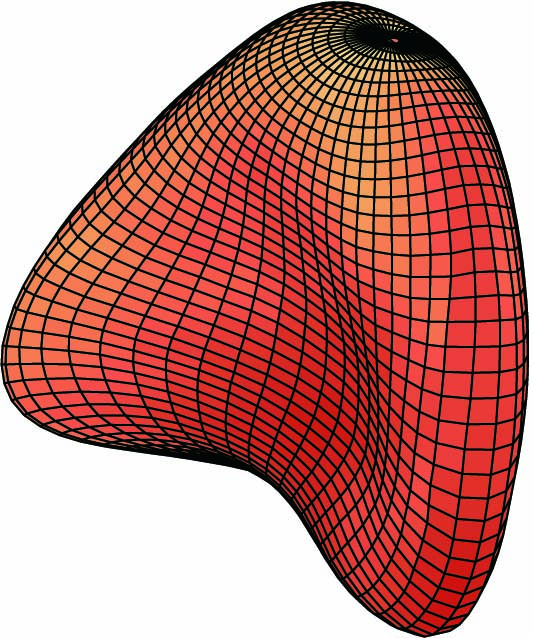}
		\qquad
 		\includegraphics[height = 3.5cm]{Colorbar2-1-1.jpg}
		}
		\caption*{\Large lower PSS $\leftarrow \rightarrow$ higher PSS}
		\end{minipage}}
		}
		\caption{Shape deformation along the most significant principal components on PTSD symptom scale. The surfaces to the left have less severe or no PTSD symptoms, and the surfaces to the right have more severe PTSD symptoms. Color indicates the small patch's relative shape difference (deformation level) along the direction.}
		\label{pattern PSS}
\end{figure}

In order to understand shape patterns of subcortical surfaces with different PTSD symptom scales and childhood traumatic experience inventories, we visualize the significant principal components for each subcortical structure. From Fig.~\ref{pattern PSS}, we observe that with more severe PTSD symptoms:
\begin{itemize}
	\item Left hippocampus surface moves along the positive direction of 2nd principal component, which shows a shrunken anterior end and curved body part;
	\item Moving along the positive direction of 4th, 5th and 6th principal components, the left amygdala surface mainly deforms at the ''head'' end, where the central nucleus lies. The ''head'' end tends to indent;
	\item The surfact of the left putamen's concave middle part has greater curvature, and the end part gets thinner and sharper along the positive direction of 6th principal component.
\end{itemize}

\fboxsep=5pt
\fboxrule=2pt

\begin{figure}[!ht]
		\centering
		\resizebox{.9\linewidth}{!}{
		\fbox{\begin{minipage}{\dimexpr\textwidth-2\fboxsep-2\fboxrule\relax}
		\subfloat[\Large 5th Principal Component of Left Hippocampus]{
		\qquad
 		\includegraphics[height = 4cm]{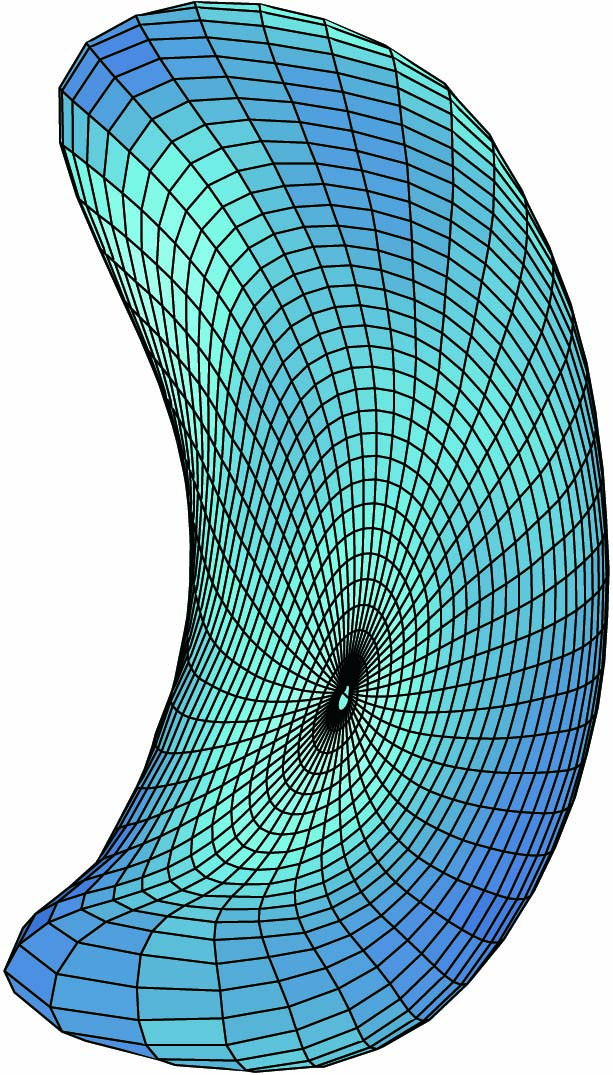}\hspace{0.55cm}
 		\includegraphics[height = 4cm]{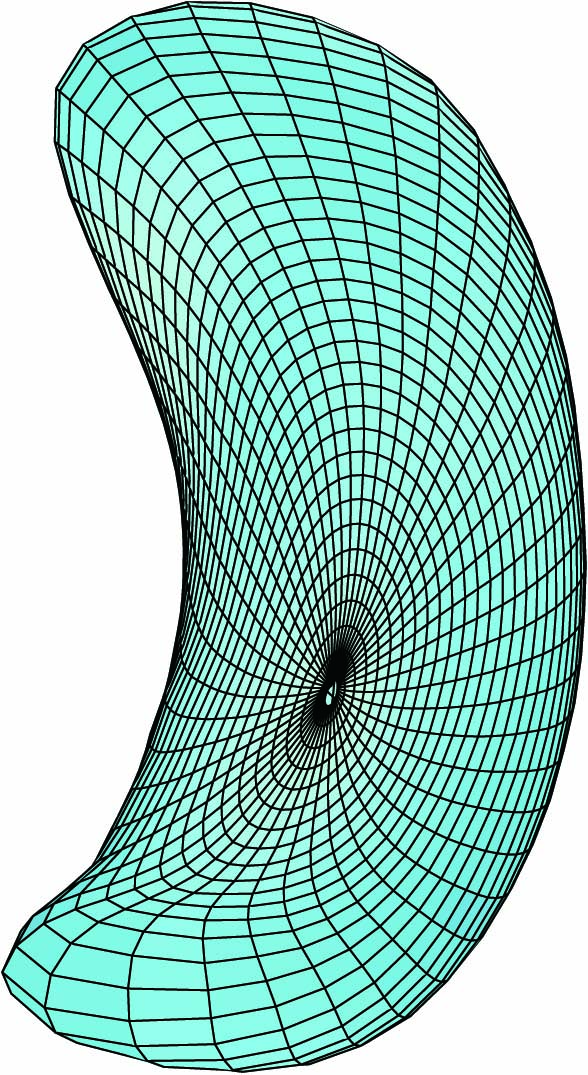}\hspace{0.55cm}
 		\includegraphics[height = 4cm]{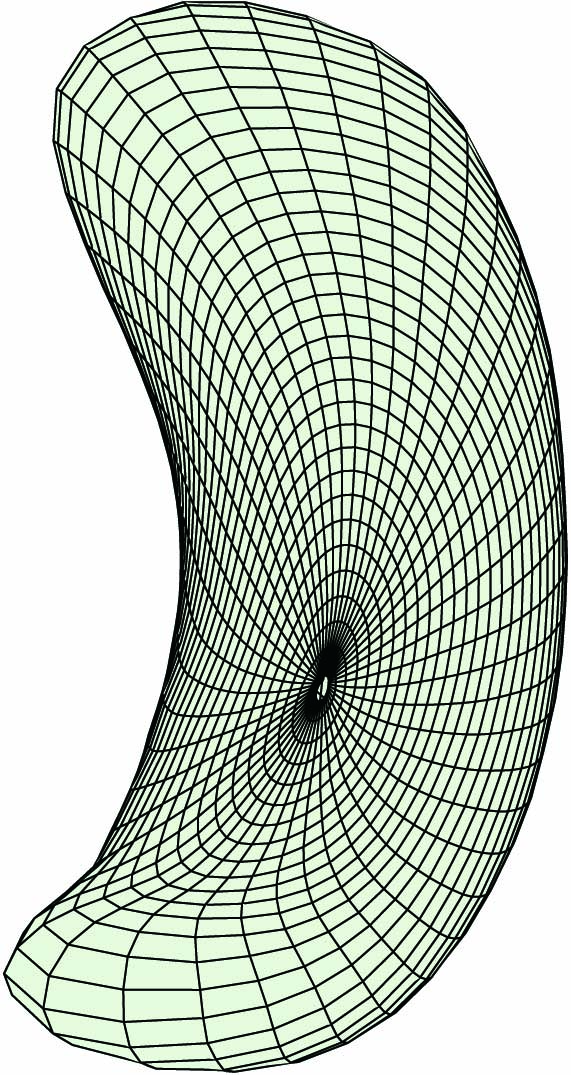}\hspace{0.55cm}
 		\includegraphics[height = 4cm]{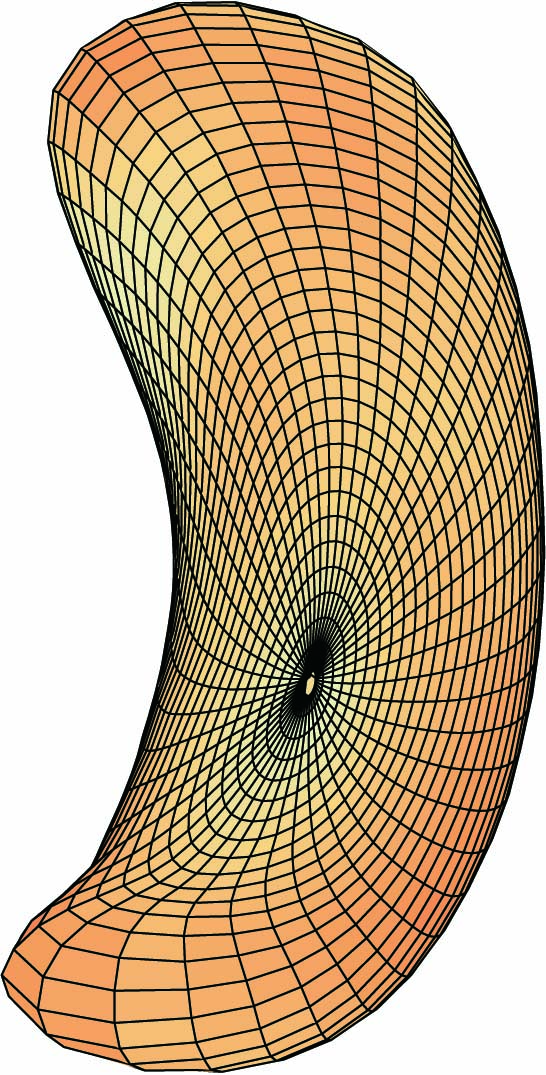}\hspace{0.55cm}
 		\includegraphics[height = 4cm]{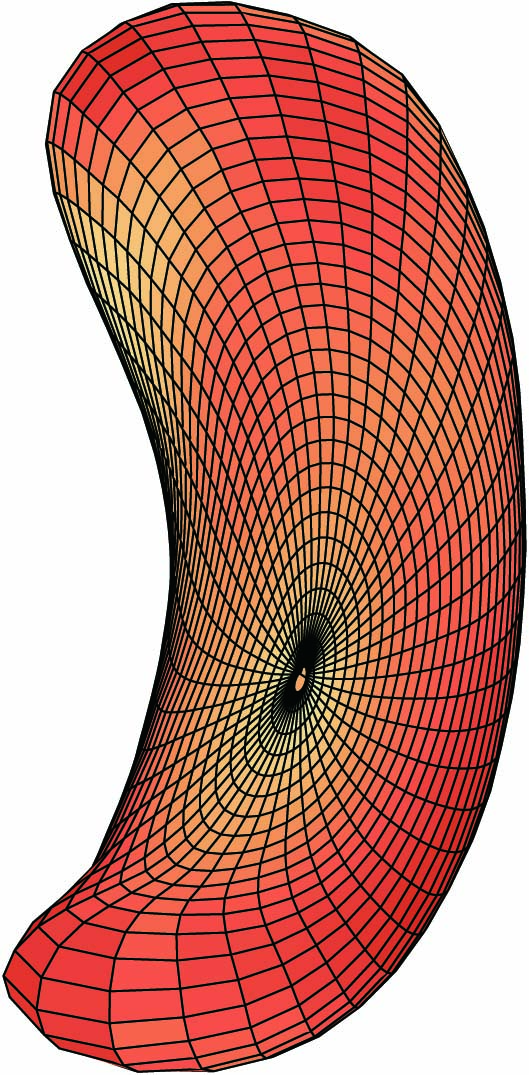}
		\qquad
 		\includegraphics[height = 4cm]{Colorbar2-1-1.jpg}
		}
		\caption*{\Large lower CTQTOT $\leftarrow \rightarrow$ higher CTQTOT}
		\end{minipage}}
		}
		
		\resizebox{.9\linewidth}{!}{
		\fbox{\begin{minipage}{\dimexpr\textwidth-2\fboxsep-2\fboxrule\relax}
		\subfloat[\Large 2nd and 11th Principal Component of Left Amygdala]{
		\qquad
 		\includegraphics[width = .16\linewidth]{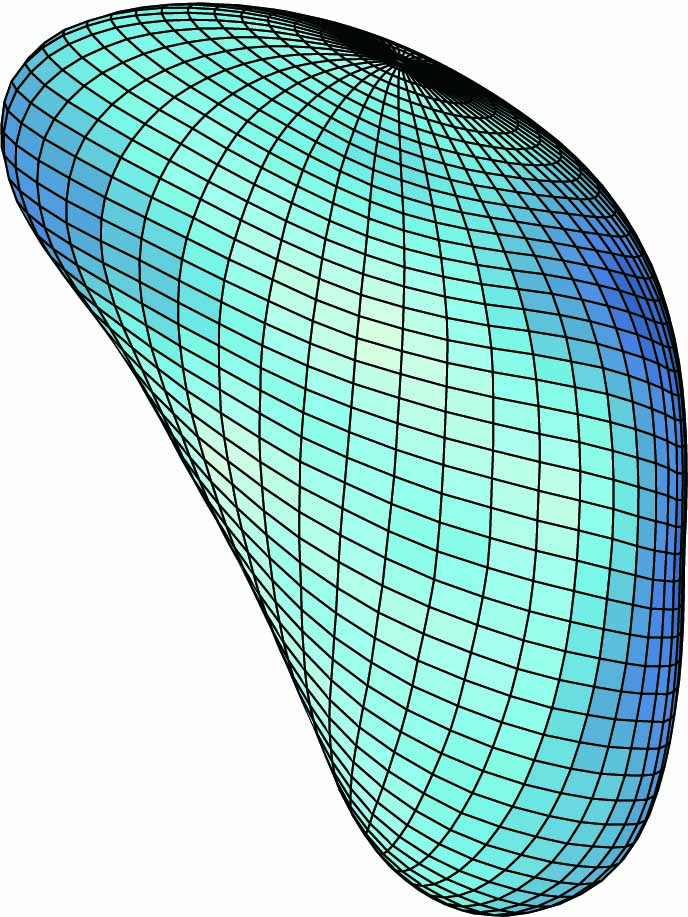}
 		\includegraphics[width = .16\linewidth]{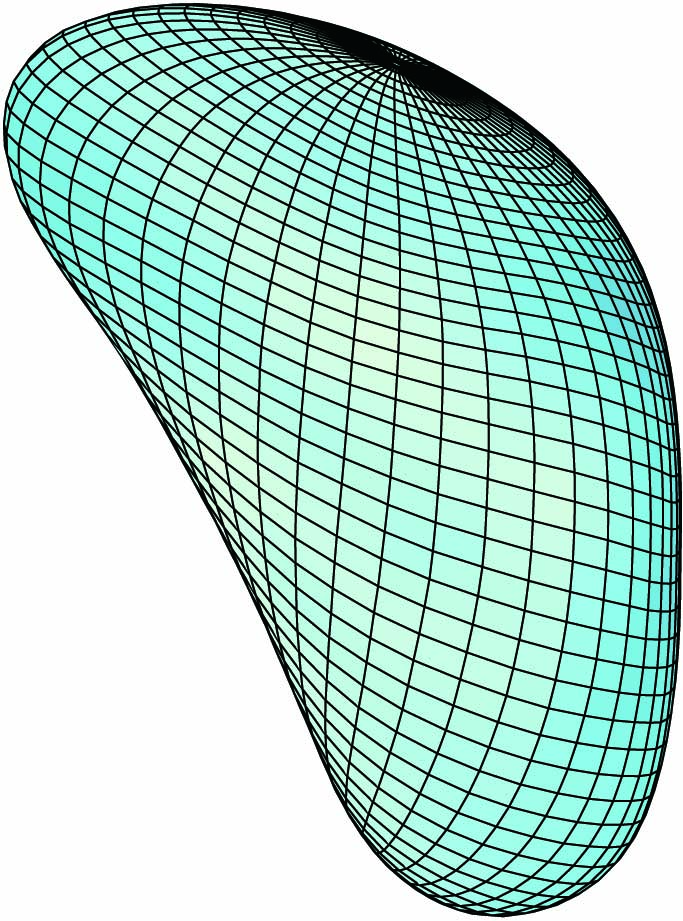}
 		\includegraphics[width = .16\linewidth]{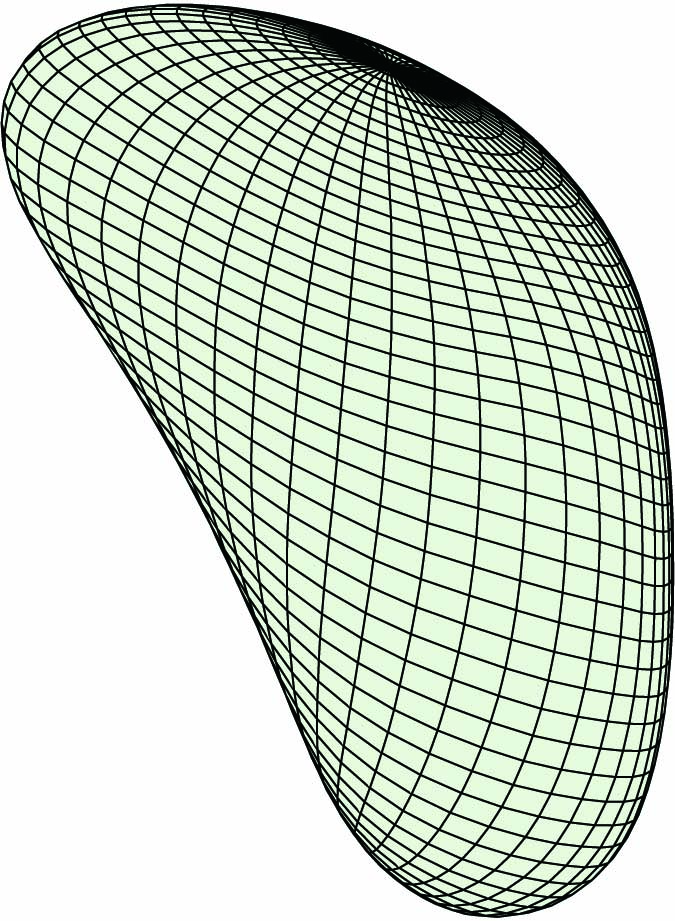}
 		\includegraphics[width = .16\linewidth]{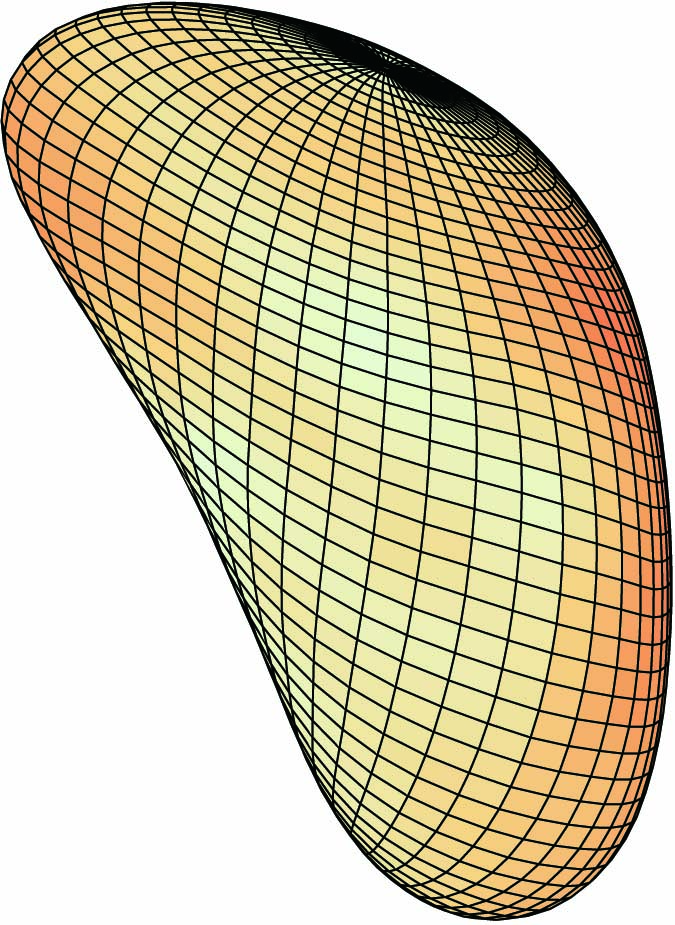}
 		\includegraphics[width = .16\linewidth]{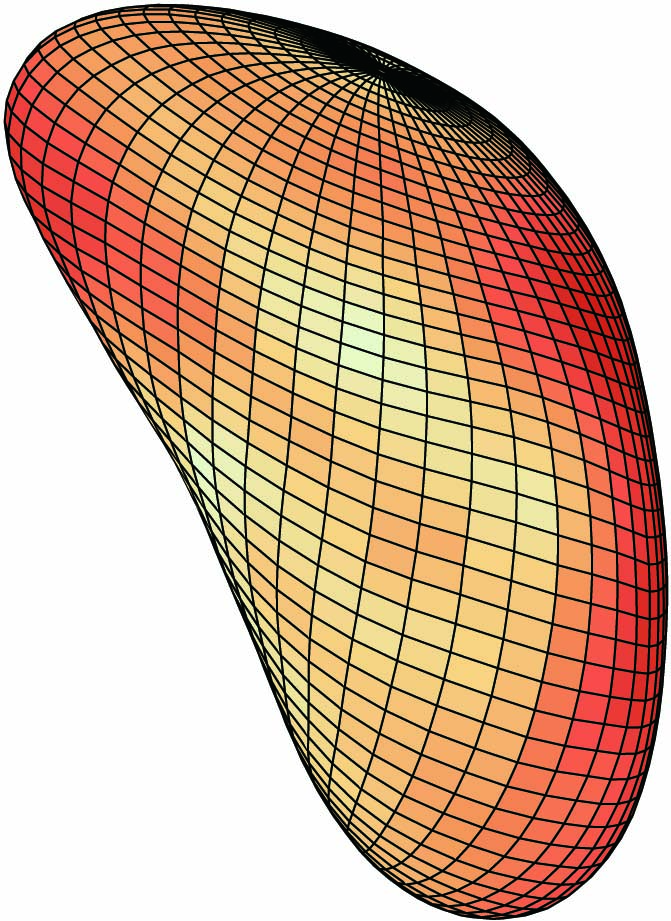}
		\qquad
 		\includegraphics[height = 3.5cm]{Colorbar2-1-1.jpg}
		}
		\caption*{\Large lower CTQTOT $\leftarrow \rightarrow$ higher CTQTOT}
		\end{minipage}}
		}
		
		\resizebox{.9\linewidth}{!}{
		\fbox{\begin{minipage}{\dimexpr\textwidth-2\fboxsep-2\fboxrule\relax}
		\subfloat[\Large 2nd and 4th Principal Component of Left Putamen]{
		\qquad
		\includegraphics[width = .16\linewidth]{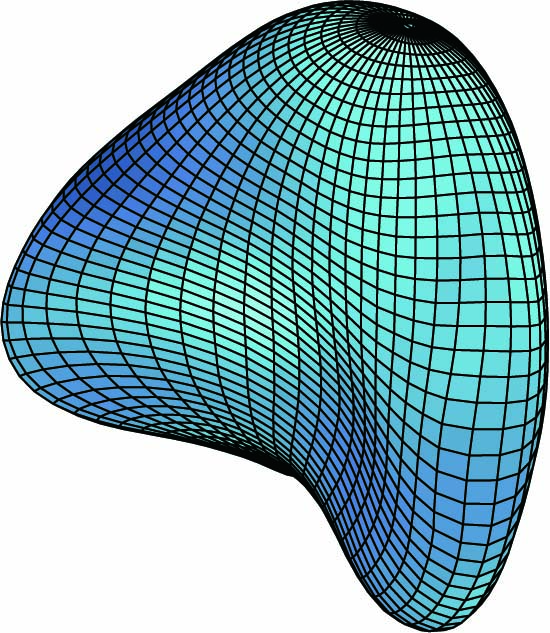}
 		\includegraphics[width = .16\linewidth]{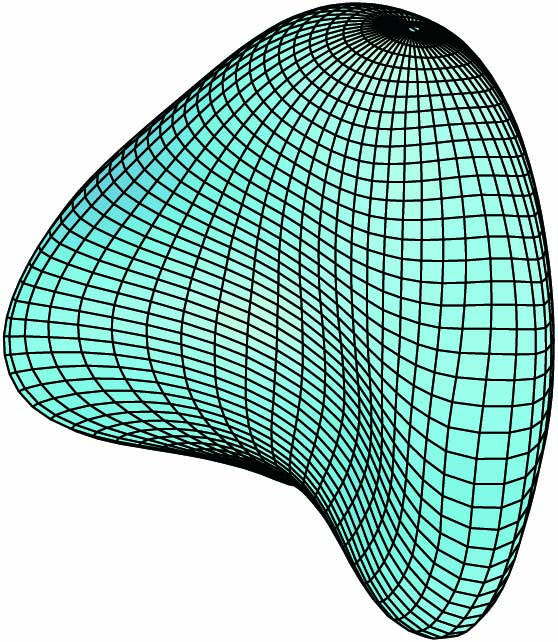}
 		\includegraphics[width = .16\linewidth]{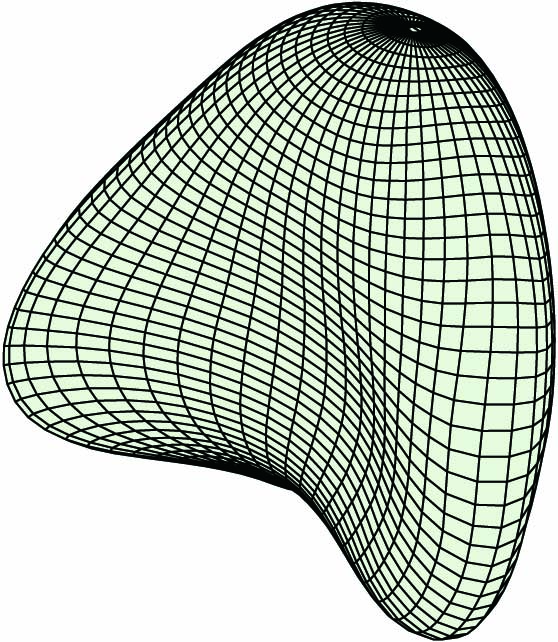}
 		\includegraphics[width = .16\linewidth]{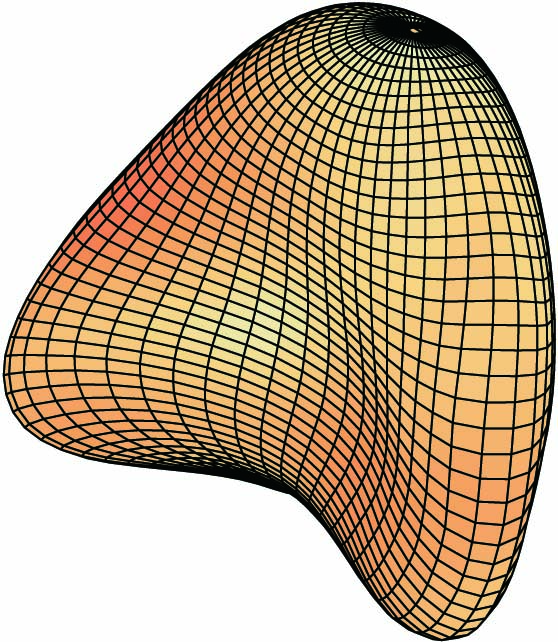}
 		\includegraphics[width = .16\linewidth]{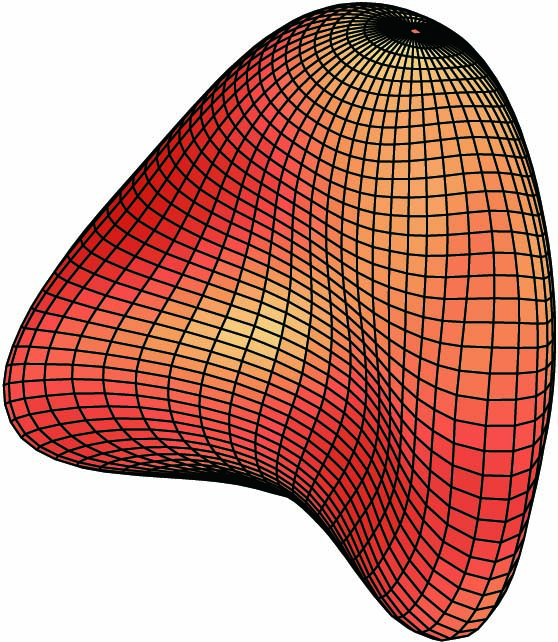}
		\qquad
 		\includegraphics[height = 3.5cm]{Colorbar2-1-1.jpg}
		}
		\caption*{\Large lower CTQTOT $\leftarrow \rightarrow$ higher CTQTOT}
		\end{minipage}}
		}
		\caption{Shape deformation along the most significant principal components on childhood traumatic experience inventory. The surfaces to the left have less or no childhood traumatic experience, and the surfaces to the right have more childhood traumatic experience. Color indicates the small patch's relative shape difference (deformation level) along the direction.}
		\label{pattern CTQTOT}
\end{figure}

Figure~\ref{pattern CTQTOT} shows that with more childhood traumatic experience:
\begin{itemize}
	\item Left hippocampus surface has thinner anterior and posterior ends moving along the negative direction of 5th principal component;
	\item Left amygdala surface has shrunken head end and the left side appears more flattened;
	\item Left putamen surface's middle part is hollower and the end part gets sharper along the negative direction of 2nd and 4th principal component.
\end{itemize}

\fboxsep=5pt
\fboxrule=2pt

By visualizing these shape patterns we find that the shape changes of three subcortical structures following different levels of PTSD symptom and childhood traumatic experience are basically consistent. This, in turn, supports the results of regression models. Subjects with more childhood traumatic experience potentially have more severe PTSD, and the shape of subcortical structures deforms in the same direction.
Additional displays of deformations along the most significant principal components for PTSD symptom scale and childhood traumatic experience inventory are presented in Supplementary Figs. 5 and 6. The interactive graphs to move along different levels of PSS and CTQTOT are also presented in the Supplementary Material.

\subsection{Result Comparisons}
The average inter- and intra-class total point-wise distances for three subcortical structures are computed with Eqns.~\ref{inter_distance} and \ref{intra_distance} for the elastic shape analysis and vertex-wise analysis separately. The results are shown in Table~\ref{distance result}. The inter-class distances computed under vertex-wise analysis are smaller than the intra-class distances, which indicates that the shape differences introduced by misalignment are larger than the differences between populations. That makes it difficult for the vertex-wise analysis to identify shape differences between healthy and PTSD groups. In contrast, with elastic surface registration, the average shape differences between different groups (healthy and PTSD) are larger than the differences within the groups. 
This provides a strong evidence that elastic shape analysis has the ability to discover differences among subcortical shapes that are correlated with PTSD.

\begin{table}[h] 
				\centering
				\caption{Average inter- and intra-class total point-wise distances}
				\label{distance result}
				\begin{threeparttable}
				\centering
				\begin{tabular}{c|c|c|c|c|}
				\textbf{} & \multicolumn{2}{|c|}{\textbf{VERTEX-WISE ANALYSIS}} & \multicolumn{2}{|c|}{\textbf{ELASTIC SHAPE ANALYSIS}}\\
				\hline
				\textbf{Subcortical Structure} & \textbf{Inter-class} & \textbf{Intra-class} & \textbf{Inter-class} & \textbf{Intra-class}\\
				\hline
				amygdala & 603.1 & 608.7 & 135.8 & 130.5\\
				hippocampus & 1263.0 & 1363.9 & 317.7 & 310.7\\
				putamen & 751.3 & 823.8 & 186.4 & 182.6\\
				\end{tabular}
		  \end{threeparttable}
\end{table}
 
To compare the two methods in regression models,  we replace the elastic shape analysis principal scores with vertex-wise analysis principal scores and repeat the experiments. Table~\ref{model comparison} lists the model adjusted $R^2$ values of the two methods. Regression models trained with elastic shape analysis principal scores outperform those trained with vertex-wise analysis principal scores in all 6 models (note that models 3 and model 7 are independent of principal scores). Besides, with same number of principal components, those computed with elastic shape analysis method contain more shape information (shape variation) because of the geometric properties discussed in section \ref{introduction}. 

\begin{table}[h] 
				\centering
				\caption{Adjusted $R^2$ of Regression Models Under Two Methods}
				\label{model comparison}
				\begin{threeparttable}
				\centering
				\begin{tabular}{c|c|c}
				\textbf{Model \#} & \textbf{$R^2$ of Vertex-wise Analysis} & \textbf{$R^2$ of Elastic Shape Analysis}\\
				\hline
				1 & 0.59 & \textbf{0.64}\\
				2 & 0.47 & \textbf{0.48}\\
				3 & 0.19 & 0.19\\
				4 & 0.23 & \textbf{0.31}\\
				5 & 0.56 & \textbf{0.70}\\
				6 & 0.24 & \textbf{0.39}\\
				7 & 0.11 & 0.11\\
				8 & 0.18 & \textbf{0.29}\\
				\end{tabular}
		  \end{threeparttable}
\end{table}

\begin{figure}[!ht]
		\centering
		\resizebox{\linewidth}{!}{
        \fbox{
 		\subfloat[Model 1]{	
 		\includegraphics[width = .25\linewidth]{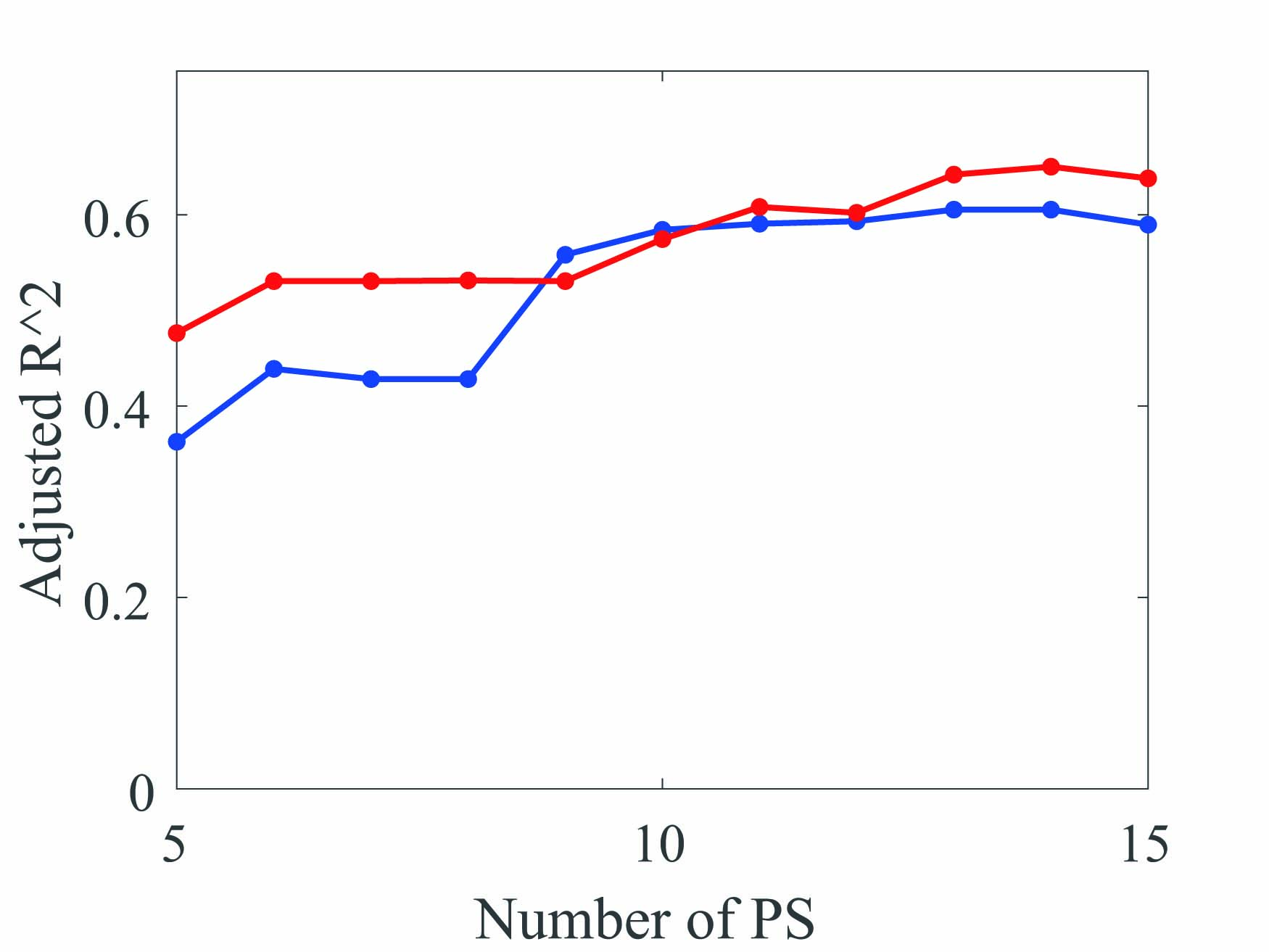}
 		}
 		\subfloat[Model 2]{
         \includegraphics[width = .25\linewidth]{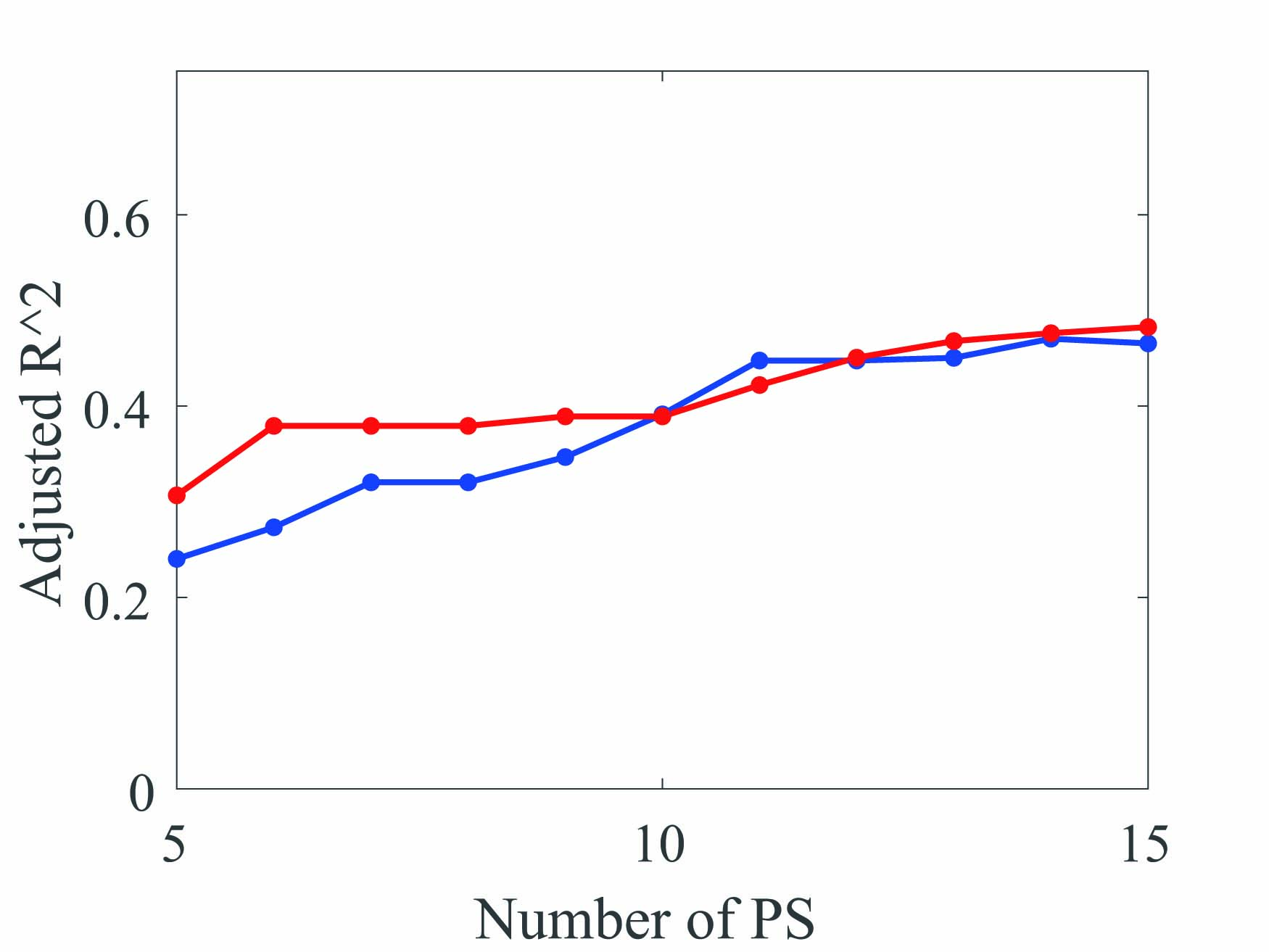}
         }
 		\subfloat[Model 3]{
         \includegraphics[width = .25\linewidth]{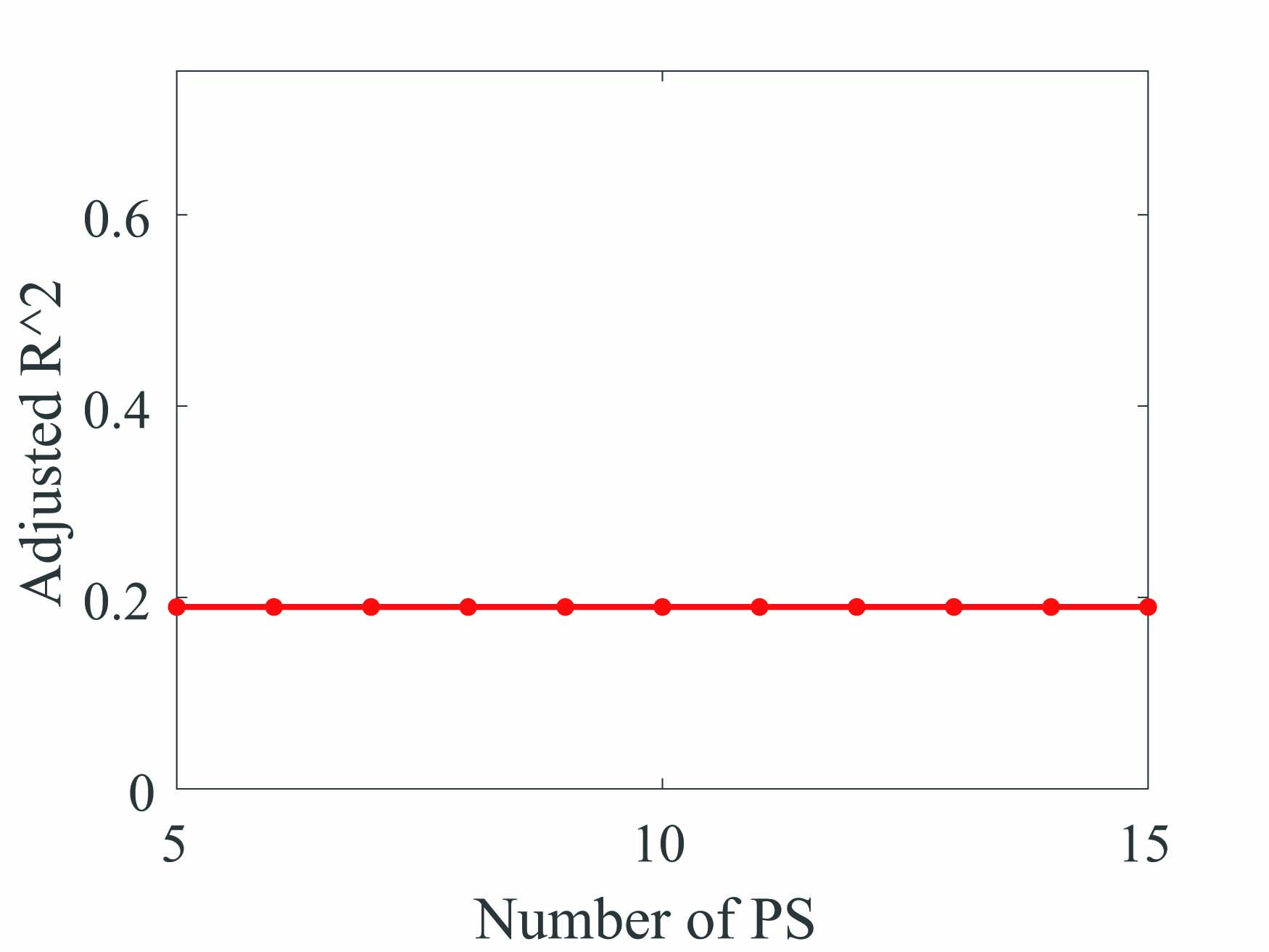}
         }
 		\subfloat[Model 4]{
 		\includegraphics[width = .25\linewidth]{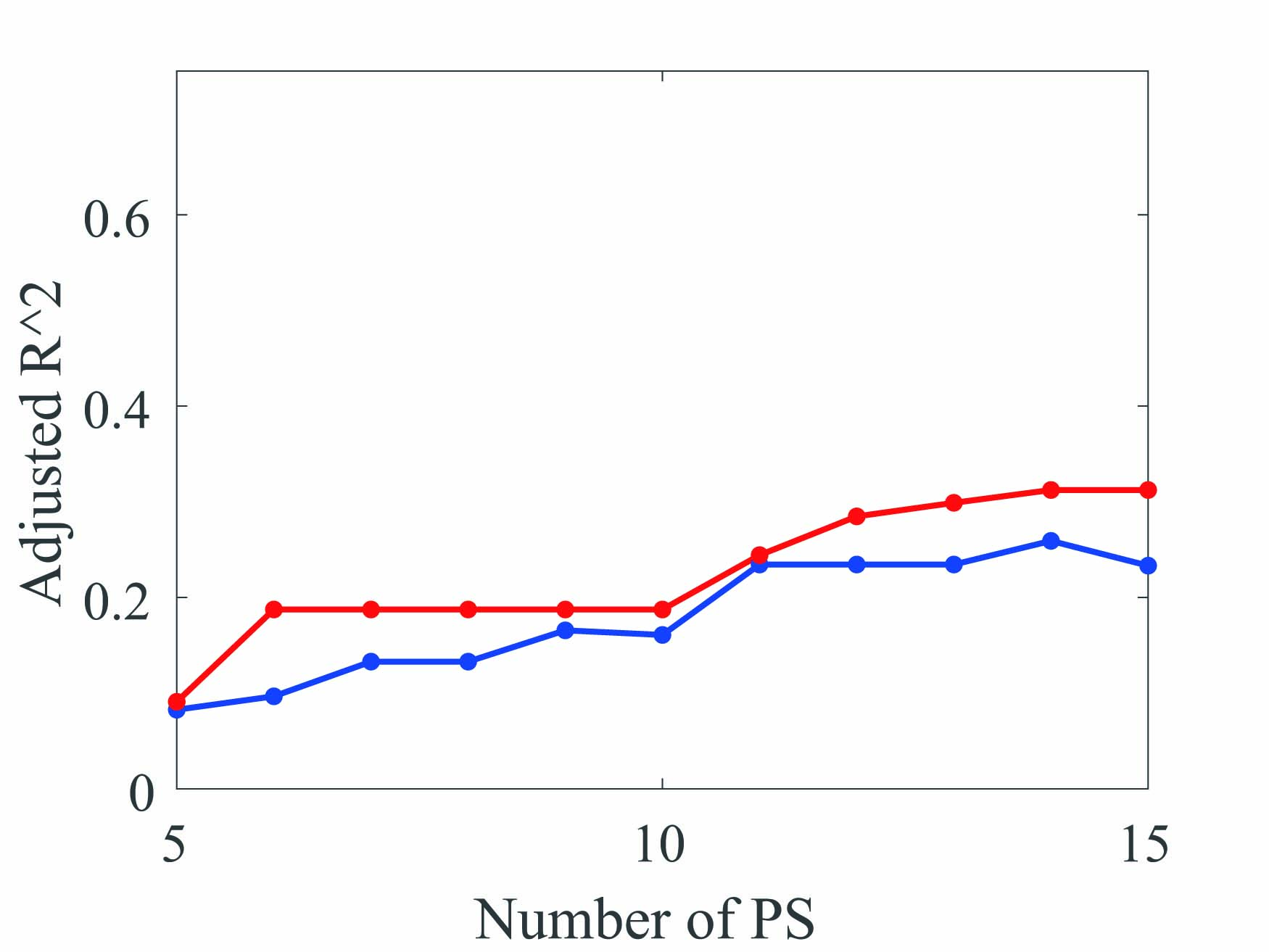}
 		} 
   }}
		\resizebox{\linewidth}{!}{
        \fbox{
 		\subfloat[Model 5]{
 		\includegraphics[width = .25\linewidth]{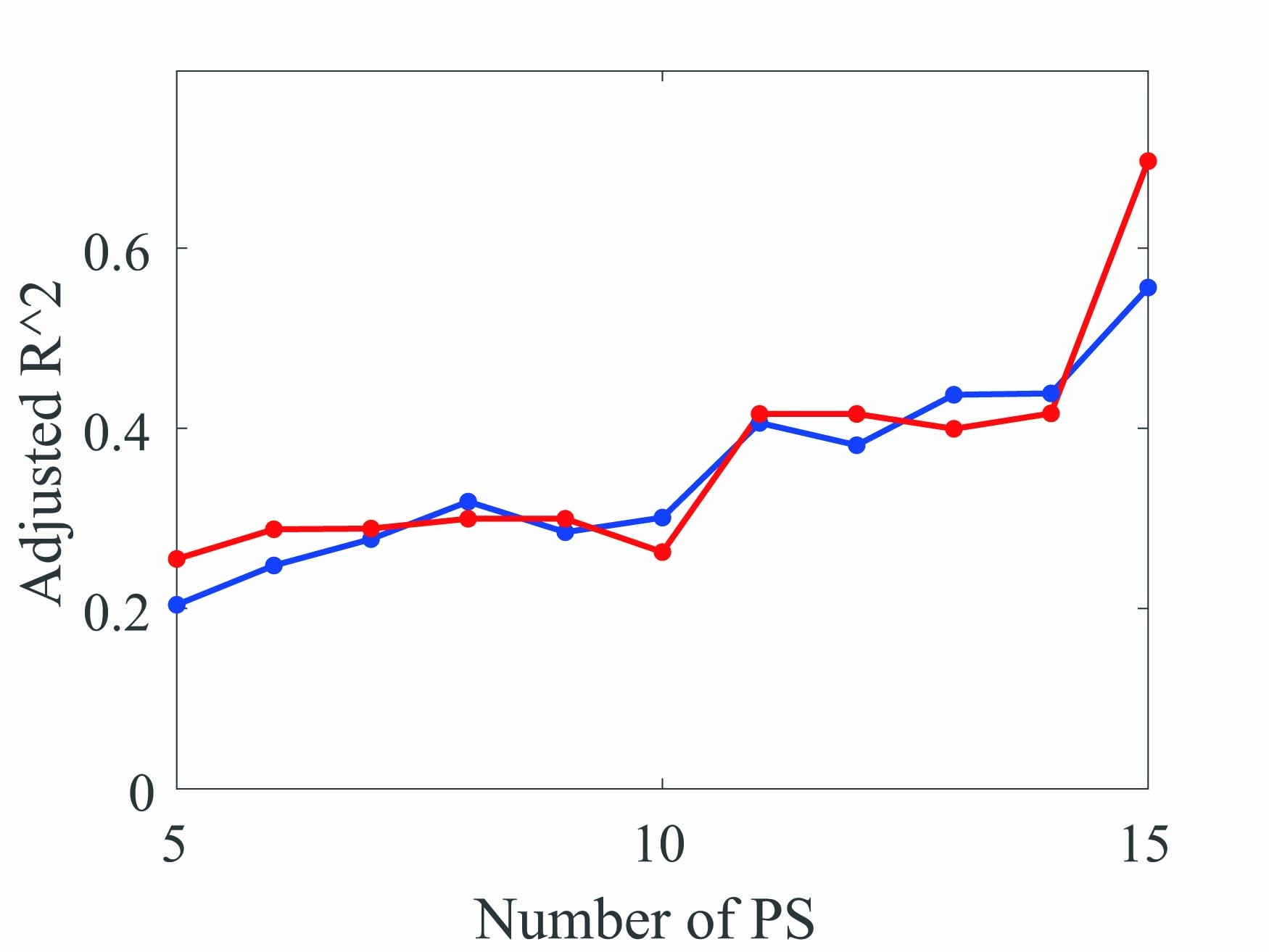}
 		}
 		\subfloat[Model 6]{
         \includegraphics[width = .25\linewidth]{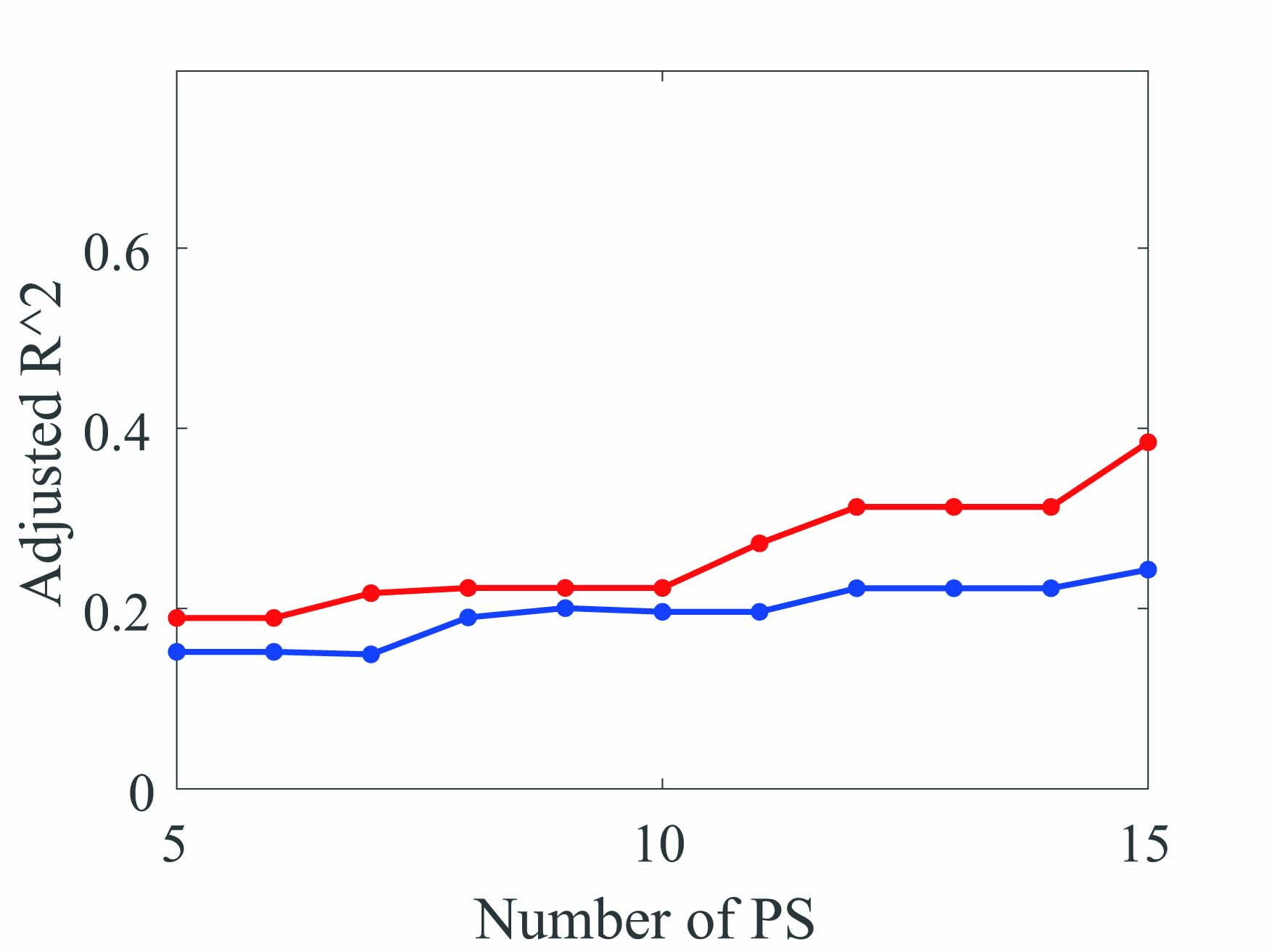}
         }
 		\subfloat[Model 7]{
         \includegraphics[width = .25\linewidth]{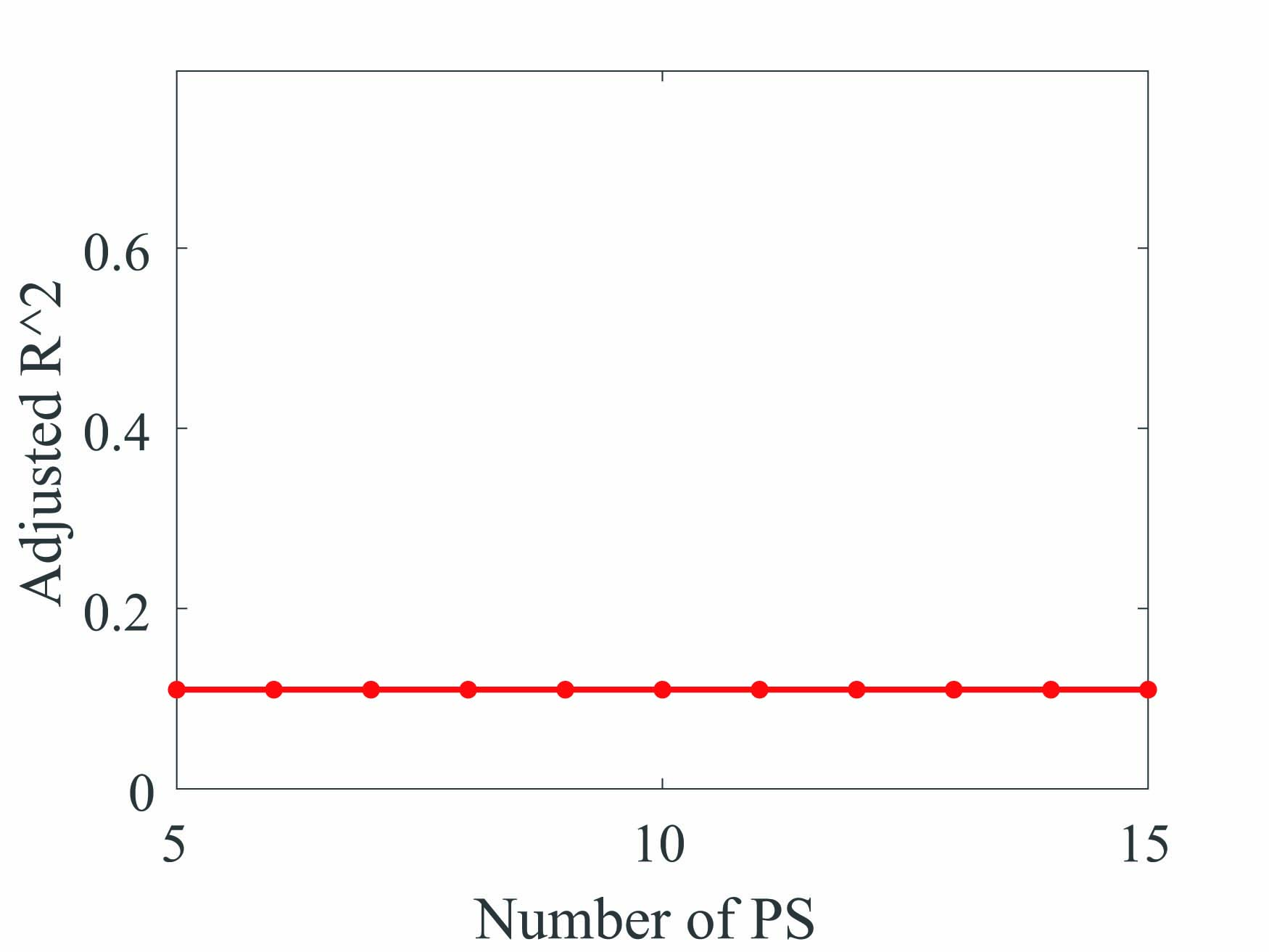}
         }
 		\subfloat[Model 8]{
 		\includegraphics[width = .25\linewidth]{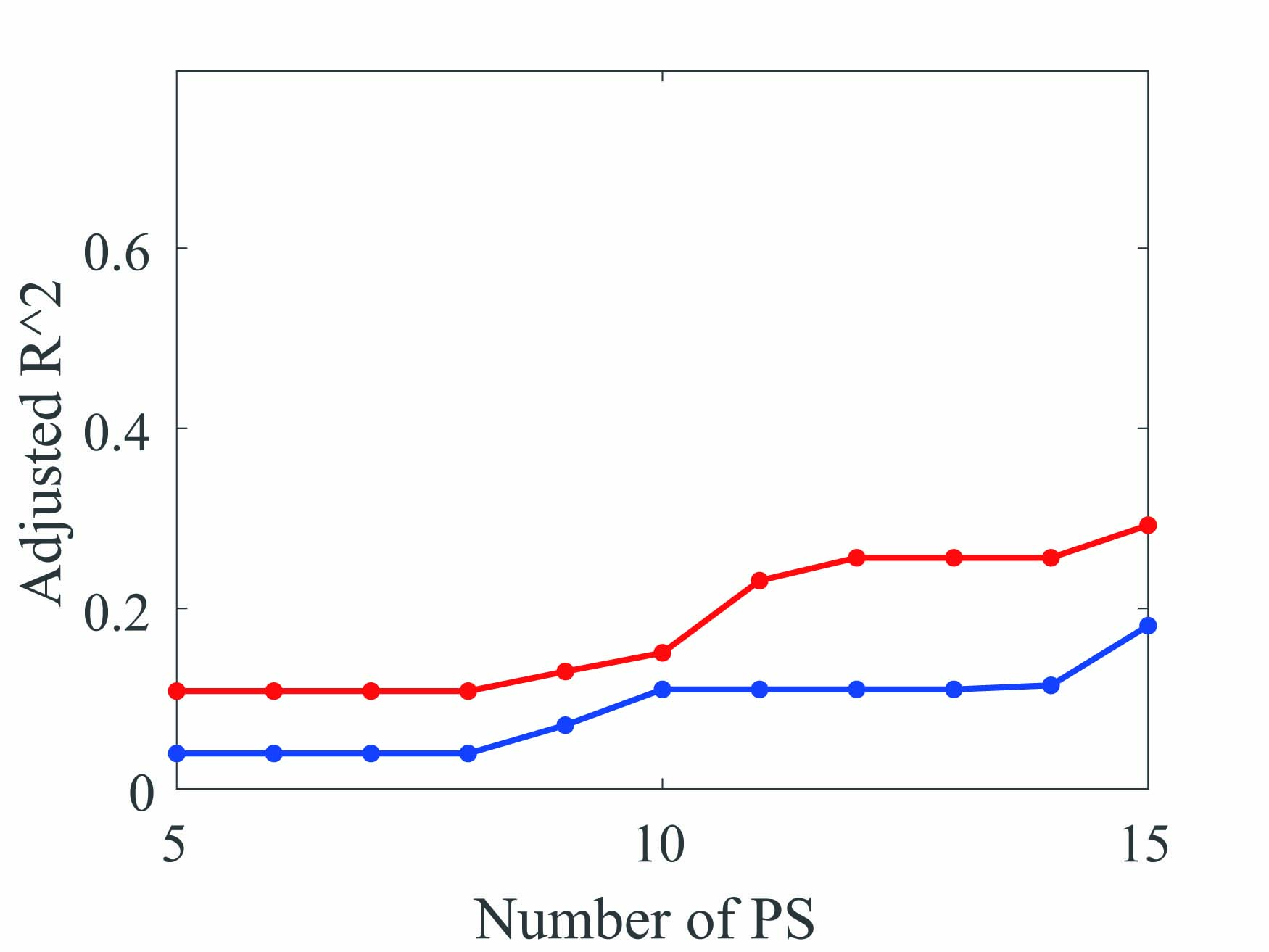}
 		}
		}}
		\caption{Models' $R^2$ trained with different number of principal components (5-15) for both methods. In each plot, red line stands for elastic shape analysis and blue line stands for vertex-wise analysis.} 
		\label{ModelR2}
\end{figure}

We also train the first eight models with different number of principal components from 5 to 15 under both methods to test and compare their performances. In Fig.~\ref{ModelR2}, the red lines stand for the $R^2$ values for models trained under elastic shape analysis, and blue lines stand for those under vertex-wise analysis. Models 3 and 7 are independent of principal components, so the $R^2$values don't change with different number of principal components, and are same for both the methods. In the other six models, most of the red lines are above blue lines, which implies that the models trained with elastic shape analysis principal scores tend to have higher $R^2$ under different number of principal components.

\fboxsep=0pt 
\fboxrule=1pt 

From these results, we conclude that elastic shape analysis is more effective and accurate in identifying the shape differences of subcortical structures correlated with PTSD and childhood traumatic experience, when compared to the widely applied vertex-wise analysis method.

\section{Discussion}

We have proposed a comprehensive shape-analysis approach that treats the brain structures as continuous surfaces instead of a collection of discrete points. One of the key aspects of the approach is that it incorporates crucial registration steps such as rigid motions, 
global scaling, and parameterizations of surfaces in a unified way. It uses a novel SNRF representation and an elastic metric that appropriately measures geodesic distances between shapes while registering them. Incorporating these important registration steps when comparing shapes helps reduce errors due to additional pre-processing and registration steps that are routinely employed in existing shape analysis approaches and helps enhance the accuracy of the proposed method.
Another important feature of the proposed approach is that it registers shapes by pairwise deformations and comparisons, and it does not need a standardised brain template for registration and shape analysis. This is a useful feature that is not present in existing shape analysis methods that usually employ a standardized brain template for registration and subsequent shape analysis (for example \cite{klein:2011}). In addition, the proposed approach is able to compute a standardized template that represents the average shape of the sample of images in the data via the Karcher mean, and is further able to provide confidence intervals around this Karcher mean that provides measures of uncertainty corresponding to the brain shapes in the population. 

Our analysis of the MRI neuroimaging data for trauma-exposed participants illustrated that the proposed approach was able to capture more than 95\% of the variability in subcortical shapes using a moderate number of principal components, whereas a considerably higher number of components were needed to capture similar levels of variability under an alternate state-of-the-art ICP method. We conclude therefore that the elastic shape analysis comprises a more parsimonious characterization of the shape of subcortical brain regions. This provides a benefit for a number of modeling techniques that would benefit from sparser representation of the neural features of interest. 

These principal components were then used as shape features to predict continuous clinical measures in PTSD in conjunction with additional exposure variables such as trauma and their interactions. This joint analysis is a significant advantage over the vertex-wise analysis, where such interactions are challenging to include because of an inflated number of model parameters. The predictive analysis yielded high $R^2$ values that were considerably higher than what is typically observed in the psychiatric neuroimaging literature. We were able to explain a unique 29\% of the variance in PTSD symptom severity using the principal scores, above and beyond effects of age or depressive symptoms. In contrast, large collaborative meta-analyses of PTSD neuroimaging biomarkers find small effect sizes ranging from $d=0.06-0.17$ across subcortical volumes, major white matter tracts, and regional cortical thickness \citep{enigma, dennis2019altered, wang2020cortical}, and are often not able to co-vary for potential comorbid conditions.

Furthermore, in comparison with the FSL FIRST analysis, elastic shape analysis produced a 5\% increase in sensitivity for the association with PTSD symptoms and a 14\% increase for the association with childhood trauma exposure. The current findings suggest that these minor effects may arise in part from methodological issues with signal detection and precision in post-acquisition analysis of the images. This is encouraging and suggests that precision psychiatric biomarkers may become more feasible and translatable with additional development of analytic tools for characterizing brain structural alterations. We must also acknowledge, however, the major role that heterogeneity in patient populations and symptom presentation play in moderating effect sizes in PTSD. We conjecture that the increased sensitivity observed under elastic shape analysis is due to the incorporation of accurate shape features (principal components), along with the subsequent capacity to incorporate supplementary variables and their interactions in our analysis. 

We identified an association between PTSD symptom severity and complex alterations in the hippocampus, amygdala, and putamen. With increasing PTSD symptom severity, the left hippocampus showed shrinkage of the medial wall of the head as well as lateral aspects of the tail, producing a more curved body shape. Although elastic shape analysis is not designed to investigate specific subfields of the hippocampus, this could be consistent with the location of CA1 and/or subiculum along the longitudinal axis of the hippocampus, consistent with prior work \citep{chen2018smaller, bae2020volume}. The left amygdala showed an indentation in superior aspects located near central subnuclei. This differs from prior literature in male veterans showing either no shape differences \citep{bae2020volume}, or larger central and medial nuclei \citep{morey2020amygdala, klaming2019expansion}. In contrast, our study was conducted in women exposed to civilian traumas such as interpersonal violence. Lastly, the left putamen showed greater concavity with thinning and sharpening of the medial end, near the nucleus accumbens. The link with brain regions involved in motivation and reward is interesting given the affective and motivational blunting observed in the numbing symptoms of PTSD, although striatal morphology has received very little attention in studies of trauma and PTSD. 

The utility of the principal components as shape features are provided via a Matlab GUI interactive tool that enables one to visualize how the brain shape changes as the first few principal components are varied (provided in Supplementary Materials). Such a tool provides a novel way to visualize changes in the brain shape that is expected to have an important impact for investigators.

\section{Conclusion}
This study uses the elastic shape analysis to compute shape summaries (mean, covariance, PCA) of subcortical data from the Grady Trauma Project (GTP). Having obtained PCA-based low-dimensional representation of shapes, we build regression models to predict PTSD clinical measures that use shapes of hippocampus, amygdala, and putamen as predictors and that have considerably great predictive power. Furthermore, we localize and visualize the subcortical shape deformations related to change in PTSD severity. This tool can also provide physicians and clinicians a novel way to visualize localized changes or deformations in the brain anatomy for statistically significant shape features or principal components. Prospective studies can be carried out in larger data sizes and involving additional subcortical structures to improve predictions of PTSD clinical measures.

\section*{Acknowledgement} 
This research was conducted in part with support from NIH R01 MH120299 and NSF DMS 1953087. 

\section*{Data and Code Availability Statement} 
The Supplementary Materials are available online using the following Dropbox link 
\\(https://www.dropbox.com/sh/1sa45k5do61koet/AACuZMTQtPmxpMNHkJv8ZUCSa?dl=0). 

The codes to run the elastic shape analysis pipeline and build regression models can be found in the GitHub repository (https://github.com/wuyx5/Elastic-Shape-Analysis-PTSD). Due to privacy and ethical considerations, the data cannot be made openly available at this stage. However, the data will be shared upon reasonable request.

\section*{Ethics Statement}
Study procedures were approved by the institutional review board of Emory University, informed consent was obtained for experimentation with human subjects, and procedures were consistent with the Declaration of Helsinki.

\section*{Declaration of Competing Interests}
The authors declare that they have no known competing financial interests or personal relationships that could have appeared to influence the work reported in this paper.

\section{Bibliography}
\nocite{*}
\bibliography{Bib} 
\bibliographystyle{plainnat}

\end{document}